\documentclass[12pt]{article}
\usepackage{graphicx,psfrag,epsf,color}
\usepackage{setspace}
\usepackage{amssymb}
\usepackage{epsfig, graphicx, amsmath}
\usepackage{natbib}
\usepackage{multirow}
\usepackage{bm, color}
\usepackage{comment}
\usepackage[ruled]{algorithm}
\usepackage{algpseudocode}
\usepackage{enumerate}
\usepackage{url} 
\usepackage{array}
\usepackage{booktabs}
\usepackage{float}
\usepackage[font=small,labelfont=bf]{caption}
\setcitestyle{citesep={;}}
\newtheorem{lemma}{Lemma}
\newtheorem{theorem}{Theorem}

\newcommand{\blind}{0}

\newcommand{\Var}{{\mbox{Var}}}

\newcommand{\prob}{{\mbox{Pr}}}

\newtheorem{prop}{Proposition}

\def\EE{\mathbb{E}}

\newcommand{\tsb}[1]{\textnormal{#1}}
\newcommand{\TV}{\tsb{\textrm{TV}}}
\newcommand{\KL}{\tsb{\textrm{KL}}}

\def\begar{$$\begin{array}{lll}}
\def\endar{\end{array}$$}
\def\begarlab{\begin{equation} \begin{array}{lll} \label}
\def\endarlab{\end{array} \end{equation}}
\def\argmax{\text{argmax}}
\def\argmin{\text{argmin}}

\def\ds1{{\mathrm{1 \hspace{-2.6pt} I}}}

\def\calA{{\cal A}}

\def\calD{{\cal D}}

\def\calI{{\cal I}}

\def\calN{{\cal N}}

\def\calS{{\cal S}}

\def\calV{{\cal V}}

\def\floor#1{\lfloor #1 \rfloor}

\usepackage{xr}

\def\old{\scriptsize{\tsb{old}}}
\def\new{\scriptsize{\tsb{new}}}
\newcommand{\change}[1]{\textcolor{black}{#1}}

\begin{document}

\def\spacingset#1{\renewcommand{\baselinestretch}%
{#1}\small\normalsize} \spacingset{1}
%
%
\if0\blind
{
			\title{\bf Value Enhancement of Reinforcement Learning via Efficient and Robust Trust Region Optimization}
			\date{\empty}
			\author{Chengchun Shi$^{a}$\thanks{
					 The first two authors contribute equally to this paper. 
				}\hspace{.2cm}, Zhengling Qi$^{b*}$\thanks{qizhengling@gwu.edu;zhoufan@mail.shufe.edu.cn}, Jianing Wang$^{c}$ and Fan Zhou$^{c\dagger}$\\
				$^a$Department of Statistics, London School of Economics and Political Science \\
				$^b$Department of Decision Sciences, The George Washington University\\
				$^c$Department of Statistics, Shanghai University of Finance and Economics
			}
			\maketitle
	} \fi

\baselineskip=21pt

\begin{abstract}
	Reinforcement learning (RL) is a powerful machine learning technique that enables an intelligent agent to learn an optimal policy that maximizes the cumulative rewards in sequential decision making. Most of methods in the existing literature are developed in \textit{online} settings where the data are  easy to collect or simulate. Motivated by high stake domains such as mobile health studies with limited and pre-collected data, in this paper, we study \textit{offline} reinforcement learning methods. To efficiently use these datasets for policy optimization, we propose a novel value enhancement method to improve the performance of a given initial policy computed by existing state-of-the-art RL algorithms. Specifically, when the initial policy is not consistent, our method will output a policy whose value is no worse and often better 
	than that of the initial policy. When the initial policy is consistent, under some mild conditions, our method will yield a policy whose value converges to the optimal one at a faster rate than the initial policy, achieving the desired ``value enhancement" property. The proposed method is generally applicable to any parametrized policy that belongs to certain pre-specified function class (e.g., deep neural networks). Extensive numerical studies are conducted to demonstrate the superior performance of our method. 
\end{abstract}

\noindent%
\vfill

\section{Introduction}
Reinforcement learning \citep[RL, see e.g.,][]{sutton2018reinforcement} is concerned with how agents take sequential actions in dynamic environments, with the main goal of maximizing the cumulative rewards they receive. In recent years, we have seen tremendous achievements of RL in artificial intelligence (AI). For example, \textit{AlphaGo} \citep{silver2016mastering}, one of the most successful applications in AI, makes use of reinforcement learning and deep learning algorithms for teaching machines to play the board game called Go, and has beaten many top human players. The appealing performance of RL has also been demonstrated in many scientific fields. \change{In medical applications}, RL has been used to help clinicians make better treatment decisions for patients with sepsis \citep{komorowski2018artificial}. In economics, econometricians often study dynamic discrete choice models \citep{rust1987optimal} in order to understand the behavior of rational agents, which is similar to the inverse RL problem \citep{abbeel2004apprenticeship}. In operations research, RL has been widely applied to business operations such as supply chain management, finance and logistics \citep{hubbs2020or}. For an overview of various applications of RL, we refer to Section 5 of \cite{li2017deep}. 


Our research in this paper is partly motivated by recently emerging mobile health (mHealth) studies. Advancements in mobile and sensor technologies provides us with a unique opportunity to deliver health interventions at anytime and anywhere for promoting healthy behaviors such as regular physical activities and preventing drug abuse, etc. For example, the OhioT1DM dataset \citep{marcolino2018impact} was developed for promoting blood glucose level prediction in order to improve the health and wellbeing of people with type 1 diabetes. It contains data information of 6 people for 8 weeks. For each patient, their treatment information was collected during insulin pump therapy with continuous glucose monitoring (CGM). In addition, blood glucose levels and self-reported times of meals and exercises were also constantly measured and recorded via a custom smartphone app. Finding an optimal insulin pumping policy for each patient at different scenarios may potentially improve their health  status \citep{shi2020statistical}. This matches the goal of RL algorithms.

A fundamental question we aim to investigate here is how to learn an optimal policy efficiently from the batch data in high-stake domains such as mHealth.
Solving this question faces at least two major challenges. First, different from the standard clinical trial data, mobile health data usually consist of a large number of decision points for each patient but the number of patients may be limited (e.g., in OhioT1DM dataset, 6 patients with a few thousands decision points). This posits a unique challenge for searching an optimal policy. In statistics, there is a rich literature in studying dynamic treatment regimes \citep[DTR, see e.g.][]{murphy2003optimal,chakraborty2013statistical,qian2011performance,zhao2015new,shi2018high,wang2018quantile}. For a review of DTR, see \cite{laber2014dynamic}, \cite{kosorok2019precision} and \cite{tsiatis2019dynamic}. However, these methods are mainly designed for only a few treatment decision points and often require a large number of patients in the observed data in order to be consistent. 

Second, different from online RL domains such as video games, where actively interacting with the environment is feasible and data are easy to generate or simulate, in some high stake domains, data are often pre-collected according to some experimental design and very limited. With such limited data, it is essential to study how to efficiently learn the optimal policy from the batch data. We remark that the main focus of RL in the computer science literature is for online learning. Among all the methods available, Q-learning is arguably the most popular model-free RL algorithms \citep{watkins1992q}. It derives the optimal policy by learning an optimal Q-function (see the definition of Q-function in Section \ref{sec: notation}). Follow this line of research, variants of Q-learning methods have been proposed including the fitted Q-iteration \citep[FQI,][]{ernst_tree-based_2005,yang2020theoretical}, 
deep Q-network \citep[DQN,][]{mnih2015human}, among many others. Policy-based learning is another class of RL algorithms that searches the optimal policy among a parametrized policy class. Some popular algorithms include REINFORCE and actor-critic methods \citep[see e.g.,][Chapter 13]{sutton2018reinforcement}. 
Since these algorithms are primarily motivated by the application
of developing artificial intelligence in online video games, 
their generalization to offline settings such as mobile health applications
remain largely unexplored. 


To address the first challenge, we model the observed data by a time-homogeneous Markov decision process \citep[MDP][]{puterman1994markov}. This framework is particularly suitable to model the data collected from mobile health studies where the total number of decision points are often large  \cite[see e.g.,][]{liao2019off,liao2020batch}. The assumptions of Markov and time homogeneity enable a consistent estimation of the optimal policy even with only a few patients.

To address the second challenge, we develop a novel procedure to derive the optimal policy. Recently, a few algorithms have been developed in the statistics literature for policy optimization in mHealth applications \citep{ertefaie2018constructing,luckett2019estimating,liao2020batch,hu2021personalized}. In particular, \cite{liao2020batch} proposed a statistically efficient batch policy learning method under the average reward MDP. However, due to the policy dependent structure of nuisance functions such as Q-function and the marginal density ratio, their proposed algorithm is computational inefficient as it requires updating the nuisance functions estimation in each iteration of their policy gradient decent algorithm. 
Instead of proposing a specific algorithm for policy optimization, we devise a ``value enhancement" method that is generally applicable to any given initial policy computed by some state-of-the-art RL algorithm to improve their performance. 
Basically, after employing some computational efficient RL algorithm and obtaining an initial policy,  we take a one-step update of this policy via efficiently estimating the value enhancement component (defined in Section \ref{sec:TRPO}) and solving a \change{constrained} optimization problem, thus taking advantage of computational efficiency from existing state-of-art RL algorithms without requiring iteratively updating the nuisance functions. More importantly, the proposed procedure guarantees that when the initial policy is not consistent, the output policy by the proposed algorithm is no worse and often better than the initial policy. If consistent, our method will yield a policy whose value converges to the optimal one at a faster rate, achieving the desired ``value enhancement" property. 
Recently, in the computer science literature, \cite{kallus2020statistically} developed an offline policy gradient algorithm that considered statistically efficient estimation of the policy gradient. Our proposal differs from theirs in that we focus on developing a general value enhancement tool that is applicable to any existing RL algorithms to improve their performances. 

Our method is  inspired by Lemma 6.1 in \cite{kakade2002approximately} and the trust region policy optimization algorithm by \cite{schulman2015trust}, which was originally designed for the online setting. 
A key observation is that, the value difference between any two policies can be decomposed into a first-order component and a higher-order remainder term. 
The higher-order term can be lower bounded, based on which a minorization function can be constructed for the value function of any policy. 
One big advantage of working with this minorization function is that it intrinsically disentangles the policy-dependent structure of nuisance functions. This ensures the computational efficiency of the proposed algorithm. 

The key ``value enhancement" property relies crucially on statistically efficient estimation of the first-order term in the decomposition. In online settings, \cite{schulman2015trust} proposed to simulate data trajectories to estimate this quantity. However, in offline settings, it remains unknown how to effectively evaluate this quantity based on the observed data. 
By leveraging semi-parametric statistics, we develop a triply robust estimator for the first-order term that is shown to achieve the efficiency bound when compared to the initial policy. By optimizing the proposed estimator, we are able to improve the value of the initial policy. The triply robustness property guarantees that the ``value enhancement" property holds even when some nuisance function models are misspecified. 
The semi-parametric efficiency guarantees that the value can be enhanced at a sufficiently fast rate. This ensures the statistical efficiency of the proposed algorithm, which is necessary in the offline setting.

In theory, we establish the value enhancement property 
under mild conditions on the nuisance function estimators. In particular, we only require them to converge at a nonparametric rate. See Section \ref{sec:theory} for details. This nice property is achieved mainly due to the
innovative way we put together these nuisance function estimators, which leads to the triply-robust estimator with a parametric convergence rate for the first-order term. In addition, we remark that all our theoretical results related to estimation are established in terms of total decision points, thus showing the proposed method is generally applicable even when the number of trajectories is small but the length of each trajectory is large, which is commonly seen in the mobile health applications.

The rest of this paper is organized as follows. In Section 2, we introduce the offline RL problem in the framework of a time-homogeneous MDP and review the trust region algorithm. 
In Section 3, we present our value enhanced policy optimization method and the related estimation. In Section 4, we study statistical properties of our algorithm. In Section 5, extensive numerical studies including a toy example demonstrating the value enhancement property, a real-data driven simulation study and a real data application are conducted to demonstrate the superior performance of the proposed method. Finally, we conclude our paper in Section \ref{sec:dis}. 
All technical proofs and details can be found in the 
online supplementary material.

%
\section{Preliminaries}
This section is organised as follows. We first introduce the offline data structure and describe the model setup in Section \ref{sec:problemformulation}. In Section \ref{sec: notation}, we introduce some notations needed to derive our method. In Section \ref{sec:TRPO}, we review the trust region policy optimization (TRPO) method proposed by \cite{schulman2015trust} for online learning, as it is closely related to our approach.
\subsection{Value function and the optimal policy}\label{sec:problemformulation}
Consider a single trajectory $\{(S_{t},A_{t},R_{t})\}_{t\ge 0}$ where $(S_{t},A_{t},R_{t})$ denotes the state-action-reward triplet collected at time $t$. We use $\calS$ and $\calA$ to denote the state and action space, respectively. We assume  $\calS$ and $\calA$ are discrete, and rewards $R_t$ are uniformly bounded. The discrete state space assumption is imposed only to simplify the presentation and the theoretical analysis. Our proposed method is equally applicable to settings with continuous state space as well. 
The observed data consist of $N$ trajectories, corresponding to $N$ independent and identically distributed copies of $\{(S_{t},A_{t},R_{t})\}_{t\ge 0}$. For any $i=1,\cdots,N$, data collected from the $i$th trajectory can be summarized by $\{(S_{i,t},A_{i,t},R_{i,t},S_{i,t+1})\}_{0\le t< T_i}$, where $T_i$ denotes the termination time for the $i$-th trajectory. 

A policy defines the agent's way of choosing the action at each decision time. A history-dependent policy $\pi$ is a sequence of decision rules $\{\pi_t\}_{t\ge 0}$ such that each $\pi_t$ maps the observed data history $\overline{S}_{t}=S_{t} \cup \{S_{j},A_{j},R_{j}\}_{0\le j<t}$ to a probability mass function on $\mathcal{A}$, denoted by $\pi_t(\cdot|\overline{S}_{t})$. When each $\pi_t$ outputs a value in $\calA$, $\pi$ is referred to as a deterministic policy.  
Under $\pi$, the agent will set $A_{t}=a$ with probability $\pi_t(a|\overline{S}_{t})$ at the $t$th decision point. Suppose there exists some  function $\widetilde{\pi}$ such that $\pi_t(\cdot|\overline{S}_{t})=\widetilde{\pi}(\cdot|S_{t})$ 
almost surely for any $t$, then $\pi$ is referred to as a stationary policy. 

The primary goal of RL is to identify an optimal policy that maximizes the cumulative reward that the agent receives. To formally state this objective, we define the value function
\begin{eqnarray}\label{eqn:value}
	V^{\pi}(s)=\sum_{t=0}^{+\infty} \gamma^t \EE^{\pi} (R_{t}|S_{0}=s),
\end{eqnarray}
where $\EE^{\pi}$ denotes the expectation assuming the actions are selected according to $\pi$, and $0 \leq \gamma<1$ denotes some discounted factor that balances the tradeoff between immediate and future rewards. We aim to learn a policy that maximizes the following integrated value function,
\begin{eqnarray}\label{eqn:intvalue}
	\calV(\pi)=\sum_{s\in \calS} V^{\pi}(s) \nu(s),
\end{eqnarray}
where $\nu$ denotes some \textit{known} reference distribution function on $\calS$. The known of reference distribution is a typical assumption in RL literature. We assume $\nu(s)$ is 
uniformly bounded away from zero for any $s\in \calS$. 
When $N$ is large, one may alternatively set $\nu$ to the distribution function of $S_{0}$ and estimate it via the empirical distribution function of the data samples $\{S_{i,0}\}_{1\le i\le N}$. 

The following assumptions allow us to focus on stationary policies and serve as the foundations of the existing state-of-the-art RL algorithms:\\ 
\noindent (A1) {\textit {Markov assumption with stationary transitions}}: there exists a Markov transition function $p$ such that for any $t\ge 0$, $a\in \calA$ and $s,s'\in \calS$,
\vspace{-0.1cm}
\begin{eqnarray*}
	\prob(S_{t+1}=s'|A_{t}=a,S_{t}=s,\{S_{j},A_{j},R_{j}\}_{0\le j<t})=p(s'|a,s),
\end{eqnarray*}

\noindent (A2) {\textit {Conditional mean independence assumption with stationary reward functions}}: there exists some function $r$ such that
for any $t\ge 0$, $s\in \calS$ and $a\in \calA$,
\begin{eqnarray*}
	\EE (R_{t}|S_{t}=s,A_{t}=a,\{S_{j},A_{j},R_{j}\}_{0\le j<t})= \EE (R_{t}|S_{t}=s,A_{t}=a)=r(s,a),
\end{eqnarray*}

These two assumptions require the future state and the conditional mean of the immediate reward to be independent of the past observations given the current state-action pair at each decision time $t$. Under these assumptions, there exists an optimal stationary policy $\pi^{\tiny{opt}}$ whose value function $V^{\pi^{\tiny{opt}}}(s)$ is no worse than $V^{\pi}(s)$ for any history-dependent policy $\pi$ and any $s\in \calS$ \citep[Section 6.2]{puterman1994markov}. Consequently, it also maximizes the integrated value function $\calV(\pi)$.  (A1) and (A2) are testable from the observed data. See the goodness-of-fit test proposed by \cite{shi2020does}. In practice, to ensure the Markov property satisfied, we can construct the state by concatenating measurements over multiple decision points till the Markov property is satisfied (see Section \ref{sec:ohio} for details). To guarantee the transition probability $p$ and the reward function $r$ are time-homogeneous, we can include some auxiliary variables (e.g., time of the day) in the state. 
Hereafter, we restrict our attentions to the class of stationary policies that belong to certain pre-specified class $\Pi$ (e.g., linear or neural networks). 
For any $\pi\in \Pi$, we use $\pi(\bullet|s)$ to denote the probability mass function on $\mathcal{A}$ that the agent will follow when the state value is $s$. We aim to learn $\pi^*\in\argmax_{\pi\in \Pi} \calV(\pi)$ based on the observed data.  

To conclude this section, we impose one additional assumption that is commonly assumed in the literature to handle offline data \citep{sutton2018reinforcement}: 

\noindent (A3) 
The data are generated by some fixed stationary policy denoted by $b$. 
In addition, the stochastic process $\{(A_t, S_t)\}_{t\ge 0}$ is stationary.


Under (A1) and (A3), the process $\{(A_t, S_t)\}_{t\ge 0}$ forms a time-homogeneous Markov chain.  We use $p_{\infty}$ to denote the stationary distribution of the state-action pair. Suppose $p_{\infty}(a,s)$ is uniformly bounded away from zero for any $a\in \calA,s\in\calS$. 
The stationarity of $\{(A_t,S_t)\}_{t\ge 0}$ is assumed for convenience, since the Markov chain will eventually reach stationarity. For the ease of presentation, throughout this paper, we assume (A1)-(A3) hold. 
\subsection{Additional notations}\label{sec: notation}
For any $a\in \calA$, $s\in \calS$, we define the following action-value function associated with a given policy $\pi$ as $
	Q^{\pi}(a,s)=\sum_{t=0}^{+\infty} \gamma^t \EE^{\pi} (R_{t}|A_{0}=a,S_{0}=s),
$
better known as the $Q$-function. By definition, it is equal to the discounted cumulative reward the agent receives when the initial action-state pair equals $(a, s)$ and all subsequent actions follow $\pi$. By \eqref{eqn:value}, we have $V^{\pi}(s)=\sum_{a\in \calA} \pi(a|s) Q^{\pi}(a,s)$ for any $\pi$. 

The advantage function $A^{\pi}$ over $\calA \times \calS$ associated with $\pi$ is defined by the difference between the Q-function and the value function, i.e., $A^{\pi}(a,s)=Q^{\pi}(a,s)-V^{\pi}(s)$ for any $a\in\calA$ and $s\in\calS$. It represents the gain of the expected cumulative reward by performing the action $a$ rather than following $\pi$ at the initial stage. By definition, we obtain\vspace{-0.1cm}
\begin{eqnarray}\label{eqn:advanvalue}
	\sum_{a\in \calA} \pi(a|s)A^{\pi}(a,s)=0,\,\,\,\,\,\,\,\,\hbox{for~any}~s,\pi.
\end{eqnarray}

We next introduce the discounted visitation probability. For any $t\ge 0$, let $p_t^{\pi}(s'|a,s)$ denote the $t$-step visitation probability $\Pr^{\pi}(S_{t}=s'|A_{0}=a,S_{0}=s)$ assuming the actions are selected according to $\pi$ at time $1,\cdots,t-1$. When $t=0$, $p_t^{\pi}(s'|a,s)$ becomes the point mass function $\mathbb{I}(s'=s)$ where $\mathbb{I}(\cdot)$ denotes the indicator function. 
When $t=1$, $p_t^{\pi}$ equals the transition function $p$ defined in Condition (A1). We define the conditional discounted visitation probability function as $d^{\pi}(s'|a,s)=(1-\gamma)\sum_{t\ge 0} \gamma^{t} p_t^{\pi}(s'|a,s)$. Let\vspace{-0.1cm}
\begin{eqnarray*}
	d^{\pi,\nu}(s')=\sum_{s\in \calS}\sum_{a\in \calA} \pi(a|s) d^{\pi}(s'|a,s)\nu(s),
\end{eqnarray*}

\vspace{-0.1cm}
\noindent be the integrated discounted visitation probability function. Assume the actions are selected according to $\pi$ and the initial state follows the distribution $\nu$. Then $d^{\pi,\nu}$ corresponds to the probability mass function of a state $S^*$ that is a mixture of the random variables $\{S_t\}_{t\ge 0}$ with the corresponding mixture weights $\{(1-\gamma)\gamma^t\}_{t\ge 0}$. 

Finally, we introduce the discounted stationary probability ratio. 
For any $\pi$, define
\begin{eqnarray}\label{eqn:condw}
	\omega^{\pi}(a',s';a,s)=\frac{(1-\gamma)\{\mathbb{I}(s'=s,a'=a)+\sum_{t\ge 1} \gamma^{t}\pi(a'|s')p^{\pi}_t(s'|a,s) \}}{p_{\infty}(a',s')}.
\end{eqnarray}
By definition, the denominator in \eqref{eqn:condw} corresponds to the stationary distribution of $(A_t,S_t)$. When the system follows $\pi$, the numerator corresponds to the probability mass function of the state-action pair $(A^*,S^*)$ that is a mixture of $\{(A_t,S_t)\}_{t\ge 0}$ conditional on the event that $(A_0,S_0)=(a,s)$. We thus refer to $\omega^{\pi}$ as the conditional discounted stationary probability ratio. Similarly, define $\omega^{\pi,\nu}(a',s')=\sum_{a,s} \pi(a|s)\nu(s) \omega^{\pi}(a',s';a,s)$ as the integrated discounted stationary probability ratio. We remark that these probability ratios play an important role in constructing debiased value function estimators for off-policy evaluation \citep{kallus2019efficiently,pmlr-v139-shi21d}. 

\subsection{Trust region policy optimization}\label{sec:TRPO}
The following equation forms the basis of TRPO \citep[Lemma 6.1]{kakade2002approximately},\vspace{-0.1cm}
\begin{eqnarray}\label{eqn:key}
	(1-\gamma)\{\calV(\pi_{\scriptsize{\tsb{new}}})-\calV(\pi_{\scriptsize{\tsb{old}}})\}=\sum_{a\in \calA,s\in \calS} \pi_{\scriptsize{\tsb{new}}}(a|s)A^{\pi_{\scriptsize{\tsb{old}}}}(a,s) d^{\pi_{\scriptsize{\tsb{new}}},\nu}(s),
\end{eqnarray}
for any two policies $\pi_{\scriptsize{\tsb{old}}}$ and $\pi_{\scriptsize{\tsb{new}}}$, where $A^{\pi}$ and $d^{\pi ,\nu}$ are defined in Section \ref{sec: notation}. 
Based on Equation \eqref{eqn:key}, for any given initial policy $\pi_{\scriptsize{\tsb{old}}}\in \Pi$, it is tempting to directly search $\pi_{\scriptsize{\tsb{new}}}\in \Pi$ that maximizes an estimates of the right-hand-side (RHS) of \eqref{eqn:key}. However, such a direct search method is computationally challenging, due to the complex dependency of $d^{\pi_{\scriptsize{\tsb{new}}}, \nu}$ on $\pi_{\scriptsize{\tsb{new}}}$. 
In general, it is difficult to construct an estimator that has an explicit form of solution in terms of $\pi_{\scriptsize{\tsb{new}}}$ for  $d^{\pi_{\scriptsize{\tsb{new}}},\nu}$. 
The gradient of the corresponding estimator for \eqref{eqn:key} is extremely difficult to compute, and thus gradient-type methods are hard to apply. 

To make the computation feasible and efficient, \cite{schulman2015trust} considered approximating the RHS of \eqref{eqn:key} by\vspace{-0.1cm}
\begin{eqnarray}\label{eqn:eta1}
	\eta_1(\pi_{\scriptsize{\tsb{new}}},\pi_{\scriptsize{\tsb{old}}})=\sum_{a\in \calA,s\in \calS} \pi_{\scriptsize{\tsb{new}}}(a|s)A^{\pi_{\scriptsize{\tsb{old}}}}(a,s) d^{\pi_{\scriptsize{\tsb{old}}},\nu}(s).
\end{eqnarray}

\vspace{-0.1cm}
\noindent Note that the discounted visitation probability in \eqref{eqn:eta1} now depends on $\pi_{\scriptsize{\tsb{old}}}$, rather than $\pi_{\scriptsize{\tsb{new}}}$. The quantity $(1-\gamma)\calV(\pi_{\scriptsize{\tsb{old}}})+\eta_1(\pi_{\scriptsize{\tsb{new}}},\pi_{\scriptsize{\tsb{old}}})$ can be viewed as a first-order approximation of $(1-\gamma)\calV(\pi_{\scriptsize{\tsb{new}}})$. To elaborate this more, note that $\eta_1(\pi_{\scriptsize{\tsb{new}}},\pi_{\scriptsize{\tsb{old}}})$ can be rewritten as $
	\sum_{a\in \calA,s\in \calS} \{\pi_{\scriptsize{\tsb{new}}}(a|s)-\pi_{\scriptsize{\tsb{old}}}(a|s)\}A^{\pi_{\scriptsize{\tsb{old}}}}(a,s) d^{\pi_{\scriptsize{\tsb{old}}},\nu}(s),
$ by \eqref{eqn:advanvalue}. It then follows from \eqref{eqn:key}  that \vspace{-0.1cm}
\begin{eqnarray*}
	(1-\gamma)\calV(\pi_{\scriptsize{\tsb{new}}})&=&(1-\gamma)\calV(\pi_{\scriptsize{\tsb{old}}})+\underbrace{\sum_{a\in \calA,s\in \calS} \{\pi_{\scriptsize{\tsb{new}}}(a|s)-\pi_{\scriptsize{\tsb{old}}}(a|s)\}A^{\pi_{\scriptsize{\tsb{old}}}}(a,s) d^{\pi_{\scriptsize{\tsb{old}}},\nu}(s)}_{\eta_1(\pi_{\scriptsize{\tsb{new}}},\pi_{\scriptsize{\tsb{old}}})}\\
	&+&\underbrace{\sum_{a\in \calA,s\in \calS} \{\pi_{\scriptsize{\tsb{new}}}(a|s)-\pi_{\scriptsize{\tsb{old}}}(a|s)\}A^{\pi_{\scriptsize{\tsb{old}}}}(a,s) \{d^{\pi_{\scriptsize{\tsb{new}}},\nu}(s)-d^{\pi_{\scriptsize{\tsb{old}}},\nu}(s)\}}_{\eta_2(\pi_{\scriptsize{\tsb{new}}},\pi_{\scriptsize{\tsb{old}}})},
\end{eqnarray*}

\vspace{-0.1cm}
\noindent where $\eta_2(\pi_{\scriptsize{\tsb{new}}},\pi_{\scriptsize{\tsb{old}}})$ corresponds to a higher-order remainder term. To quantify this higher-order remainder term, for any two probability distributions $\mu_1,\mu_2$ on $\calA$, we use $\calD_{\TV}(\mu_1,\mu_2)$ to denote the total variation distance $2^{-1} \sum_{a\in \calA} |\mu_1(a)-\mu_2(a)|$. Let 
$$
\calD_{\KL}(\mu_1,\mu_2)=\sum_{a\in \calA} \mu_1(a)\log \{\mu_1(a)/\mu_2(a)\}
$$ 
denote the Kullback–Leibler (KL) divergence from $\mu_1$ to $\mu_2$. 
In Lemma \ref{prop1} (see Appendix \ref{sec:moreTRPO}), we show that
\begin{equation}\label{eqn:lowerbound}
\begin{aligned}
|\eta_2(\pi_{\scriptsize{\tsb{new}}},\pi_{\scriptsize{\tsb{old}}})| &\leq c^* \left[\EE_{S^*\sim d^{\pi_{\old},\nu}} \calD_{\TV}\left(\pi_{\scriptsize{\tsb{old}}}(\bullet|S^*),\pi_{\scriptsize{\tsb{new}}}(\bullet|S^*)\right)\right]^2\\
& \le c^* \EE_{S^*\sim d^{\pi_{\scriptsize{\tsb{old}}},\nu}}\calD_{\KL}\left(\pi_{\scriptsize{\tsb{old}}}(\bullet|S^*),\pi_{\scriptsize{\tsb{new}}}(\bullet|S^*)\right),
\end{aligned}
\end{equation}
for some positive constant $c^*>0$. The first inequality in \eqref{eqn:lowerbound} implies that $|\eta_2(\pi_{\scriptsize{\tsb{new}}},\pi_{\scriptsize{\tsb{old}}})|$ is indeed a second-order term. 
Based on this observation, we consider a policy optimization  procedure by maximizing a lower bound of \eqref{eqn:key}, given by\vspace{-0.1cm}
\begin{eqnarray}\label{eqn:opt}
	\pi_{\scriptsize{\tsb{new}}} \in \argmax_{\pi\in \Pi} \; [\eta_1(\pi,\pi_{\scriptsize{\tsb{old}}})-c^* \EE_{S^*\sim d^{\pi_{\old},\nu}}\calD_{\KL}\left(\pi_{\scriptsize{\tsb{old}}}(\bullet|S^*),\pi(\bullet|S^*)\right)].
\end{eqnarray}
%

\vspace{-0.1cm}
\noindent Iteratively solving the above optimization yields a type of minorization-maximization (MM) algorithm \citep{hunter2004tutorial} as we can see when $\pi = \pi_{\old}$, the objective function in \eqref{eqn:opt} becomes $0$ and $(1-\gamma)\calV(\pi)$ becomes $(1-\gamma)\calV(\pi_{\scriptsize{\tsb{old}}})$. So one can guarantee that \eqref{eqn:key} is always nonnegative after optimization.  Therefore this type of algorithm can greatly reduce the computational cost by circumventing computing $d^{\pi_{\new},\nu}$ and meanwhile monotonically improve the integrated value function. However, in practice, it may be hard to robustly choose the penalty coefficients $c^*$ in \eqref{eqn:opt}. To resolve this issue, one can consider iteratively solving the following equivalent optimization problem with a so-called trust region constraint:
\begin{eqnarray}\label{eqn:TRPO}
\begin{split}
&\pi_{\scriptsize{\tsb{new}}} \in \argmax_{\pi\in \Pi} \; \eta_1(\pi,\pi_{\scriptsize{\tsb{old}}})\\ 
&\textrm{subject to~~}\EE_{S^*\sim d^{\pi_{\scriptsize{\tsb{old}}},\nu}}\calD_{\KL}\left(\pi_{\scriptsize{\tsb{old}}}(\bullet|S^*),\pi(\bullet|S^*)\right)\le \delta,
\end{split}
\end{eqnarray}
for some constant $\delta>0$. 
This yields the TRPO algorithm. 

\section{Value Enhanced Policy Optimization}
In this section, we first present the motivation of our method. 
To implement TRPO, we need an estimate for $\eta_1(\pi,\pi_{\scriptsize{\tsb{old}}})$. 
In online settings, \cite{schulman2015trust} proposed to simulate trajectories following the policy $\pi_{\scriptsize{\tsb{old}}}$ to estimate $\eta_1(\pi,\pi_{\scriptsize{\tsb{old}}})$. In offline settings, it remains unknown how to effectively evaluate this quantity based on the observed data. By its definition, we note that $\eta_1$ depends on the nuisance functions $A^{\pi_{\old}}$ and $d^{\pi_{\old},\nu}$. A naive method is to first estimate these quantities (denote by $\widetilde{A}$ and $\widetilde{d}^{\nu}$) and then use the corresponding plug-in estimators $\widetilde{\eta}_1=\EE_{S^*\sim \widetilde{d}^{\nu}} \sum_{a\in \calA} \pi(a|S^*)\widetilde{A}(a,S^*)$ to estimate $\eta_1(\pi,\pi_{\old})$. However, such a procedure suffers from the following three main drawbacks: 
\begin{enumerate}[(I)]
	\item Iteratively computing the optimization problem \eqref{eqn:TRPO} can still be computationally expensive as the policy-dependent nuisance functions need to be updated at each iteration, especially when we do not have the closed-form expression for estimating these nuisance functions such as $\widetilde{d}^{\nu}$. Therefore it may not be desirable to directly implement TRPO method.
	\item When either $\widetilde{A}$ or $\widetilde{d}^{\nu}$ is not consistent, $\widetilde{\eta}_1$ might not be consistent. Consequently, there is no guarantee that the resulting new policy $\pi_{\new}$ can outperform $\pi_{\old}$. 
	
	\item To ensure both $\widetilde{A}$ and $\widetilde{d}^{\nu}$ are both consistent, one might consider estimating these functions nonparametrically. 
	Even when both of them are consistent, the plug-in estimator $\widetilde{\eta}_1$ might not be rate-optimal, i.e., $(NT)^{-1/2}$, due to that the nonparametric estimators $\widetilde{A}$ and $\widetilde{d}^{\nu}$ usually converge much slower than $(NT)^{-1/2}$. Consequently, compared with $\pi_{\old}$, the improvement by $\pi_{\new}$ may be marginal, resulting in a slow convergence rate to the optimal policy.
\end{enumerate}
To address the first concern, we propose to first apply some existing state-of-the-art offline RL method to obtain a good initial policy. Several methods can be applied here, including the conservative Q-learning \citep[CQL,][]{kumar2020conservative}, FQI, V-learning, among others. CQL and neural FQI use neural networks to index the policy class and V-learning considers a parametrized policy class indexed by a finite-dimensional vector. Note that these methods basically rely on the estimation of value or Q-functions. They are more computationally efficient than the iterative procedure described in (I). 
The value functions under initial policies obtained by these algorithms, if consistent, may converge at a slow rate as a trade-off for fast computation. In the second step, we propose to solve \eqref{eqn:TRPO} to improve their performances. This corresponds to a one-step update of the initial policy. One may also update this new policy for a few times to ensure the final estimated policy achieves a fast convergence rate. To remove the dependence between the initial policy and our policy optimization, we incorporate a data-splitting strategy, which is commonly seen in statistics and machine learning, e.g., \cite{chernozhukov2018double,kallus2019efficiently}. The detailed procedure is described in Section \ref{sec:overalg}. 

To address the second and the third concerns, we develop an efficient and robust estimating procedure for $\eta_1$, which is described in Section \ref{sec:EReta1}. Specifically, when the input policy $\pi_{\old}$ is consistent, we can guarantee that the output policy $\pi_{\new}$ by solving \eqref{eqn:TRPO} achieves the desired ``value enhancement" property. We call this set of methods ``value enhanced policy optimization (VEPO)". An overview of our algorithm is given in Section \ref{sec:overalg}, which integrates Q-learning, discounted stationary probability ratio estimation, transition dynamics estimation and policy search. We then discuss each component in the rest of the section. 

\subsection{An efficient and multiply robust estimator for $\eta_1$}\label{sec:EReta1}
For a given $\pi_{\old}\in \Pi$, we first outline three potential approaches (see (i)-(iii) below) to estimating $\eta_1(\pi,\pi_{\old})$ from the observed data. Each of these methods requires some nuisance functions to be consistently estimated. We then present our proposal that combines these three methods to achieve efficient and triply robust estimation.

\noindent \textbf{(i) Plug-in estimator:} $\widetilde{\eta}_1^{(1)}=\EE_{S^*\sim \widetilde{d}^{\nu}} \sum_{a\in \calA} \pi(a|S^*)\widetilde{A}(a,S^*)$. This is the plug-in method discussed earlier. The validity of $\widetilde{\eta}_1^{(1)}$ requires the consistent estimation of $d^{\pi_{\old},\nu}$ and $A^{\pi_{\old}}$.
	
\noindent \textbf{(ii) Importance sampling (IS) estimator I:} \vspace{-0.1cm} 
\begin{eqnarray}\label{eqn:IS1}
	\widetilde{\eta}_1^{(2)}=\frac{1}{\sum_i T_i} \sum_{i=1}^N \sum_{t=0}^{T_i-1} \EE_{S^*\sim \widetilde{d}^{\nu}} \sum_{a \in \calA}  \{\pi(a|S^*)-\pi_{\old}(a|S^*)\} \widetilde{\omega}(A_{i,t},S_{i,t};a,S^*) R_{i,t},
\end{eqnarray}

\vspace{-0.1cm}
\noindent where $\widetilde{\omega}$ denotes some estimator for the conditional discounted probability ratio $\omega^{\pi_{\old}}$. See \eqref{eqn:condw} for a detailed definition. The validity of $\widetilde{\eta}_1^{(2)}$ requires consistent estimation of both $d^{\pi_{\old},\nu}$ and $\omega^{\pi_{\old}}$. Such an IS estimator is motivated by the work of \cite{liu2018breaking} on the off-policy value evaluation. A key observation is that, under (A2) and (A3), $Q^{\pi}(a,s)$ can be represented by\vspace{-0.1cm}
\begin{eqnarray*}
	\sum_{a',s'} \{\mathbb{I}(s'=s,a'=a)+\sum_{t\ge 1} \gamma^{t}\pi(a'|s')p_{t}^{\pi}(s'|a,s) \} r(s',a')=\frac{1}{1-\gamma}\EE {\omega}^{\pi}(A_t,S_t;a,s) R_t,
\end{eqnarray*}

\vspace{-0.1cm}
\noindent for any $t$, $s$ and $a$. This yields the following IS estimator for $A^{\pi}(a,s)$: \vspace{-0.1cm}
\begin{eqnarray*}
\frac{1}{\sum_i T_i} \sum_{i=1}^N \sum_{t=0}^{T_i-1} \sum_{a' \in \calA}  \{\mathbb{I}(a'=a)-\pi(a'|s)\} \widetilde{\omega}(A_{i,t},S_{i,t};a',s) R_{i,t}.
\end{eqnarray*}

\vspace{-0.1cm}
\noindent Plugging in the above estimator for $A^{\pi_{\old}}(a,s)$ and $\widetilde{d}^{\nu}$ for $d^{\pi_{\old},\nu}$ yields \eqref{eqn:IS1}.  
	
\noindent \textbf{(iii) IS estimator II:} \vspace{-0.1cm} 
\begin{eqnarray}\label{eqn:IS2}
\widetilde{\eta}_1^{(3)}=\frac{1}{\sum_i T_i} \sum_{i=1}^N \sum_{t=0}^{T_i-1} \sum_{a\in \calA}\pi(a|S_{i,t})\widetilde{A}(a,S_{i,t})\widetilde{\omega}^{\nu}(A_{i,t},S_{i,t}),
\end{eqnarray}

\vspace{-0.1cm}
\noindent where $\widetilde{\omega}^{\nu}$ denotes some estimator for the integrated probability ratio $\omega^{\pi_{\old},\nu}$. The validity of $\widetilde{\eta}_1^{(3)}$ requires consistent estimation of $d^{\pi_{\old},\nu}$ and the integrated probability ratio $\omega^{\pi_{\old},\nu}$. To motivate this estimator, we observe that the expectation $\EE_{S\sim d^{\pi,\nu}} f(S)$ can be rewritten as $\EE f(S_t) \omega^{\pi,\nu}(A_t,S_t)$, for any function $f$, policy $\pi$ and decision point $t$. Consequently, we can represent $\eta_1$ by \vspace{-0.1cm}
\begin{eqnarray*}
	\EE \sum_{a\in \calA} \pi(a|S_t) A^{\pi_{\old}}(a,S_t) \omega^{\pi_{\old},\nu}(A_t,S_t).
\end{eqnarray*}
This yields the IS estimator in \eqref{eqn:IS2}. 

We note that each of the above estimator may be severely biased when the corresponding estimated nuisance functions fail to be consistent. Toward that end, we develop a multiply robust estimator by carefully combining the estimating strategies used in (i)-(iii). Meanwhile, the resulting estimator requires much weaker assumptions to achieve consistency. Let $o$ be a shorthand for a data tuple $(s,a,r,s')$. The key to constructing our estimator is the following estimating function, 
\begin{eqnarray*}
	\psi(o;\pi,\pi_{\old},\widetilde{V},\widetilde{A},\widetilde{\omega},\widetilde{d})=\psi_1(\pi,\pi_{\old},\widetilde{A},\widetilde{d})+ \psi_2(o;\pi,\pi_{\old},\widetilde{V},\widetilde{A},\widetilde{\omega},\widetilde{d})+\psi_3(o;\pi,\pi_{\old},\widetilde{A},\widetilde{\omega},\widetilde{d}),
\end{eqnarray*}
for some given nuisance functions $\widetilde{V}$, $\widetilde{A}$, $\widetilde{\omega}$ and $\widetilde{d}$, where,
\begin{eqnarray*}
	\psi_1(\pi,\pi_{\old},\widetilde{A},\widetilde{d})= \EE_{S^*\sim \widetilde{d}^{\nu}} \sum_{a\in \calA} \pi(a|S^*)\widetilde{A}(a,S^*),\\
	\psi_2(o;\pi,\pi_{\old},\widetilde{V},\widetilde{A},\widetilde{\omega},\widetilde{d})=\frac{1}{1-\gamma} \EE_{S^*\sim \widetilde{d}^{\nu}}\sum_{a^*\in \calA}\{\pi(a^*|S^*)-\pi_{\old}(a^*|S^*)\}
	\widetilde{\omega}(a,s;a^*,S^*)\\
	\times \{r+\gamma \widetilde{V}(s')-\widetilde{V}(s)-\widetilde{A}(a,s)\}\\
	\psi_3(o;\pi,\pi_{\old},\widetilde{A},\widetilde{\omega},\widetilde{d})=\sum_{a^*\in \calA} \frac{ \widetilde{\omega}^{\nu}(a,s)}{1-\gamma}\left[\gamma\EE_{\substack{a'\sim \pi_{\old}(\bullet|s')\\S^*\sim \widetilde{d}(\bullet|a',s')}}\widetilde{A}(a^*,S^*)\pi(a^*|S^*)\right.\\
	-\EE_{S^*\sim \widetilde{d}(\bullet|a,s)}\widetilde{A}(a^*,S^*)\pi(a^*|S^*)+\left.(1-\gamma)\pi(a^*|s)\widetilde{A}(a^*,s)\right],
\end{eqnarray*} 
where the nuisance functions $\widetilde{d}^{\nu}$ and $\widetilde{\omega}^{\nu}$ are determined by $\widetilde{d}$ and $\widetilde{\omega}$, given by $\widetilde{d}^{\nu}(\bullet)=\sum_{a,s}\pi_{\old}(a|s)\nu(s)\widetilde{d}(\bullet|a,s)$ and $\widetilde{\omega}^{\nu}(\bullet,\bullet)=\sum_{a,s}\pi_{\old}(a|s)\nu(s) \widetilde{\omega}(\bullet,\bullet;a,s)$. 

By definition, $\psi$ consists of three terms. The first term $\psi_1$ is essentially the plug-in estimator that depends only on $\widetilde{A}$ and $\widetilde{d}$. The second and third terms, i.e., $\psi_2$ and $\psi_3$, are the augmentation terms. Let $O_t=(S_t,A_t,R_t,S_{t+1})$ for any $t$, we have $\EE \psi_2(O_t;\pi,\pi_{\old},\widetilde{V},\widetilde{A},\widetilde{\omega},\widetilde{d})=0$ when $\widetilde{A}=A^{\pi_{\old}}$, $\widetilde{V}=V^{\pi_{\old}}$ and $\EE \psi_3(O_t;\pi,\pi_{\old},\widetilde{A},\widetilde{\omega},\widetilde{d})=0$ when $\widetilde{d}=d^{\pi_{\old}}$. See Appendix \ref{sec:moreprop2} 
for details. 
The purpose of adding these two terms is to offer an additional protection against the potential bias of $\psi_1$ resulting from the biases of $\widetilde{A}$ and $\widetilde{d}$. Therefore we have the following proposition. 
\begin{prop}\label{prop2}
	Suppose $\sum_{a} \pi_{\old}(a|s)\widetilde{A}(a,s)=0$ for any $s$. Then 
	$ \psi(O_t;\pi,\pi_{\old},\widetilde{V},\widetilde{A},\widetilde{\omega},\widetilde{d})$ is unbiased to $\eta_1$ as long as one of the following three assumptions are satisfied: (B1) $\widetilde{A}=A^{\pi_{\old}}$, $\widetilde{V}=V^{\pi_{\old}}$ and $\widetilde{d}=d^{\pi_{\old}}$; (B2) $\widetilde{\omega}=\omega^{\pi_{\old}}$ and $\widetilde{d}=d^{\pi_{\old}}$; (B3) $\widetilde{A}=A^{\pi_{\old}}$ and $\widetilde{\omega}=\omega^{\pi_{\old}}$. 
\end{prop} 
The condition $\sum_{a} \pi_{\old}(a|s)\widetilde{A}(a,s)$ is automatically satisfied if we set $\widetilde{A}(a,s)=\widetilde{A}^*(a,s)-\sum_{a} \pi_{\old}(a|s)\widetilde{A}^*(a,s)=0$ for any initial advantage estimator $\widetilde{A}^*$. We remark that if $Q^{\pi_{\old}}$ is correctly specified, so do $A^{\pi_{\old}}$ and $V^{\pi_{\old}}$. Based on this estimating function, a triply-robust estimator for $\eta_1$ is given by \vspace{-0.1cm}
\begin{eqnarray}\label{eqn:naturalest}
	\frac{1}{\sum_i T_i}\sum_{i=1}^N \sum_{t=0}^{T_i-1} \psi(O_{i,t};\pi,\pi_{\old},\widetilde{V},\widetilde{A},\widetilde{\omega},\widetilde{d}),
\end{eqnarray}

\vspace{-0.1cm}
\noindent where $O_{i,t}=(S_{i,t},A_{i,t},R_{i,t},S_{i,t+1})$. It remains to specify the estimation of nuisance functions. We present the details in the next section. In Section \ref{sec:theory1}, we show the resulting estimator is efficient. 

\subsection{The complete algorithm}\label{sec:overalg}
Our main idea is to construct an efficient and robust estimator for $\eta_1$ to improve the performance of an initial policy $\pi_{\old}$. To achieve this goal, we need to estimate four key nuisance functions: (a) An initial policy $\pi_{\old}$; (b) The value and advantage function $V^{\pi_{\old}}$ and $A^{\pi_{\old}}$; (c) The conditional discounted stationary probability ratio $\omega^{\pi_{\old}}$; (d) The conditional discounted visitation probability function $d^{\pi_{\old}}$. 

Correspondingly, our estimating procedure involves four key steps, described in Sections \ref{sec:oldlearn}-\ref{sec:gans} respectively. In addition to these four main estimating components, we also propose to couple the estimator in \eqref{eqn:naturalest} with a data-splitting and cross-fitting strategy. Specifically, without loss of generality, we randomly divide the indices of all trajectories $\{1,2,\cdots,N\}$ into $\mathbb{L}$ subsets $\cup_{\ell=1}^{\mathbb{L}}\{O_{i,t}\}_{i\in \mathcal{I}_{\ell},0\le t< T_i}$ with equal size, where $\mathcal{I}_{\ell}$ denotes the indices of trajectories contained in the $\ell$th data subset. We next apply the learning components in (a)-(d) to the data subsets in $\mathcal{I}_{\ell}^c=\{1,\cdots,N\} \backslash \mathcal{I}_{\ell}$ for $\ell=1,\cdots,\mathbb{L}$ and construct the estimator $\widehat{\eta}_1$ via cross-fitting. \change{Cross-fitting essentially guarantees that the dataset used to learn (a)-(d) is independent of the dataset used to construct $\widehat{\eta}_1$. This allows us to avoid imposing Donsker-typed conditions, which limit the growth rate of the VC dimension of the estimators for (a)-(d) \citep{chernozhukov2018double}, to achieve desirable properties of our procedure.} 
Then we propose to search $\pi_{\new}$ that maximizes $\widehat{\eta}_1$ subject to the trust region constraint in \eqref{eqn:TRPO} with $d^{\pi_{\old},\nu}$ replaced by its corresponding estimator. This corresponds to a one-step update of the initial policy. After computing $\pi_{\new}$, we can repeat the above procedures a few times to guarantee the final estimated policy achieves a fast convergence rates. See Section \ref{sec:ve} for details. A pseudocode summarizing our approach is given in Algorithm \ref{alg:full}.

We remark that it is not necessary to develop a robust and efficient estimating procedure for the integrated KL divergence in the trust region constraint, due to the fact that it corresponds to a higher-order remainder term for the value difference (see Lemma \ref{prop1} in Appendix \ref{sec:moreTRPO}). Our proposal works as long as $d^{\pi_{\old},\nu}$ and its estimator have the common support. This condition is automatically satisfied when the reference distribution $\nu$ is uniformly bounded away from zero on $\calS$.


\begin{algorithm}[t!]
	\caption{Value enhanced policy optimization}
	\label{alg:full}
	\begin{algorithmic}
		\item\normalsize
		\hspace{-0.7cm} {\bf{Input}}: ~A policy class $\Pi$ and the observed data.\\
		\hspace{-0.7cm} {\bf{Output}}: An updated policy $\pi_{\new}$.
		\begin{enumerate}[Step 1.]
			\item[Step 0.] Randomly split the trajectories into $\mathbb{L}$ disjoint subsets, $\cup_{\ell=1}^{\mathbb{L}}\mathcal{I}_{\ell}$. Let $\mathcal{I}_{\ell}^c=\{1,\cdots,N\}-\mathcal{I}_{\ell}$, for $\ell=1,\cdots,\mathbb{L}$.
			
			\item  For $\ell=1,\cdots,\mathbb{L}$: apply some existing state of art offline RL algorithm to obtain the input initial policy $\pi_{\old}$ using the data subset $\calI_{\ell}^c$. Denote the resulting policy as $\pi^{(\ell)}_{\old}$.

			
			\item  For $\ell=1,\cdots,\mathbb{L}$: 
			\begin{enumerate}[({2}a)]
			\item Apply fitted-Q evaluation (see \eqref{eqn:updateQ}) to estimate the Q-function $Q^{\pi^{(\ell)}_{\old}}$, based on the data subset in $\mathcal{I}_{\ell}^c$. Denote the resulting estimator by $\widehat{Q}^{(\ell)}$.
			
			\item Set $\widehat{V}^{(\ell)}(s)=\sum_{a} \pi^{(\ell)}_{\old}(a|s)\widehat{Q}^{(\ell)}(a,s)$ for any $s$ and $\widehat{A}^{(\ell)}(a,s)=\widehat{Q}^{(\ell)}-\widehat{V}^{(\ell)}(s)$ for any $a$ and $s$.
			\end{enumerate} 
			
			\item For $\ell=1,2,\cdots,\mathbb{L}$, apply the method detailed in Section \ref{sec:ratio} to learn a conditional probability ratio $\widehat{\omega}^{(\ell)}$, based on the data subset $\mathcal{I}_{\ell}^c$. 
			
			\item For $\ell=1,2,\cdots,\mathbb{L}$, apply 
   machine learning methods to approximate the conditional distribution of $S_{t+1}$ given $A_{t}$ and $S_t$, using the data subset $\mathcal{I}_{\ell}^c$, which can be used to generate the pseudo sample. The pseudo sample can then be used to approximate the distribution function $d^{\pi_{\old}^{(\ell)}}$ and $d^{\pi_{\old}^{(\ell)},\nu}$ (see Algorithms \ref{alg0} and \ref{alg0.5}). 
			
			\item Construct the estimator for $\eta_1$ and update the corresponding policy:
			\begin{enumerate}[({5}a)]
				\item Apply cross-fitting to construct the value difference estimator $\widehat{\eta}_1$ (see \eqref{eqn:esteta1}). 
				
				\item Use the estimated dynamic to construct the trust region constraint (see \eqref{eqn:constraint}).
				
				\item Search $\pi_{\new}\in \Pi$ that maximizes $\widehat{\eta}_1$ subject to \eqref{eqn:constraint}.
			\end{enumerate}
			
			\item Set all $\pi^{(\ell)}_{\old}$ to $\pi_{\new}$ for $\ell = 1, \cdots, \mathbb{L}$ and repeat Steps 2-5 a few times. 
		\end{enumerate}
	\end{algorithmic}
\end{algorithm}


\subsubsection{Step 1: Initial Policy Optimization}\label{sec:oldlearn}
First, to initial our VEPO algorithm, we need to estimate an initial policy denoted by $\pi_{\old}$. We propose to apply some existing state-of-the-art offline RL algorithm on the data subset $\calI^c_{\ell}$ and obtain resulting estimated policy denoted by $\hat \pi_{\old}^{(\ell)}$ for $\ell = 1, \cdots, \mathbb{L}$. For the ease of presentation, we often write $\hat \pi_{\old}^{(\ell)}$ as $\pi_{\old}^{(\ell)}$ when there is no confusion. Specifically, in our numerical studies, we implement three offline RL algorithms to obtain our initial policies. The first one is FQI using the idea of value iteration with function approximation \citep{sutton2018reinforcement}. It relies on the optimal Bellman equation \citep{bertsekas1996neuro}. The second one is V-learning proposed by \cite{luckett2019estimating}, which considered policy iteration with function approximation. The last one is CQL by \cite{kumar2020conservative}, which proposed to learn a lower bound of $Q$-function during the policy iteration procedure. The last method is driven by the overestimation of the value function due to the distributional mismatch between the behavior policy in the batch dataset and the learned policy. We remark that any valid offline RL method can be employed here, 
as long as assumptions in Theorem \ref{thm4} are satisfied. See details in Section \ref{sec:theory2}.

\subsubsection{Step 2: Q-learning}\label{sec:Qlearn}
Second, to estimate nuisance functions in (b), we employ a Q-learning type algorithm to learn the Q-function $Q^{\pi_{\old}^{(\ell)}}$, based on the data subset in $\mathcal{I}_{\ell}^c$. 
Denote the corresponding estimator by $\widehat{Q}^{(\ell)}$. We then construct the corresponding estimators for the value and advantage function by $\widehat{V}^{(\ell)}(s)=\sum_{a} \pi^{(\ell)}_{\old}(a|s)\widehat{Q}^{(\ell)}(a,s)$ and $\widehat{A}^{(\ell)}(a,s)=\widehat{Q}^{(\ell)}(a,s)-\widehat{V}^{(\ell)}(s)$, for any $s$ and $a$, based on the relation that $V^{\pi_{\old}^{(\ell)}}(s)=\sum_{a} \pi_{\old}^{(\ell)}(a|s) Q^{\pi_{\old}^{(\ell)}}(a,s)$, $A^{\pi_{\old}^{(\ell)}}(a,s)=Q^{\pi_{\old}^{(\ell)}}(a,s)-V^{\pi_{\old}^{(\ell)}}(s)$. Consequently, the requirement $\sum_{a} \pi^{(\ell)}_{\old}(a|s)\widehat{A}^{(\ell)}(a,s)=0$ for the estimated advantage function is automatically satisfied. 

Several algorithms can be used here to estimate the Q-function $Q^{\pi_{\old}^{(\ell)}}$.  
Here, we adopt the fitted Q-evaluation evaluation (FQE) method proposed by \cite{le2019batch}. The following Bellman's equation forms the basis of all Q-learning type algorithms: for $t \geq 0$, \vspace{-0.1cm}
\begin{eqnarray}\label{eqn:belllman}
	Q^{\pi_{\old}^{(\ell)}}(A_t,S_t)=\EE \left\{\left.R_t+\gamma \sum_a \pi_{\old}^{(\ell)}(a|S_{t+1}) Q^{\pi_{\old}^{(\ell)}}(a,S_{t+1})\right|A_t,S_t\right\}.
\end{eqnarray}
Based on this identity, we  estimate $Q^{\pi_{\old}^{(\ell)}}$ by iteratively computing \vspace{-0.1cm}
\begin{eqnarray}\label{eqn:updateQ}
	\widehat{Q}_k^{(\ell)}=\argmin_{Q_k} \sum_{i,t} \left\{R_{i,t}+\sum_{a}\pi^{(\ell)}_{\old}(a|S_{i, t+1}) \widehat{Q}_{k-1}^{(\ell)}(a,S_{i,t+1})-Q_k(A_{i,t},S_{i,t})\right\}^2,
\end{eqnarray} 

\vspace{-0.1cm}
\noindent for $k=1,2,\cdots$ with any initial $	\widehat{Q}_0^{(\ell)}$. Several supervised learning methods can be incorporated here, since \eqref{eqn:updateQ} is essentially a regression problem. In our implementation, we employ deep learning \citep{lecun2015deep} to compute $\widehat{Q}_k^{(\ell)}$ during each iteration. 

\subsubsection{Step 3: discounted stationary probability ratio estimation}\label{sec:ratio}
\change{We adopt the algorithm developed by \cite{pmlr-v139-shi21d} to estimate (c). The procedure is motivated by the following observation: 
	For any two pairs $(i,t)$ and $(i',t')$ such that $O_{i,t}$ and $O_{i',t'}$ are independent, we have for any function $f$ such that $\EE \Delta(\omega^{\pi},f,\pi;i,t,i',t')=0$, where 
 \vspace{-0.1cm}
	\begin{eqnarray}\label{eqn:omega}
	\begin{split}
	 \Delta(\omega^{\pi},f,\pi;i,t,i',t')=\omega^{\pi}(S_{i',t'},A_{i',t'};S_{i,t},A_{i,t})\Big\{\gamma\sum_a \pi(a|S_{i',t'+1}) f(S_{i',t'+1},a;S_{i,t},A_{i,t})\\-f(S_{i',t'},A_{i',t'};S_{i,t},A_{i,t})\Big\}+(1-\gamma)  f(S_{i,t},A_{i,t},S_{i,t},A_{i,t}).
	\end{split}
	\end{eqnarray}
	Conversely, for any $\omega$ that satisfies $\EE\Delta(\omega,f,\pi;i,t,i',t')=0$ for any $f$, we have $\omega=\omega^{\pi}$. See Equation (5) of \cite{pmlr-v139-shi21d} for details. }
 
For each function $f$, an estimating equation for $\omega^{\pi_{\old}^{(\ell)}}$ can be constructed based on  \eqref{eqn:omega}. Similar to the proposal in \cite{liu2018breaking}, $f$ can be treated as a discriminator to construct the following minimax loss
function \vspace{-0.1cm}
\begin{eqnarray}\label{eqn:obj}
	\argmin_{\omega\in\Omega} \sup_{f\in \mathcal{F}} \left|\EE \Delta(\omega^{\pi_{\old}^{(\ell)}},f,\pi_{\old}^{(\ell)};i,t,i',t')\right|^2,
\end{eqnarray}
for some function classes $\Omega$ and $\mathcal{F}$. 
To simplify the calculation, $\mathcal{F}$ is set to a unit ball of a reproducing kernel Hilbert space. This yields a close-form expression for the inner maximization problem in \eqref{eqn:obj}. The expectation in \eqref{eqn:obj} is then approximated by the empirical distributions of the data subset in $\mathcal{I}_{\ell}^c$. The parameters involved in $\omega$ are updated by the stochastic gradient descent algorithm. To save space, we present the details in Section \ref{sec:cdspr}.

\subsubsection{Step 4: Estimation of the underlying dynamics}\label{sec:gans}
The conditional discounted visitation probability $d^{\pi_{\old}^{(\ell)}}$ in (d) is extremely difficult to estimate when the state space is high-dimensional, as it corresponds to a mixture distribution of state variables at different decision points. A key observation is that, $d^{\pi_{\old}^{(\ell)}}$ is completely determined by the system dynamics. As long as the transition kernel $p$ can be consistently estimated, $d^{\pi_{\old}^{(\ell)}}$ can be well-approximated.

To compute $d^{\pi_{\old}^{(\ell)}}$, in continuous state space, we propose to use a Gaussian probabilistic model to estimate the transition kernel $p$ with the use of 
machine learning models to approximate the mean and covariance matrix. \change{In particular, we assume the next state $S_{t+1}$ given the current action-state $(S_t, A_t)$ follows a multi-variate Gaussian distribution $\calN(\mu(S_t, A_t), \Sigma(S_t, A_t))$, where $\mu$ and $\Sigma$ are the corresponding mean and covariance matrix functions respectively. Notice that estimating $\mu$ is essentially a regression problem, and there are many supervised learning methods available. In our implementation, we use deep neural networks to compute $\widehat{\mu}^{(\ell)}$, via
\begin{eqnarray*}
    \widehat{\mu}^{(\ell)}_j=\arg\min_{\mu_j} \sum_{i\in \mathcal{I}^c_{\ell}}\sum_{t=0}^{T_i-1}[S_{i,t+1,j}-\mu_j(S_{i,t}, A_{i,t})]^2,
\end{eqnarray*}
where $\widehat{\mu}^{(\ell)}_j$ and $S_{i,t+1,j}$ denote the $j$th element of $\widehat{\mu}^{(\ell)}$ and $S_{i,t+1}$, respectively. Next, let $\varepsilon_{i,t,j}$ denote the residual $S_{i,t+1,j}-\widehat{\mu}_j^{(\ell)}(S_{i,t}, A_{i,t})$. We employ deep learning again to compute $\widehat{\Sigma}^{(\ell)}$, via
\begin{eqnarray*}
    \widehat{\Sigma}^{(\ell)}_{j_1,j_2}=\arg\min_{\Sigma_{j_1,j_2}} \sum_{i\in \mathcal{I}_{\ell}^c}\sum_{t=0}^{T_i-1}[\varepsilon_{i,t,j_1}\varepsilon_{i,t,j_2}-\Sigma_{j_1,j_2}(S_{i,t}, A_{i,t})]^2,
\end{eqnarray*}
where $\widehat{\Sigma}^{(\ell)}_{j_1,j_2}$ denotes the $(j_1,j_2)$th entry of $\widehat \Sigma^{(\ell)}$, which is the final estimator of $\Sigma^{(\ell)}$. }

\begin{algorithm}[t!]
	\caption{Generate pseudo samples to approximate $d^{\pi_{\old}^{(\ell)}}(\bullet | s, a)$}
	\label{alg0}
	\begin{algorithmic}
		\item
		\begin{description}
			\item[\textbf{Input}:] Estimators $(\widehat \mu^{(\ell)}, \widehat \Sigma^{(\ell)})$ computed via supervised learning.
			
			\item[\textbf{for}]  $m=1$ to $M$: \textbf{do}
			\begin{enumerate}
				\item[(a)] Set $\widetilde{S}_{0}^{(m)}=s$ and $\widetilde{A}_{0}^{(m)}=a$;
				\item[(b)] For $t'=1$ to $T'$, generate  $\widetilde{S}_{t'}^{(m)}$ by  $\calN(\widehat{\mu}^{(\ell)}(\widetilde{A}_{t'-1}^{(m)},\widetilde{S}_{t'-1}^{(m)}), \widehat{\Sigma}^{(\ell)}(\widetilde{A}_{t'-1}^{(m)},\widetilde{S}_{t'-1}^{(m)}))$ and randomly sample $\widetilde{A}_{t'}^{(m)}$ from $\pi_{\old}^{(\ell)}(\bullet|\widetilde{S}_{t'}^{(m)})$.
			\end{enumerate}
			\item[\textbf{Output}] $\{\widetilde{S}_{t'}^{(m)}\}_{1\le m\le M, 0\le t\le T'}$. 
		\end{description}
	\end{algorithmic}
\end{algorithm}
\begin{algorithm}[t!]
	\caption{Generate pseudo samples to approximate $d^{\pi_{\old}^{(\ell)},\nu}$}
	\label{alg0.5}
	\begin{algorithmic}
		\item
		\begin{description}
			\item[\textbf{Input}:] Estimators $(\widehat \mu^{(\ell)}, \widehat \Sigma^{(\ell)})$ computed via supervised learning.
			
			\item[\textbf{for}]  $m=1$ to $M$: \textbf{do}
			\begin{enumerate}
				\item[(a)] Sample $\widetilde{S}_{0}^{(m),\nu}$ from $\nu$;\vspace{-0.2cm}
				\item[(b)] For $t'=0$ to $T'-1$,  randomly sample $\widetilde{A}_{t'}^{(m),\nu}$ from $\pi_{\old}^{(\ell)}(\bullet|\widetilde{S}_{t'}^{(m),\nu})$ and generate generate  $\widetilde{S}_{t'+1}^{(m),\nu}$ by  $\calN(\widehat{\mu}^{(\ell)}(\widetilde{A}_{t'}^{(m)},\widetilde{S}_{t'}^{(m),\nu}), \widehat{\Sigma}^{(\ell)}(\widetilde{A}_{t'}^{(m)},\widetilde{S}_{t'}^{(m),\nu}))$.
			\end{enumerate}	
			\item[\textbf{Output}] $\{\widetilde{S}_{t'}^{(m),\nu}\}_{1\le m\le M, 0\le t\le T'}$. 
		\end{description}
	\end{algorithmic}
\end{algorithm}

\change{To approximate the conditional distribution $d^{\pi_{\old}^{(\ell)}}(\bullet|a,s)$ for any $a$ and $s$, we can employ the Monte Carlo method and generate a sequence of pseudo samples $\{\widetilde{S}_{t'}^{(m)}\}_{1\le m\le M, 0\le t\le T'}$ based on $\widehat{\mu}^{(\ell)}$ and $\widehat{\Sigma}^{(\ell)}$. We illustrate the details in Algorithm \ref{alg0}.
For any function $f$, the integral 
$\EE_{S^*\sim d^{\pi_{\old}^{(\ell)}}(\bullet|a,s)}f(S^*)$ can then be approximated by \vspace{-0.1cm}
\begin{eqnarray}\label{eqn:app}
(1-\gamma) M^{-1} \sum_{m=1}^M \sum_{t'=0}^{T'} \gamma^t f(\widetilde{S}_{t'}^{(m)}),
\end{eqnarray}
which we denote by $\EE_{S^*\sim \widehat{d}^{(\ell)}(\bullet|a,s)}f(S^*)$ where $\widehat{d}^{(\ell)}$ denotes the estimated probability density/mass function. We use the aggregated squared total variation distance 
\begin{eqnarray}\label{sqtv}
\EE_{(S,A)\sim p_{\infty}}\mathcal{D}_{\TV}^2(\widehat{d}^{(\ell)}(\bullet|A,S),d^{\pi_{\old}^{(\ell)}}(\bullet|A,S))
\end{eqnarray}
to measure its goodness of fit. See Condition (C3) in Section \ref{sec:theory} for details. In Section \ref{sec:morecond} of the supplementary material, we show that when the conditional Gaussian model is correctly specified, the minimum eigenvalue of $\Sigma(a,s)$ is uniformly bounded away from zero for any $a,s$, and $\widehat{\Sigma}(a,s)$ is positive definite for any $a,s$, \eqref{sqtv} is upper bounded by
\begin{eqnarray*}
    3\gamma^{2T'}+\frac{3}{M}+O(1) \EE_{(A^*,S^*)\sim p_{\infty}} [\|\mu(S^*,A^*)-\widehat{\mu}^{(\ell)}(S^*,A^*)\|_2+\|\Sigma(S^*,A^*)-\widehat{\Sigma}^{(\ell)}(S^*,A^*)\|_F]^2,
\end{eqnarray*}
where $O(1)$ denotes some positive constant. As such $\widehat{d}^{(\ell)}$ is consistent as long as $T',M\to \infty$ and that the estimated conditional mean and covariance functions are consistent. We also remark that the conditional Gaussian model is widely used in the RL literature for learning the transition function \citep[see e.g.,][]{yu2020mopo}. Alternatively, a conditional Gaussian mixture model can be employed to mitigate model misspecification \citep{mdn}. }

\subsubsection{Step 5: policy optimization}\label{sec:ve}
After obtaining the four key components, we next discuss the procedure to compute the new policy $\pi_{\new}$. We propose to construct the estimator $\widehat{\eta}_1$ via cross-fitting. Specifically, we estimate it by \vspace{-0.1cm}
\begin{eqnarray}\label{eqn:esteta1}
	\widehat{\eta}_1(\pi)=\frac{1}{\sum_i T_i}\sum_{\ell=1}^{\mathbb{L}} \left\{\sum_{i\in \mathcal{I}_{\ell}} \sum_{t=0}^{T_i-1} \psi(O_{i,t};\pi, \pi^{(\ell)}_{\old},\widehat{V}^{(\ell)},\widehat{A}^{(\ell)},\widehat{\omega}^{(\ell)},\widehat{d}^{(\ell)})\right\}.
\end{eqnarray}

\vspace{-0.1cm}
\noindent Note that the nuisance functions $\pi_{\old}^{(\ell)}$, $\widehat{A}^{(\ell)}$, $\widehat{V}^{(\ell)}$, $\widehat{\omega}^{(\ell)}$ and $\widehat{d}^{(\ell)}$ are computed based on the data subset in $\mathcal{I}_{\ell}^c$ and are independent of the observations in $\mathcal{I}_{\ell}$ that are used to construct the estimating function $\psi$. 
Theoretical properties of this estimator are studied in Section \ref{sec:theory1}.

We then propose to learn $\pi_{\new}$ by solving the following constrained optimization, 
\begin{eqnarray}\label{eqn:constraint}
\begin{split}
\pi_{\new} \in &\argmax_{\pi\in\Pi}  \widehat{\eta}_1(\pi),\\
&\textrm{subject to~~}\frac{1}{\mathbb{L}}\sum_{\ell=1}^{\mathbb{L}}\EE_{S^*\sim \widehat{d}^{(\ell),\nu}}\calD_{\KL}\left(\pi^{(\ell)}_{\scriptsize{\tsb{old}}}(\bullet|S^*),\pi(\bullet|S^*)\right)\le \delta,
\end{split}
\end{eqnarray}

\vspace{-0.1cm}
\noindent where $\widehat{d}^{(\ell),\nu}$ denotes the distribution of the pseudo samples $\{\widetilde{S}_{t'}^{(m),\nu}\}_{1\le m\le M, 0\le t\le T'}$ generated according to Algorithm 3.

We remark that when the initial policy $\pi_{\old}^{(\ell)}$ is set to the behavior policy $b$, the constraint in \eqref{eqn:constraint} then requires the learned policy close to the behavior one in the batch dataset, which is commonly used in the recent developed RL algorithms, e.g., \cite{wu2019behavior}. As pointed out by \cite{levine2020offline}, one of the fundamental challenges of offline RL is the out of distribution due to the mismatch between behavior policy and the target policy. This out of distribution issue will result in an overestimation for the value function, therefore deteriorating the performance of policy learning. Restricting the learned policies to stay close to the behavior one can potentially relieve this limitation. 
In practice, we can repeat the constraint optimization in \eqref{eqn:constraint} several times by setting the all initial policy $\pi^{(\ell)}_{\old}$ for $\ell = 1, \cdots, \mathbb{L}$ to  $\pi_{\new}$ obtained from the previous iteration. This guarantees that the final estimated policy achieves a fast convergence rate. Finally, we remark that our method is not overly complicated compared to the existing state-of-the-art RL algorithms and indeed quite flexible. Although we require to learn a number of components and use deep learning models in our numerical experiments below, these components can be alternatively estimated via much simpler methods (e.g., parametric models, sieve methods or kernels). In this case, our proposed algorithm becomes more accessible.


\section{Theory}\label{sec:theory}
In this section, we systematically study the theoretical properties of our algorithm. In Section \ref{sec:theory2}, we establish the properties of our estimated optimal policy. In Section \ref{sec:theory1}, we show the proposed first-order value difference estimator $\widehat{\eta}_1$ is efficient. To simplify the theoretical analysis, we assume $T_1=\cdots=T_N=T$. All the asymptotic results are derived when either the number of trajectories $N$, or the number of decision points $T$, diverges to infinity. Results of this type provide useful theoretical guarantees for a variety of applications in reinforcement learning. We refer to theories of this type as bidirectional theories. 
We also allow the state-action space, the transition matrix $p$, the reward function $r$ and policy class $\Pi$ to depend on $N$ and $T$. Consequently, the optimal policy, the Q function and the discounted visitation probability are allowed to vary with $N$ or $T$.

\subsection{Properties of the estimated optimal policy}\label{sec:theory2}
We first show in Theorem \ref{thm3} that the value difference between the new and the old policy is $O(\sqrt{\delta})$ where $\delta$ corresponds to the threshold in the trust region constraint \eqref{eqn:constraint}. Consequently, by setting $\delta\to 0$, we can guarantee that the new policy is asymptotically no worse than the old one on average. 
\begin{theorem}\label{thm3}
	$|\calV(\pi_{\new})-\mathbb{L}^{-1}\sum_{\ell=1}^{\mathbb{L}}\calV(\pi_{\old}^{(\ell)})|\le O(1)\sqrt{\delta}$ where $O(1)$ denotes some positive constant.  
\end{theorem}

We remark that Theorem \ref{thm3} does not require any conditions on the estimated nuisance functions $\widehat{Q}^{(\ell)}$, $\widehat{\omega}^{(\ell)}$ and $\widehat{d}^{(\ell)}$. Nor does it require $\pi_{\old}^{(\ell)}$ to converge to an optimal policy $\pi^{\tiny{opt}}$. In addition, Theorem 1 holds deterministically even if the policy $\pi_{\old}^{(\ell)}$ is data-dependent. See Appendix A.1 for more details.

Note that $N\times T$ corresponds to the total number of decision points. We next consider the scenario where the input policy $\pi_{\old}^{(\ell)}$ is close to $\pi^{\tiny{opt}}$ in the sense that $\calV(\pi_{\old}^{(\ell)})=\calV(\pi^{\tiny{opt}})+O\{(NT)^{-\kappa_0}\}$ for some constant $\kappa_0>0$. This implies that $\pi_{\old}^{(\ell)}$ is consistent to $\pi^{\tiny{opt}}$ as either $N$ or $T$ diverges to infinity. 
We further assume
$\calV(\pi^*)=\calV(\pi^{\tiny{opt}})+o(1)$, as $NT\to \infty$. Recall that $\pi^*$ is defined as the optimal in-class policy that maximizes the value among $\Pi$. In other words, the value under the optimal in-class policy approaches to the optimal value function as the sample size increases (because the size of policy class also increases). To simplify the theoretical analysis, we assume $\pi^{\tiny{opt}}\in \Pi$ such that $\pi^*=\pi^{\tiny{opt}}$. Meanwhile, our theories are equally applied to settings where $\pi^{\tiny{opt}}\notin \Pi$ but the value difference $\calV(\pi^*)-\calV(\pi^{\tiny{opt}})$ converges at a sufficiently fast rate. This assumption is reasonable in practice when we either have domain knowledge on the parametric form of $\pi^{\tiny{opt}}$ or use function classes with the universal approximation capabilities (e.g., neural networks) to parametrize $\Pi$. To establish the value enhancement property, we need the following set of conditions.

\noindent (C1) Suppose $\EE_{(A,S)\sim p_{\infty}} |\widehat{Q}^{(\ell)}(A,S)-Q^{ \pi^{(\ell)}_{\old}}(A,S)|^2=O_p\{(NT)^{-2\kappa_1}\}$ for some constant $\kappa_1\ge 0$. In addition, $\widehat{Q}^{(\ell)}$ is uniformly bounded almost surely. 

\noindent (C2) Suppose $ \EE_{(A,S),(\widetilde{A}, \widetilde{S})\sim p_{\infty}} |\widehat{\omega}^{(\ell)}(\widetilde{A},\widetilde{S}; A,S)-\omega^{ \pi_{\old}^{(\ell)}}(\widetilde{A},\widetilde{S}; A,S)|=O_p\{(NT)^{-2\kappa_2}\}$ for some constant $\kappa_2\ge 0$, where $(\widetilde{A},\widetilde{S})$ and $(A,S)$ denote two independent state-action pairs generated according to  $p_{\infty}$. In addition, $\widehat{\omega}^{(\ell)}$ is uniformly bounded almost surely.\\
\noindent \change{(C3) Suppose $\EE_{(A,S)\sim p_{\infty}} |\calD_{\TV}(d^{\pi_{\old}^{(\ell)}}(\bullet|A,S),\widehat{d}^{(\ell)}(\bullet|A,S))|^2=O_p\{(NT)^{-2\kappa_3}\}$ for some constant $\kappa_3\ge 0$}. 

\noindent (C4) Suppose  $\Pi$ corresponds to certain VC type function class \citep{chernozhukov2014gaussian} with VC indices upper bounded by $O\{(NT)^{\kappa_4}\}$ for some constant $0\le \kappa_4 <\frac{\alpha}{\alpha+1}$, where $\alpha$ is defined below.

\noindent (C5) The optimal policy is unique. In addition, there exist some positive constants $\alpha,\bar{c},\bar{\epsilon}$ such that $\prob(-\epsilon \le A^{\pi^{\tiny{opt}}}(a,S^*)<0)\le \bar{c} \epsilon^{\alpha}$ for any $a\in\calA$ and $0<\epsilon\le \bar{\epsilon}$, where the random variable $S^*$ is distributed according to $d^{\pi^{\tiny{opt}},\nu}$.

\noindent (C6) The process $\{(S_t,A_t,R_t)\}_{t\ge 0}$ is exponentially $\beta$-mixing. 

Conditions (C1)-(C3) characterize the theoretical requirements on the learners in (a)-(c), respectively.
In particular, (C1)-(C2) require the squared prediction losses of the estimated Q-function and the conditional probability ratio to satisfy certain convergence rates, whereas Condition (C3) assumes the squared total variation norm between the transition function and its estimator to satisfy a certain convergence rate. If some parametric models are imposed to learn $Q^{\pi_{\old}}$, $\omega^{\pi_{\old}}$ and the transition matrix $p$, we have $\kappa_1=\kappa_2=\kappa_3=1/2$. In our setup, we only require $\kappa_{i_1}+\kappa_{i_2}>1/(2+2\alpha)$ for any disjoint $i_1,i_2\in \{1,2,3\}$. See the statement of Theorem \ref{thm4} below. This condition holds when $\min_{i\in \{1,2,3\}}\kappa_i>1/(4+4\alpha)$ and thus is  achievable for many nonparametric estimators. It is also strictly weaker than those imposed in the recent literature that require the nuisance function to converge at a rate faster than $(NT)^{-1/4}$ for off-policy value evaluation \citep[e.g.,][]{kallus2019efficiently}. 
\change{For example, when the kernel smoother \citep{feng2020accountable}, sieve method \citep{shi2020statistical,chen2022well} or deep neural networks \citep{fan2020theoretical} are used to approximate the Q-function, it can be shown that under some technical conditions, (C2) holds with $\kappa_1=\beta_1/(2\beta_1+d_S)$ and $\beta_1 > d_S / 2$ where $d_S$ denotes the dimension of the state space and $\beta_1$ denotes the H{\"o}lder exponent that characterizes the smoothness of the Q-function.} 
Similar result (i.e., optimal non-parametric convergence rate) can be obtained for the conditional probability ratio function. 
\change{As discussed in Section \ref{sec:gans}, when $T'$ and $M$ are sufficiently large, (C3) essentially requires the estimated mean and covariance functions in the conditional Gaussian model to converge at a rate of $(NT)^{-\kappa_3}$. Under some regularity conditions, an optimal non-parametric convergence rate can also be achieved.}

Condition (C4) is mild as the policy class $\Pi$ is pre-specified. When a linear policy class is employed, we have $\kappa_4=\#s$ where $\#s$ denotes the number of parameters used to index the policy class. When $\Pi$ is set to some deep neural networks, the corresponding VC-dimension is also available in the literature \citep[see e.g.,][]{harvey2017nearly}. 

The uniqueness of the optimal policy (C5) is commonly assumed in the literature \citep{ertefaie2018constructing, luckett2019estimating}. The second part of (C5) is closely related to margin-type conditions commonly used to bound the excess misclassification error \citep{tsybakov2004optimal,audibert2007fast} and the regret of individualized treatment regimes in point treatment studies \citep{qian2011performance,luedtke2016statistical,shi2020breaking}. 

To better understand the margin condition in (C5), we first observe that $A^{\pi^{\tiny{opt}}}(a,s)\le 0$ for any $a$ and $s$. To elaborate this, we note that $\pi^{\tiny{opt}}$ maximizes $V^{\pi}(s)$ for any $\pi$. Consider the following history-dependent policy $\pi^{\tiny{opt}}(a)$ that assigns $a$ at the initial decision point and follows $\pi^{\tiny{opt}}$ in the subsequent steps. The value under such a policy is given by $Q^{\pi^{\tiny{opt}}}(a,s)$. It follows that $Q^{\pi^{\tiny{opt}}}(a,s)\le V^{\pi^{\tiny{opt}}}(s)$. or equivalently, $A^{\pi^{\tiny{opt}}}(a,s)\le 0$ for any $a$ and $s$. The equality holds only when $a=\argmax_{a'} Q^{\pi^{\tiny{opt}}}(a',s)$. The argmax is well-defined by the uniqueness of the optimal policy. For $a\neq \argmax_{a'} Q^{\pi^{\tiny{opt}}}(a',s)$, the advantage function corresponds to the value difference between $\pi^{\tiny{opt}}(a)$ and $\pi^{\tiny{opt}}$. The smaller the difference, the harder it is to identify the optimal policy. To ensure $\pi^{\tiny{opt}}$ can be consistently identified, it is thus reasonable to assume $\prob(0<|A^{\tiny{opt}}(a,S^*)|\le \epsilon)$ decays to zero with $\epsilon$ as well. (C5) explicitly characterizes such dependence. \textcolor{black}{ For example, when the action space is binary, let $\tau(s)$ denote the contrast function, i.e., $\tau(s) = Q^{\pi^{\tiny{opt}}}(1,s)-Q^{\pi^{\tiny{opt}}}(0,s)$. It is immediate to see that $A^{\pi^{\tiny{opt}}}(0,s)=\min(-\tau(s),0)$ and $A^{\pi^{\tiny{opt}}}(1,s)=\min(\tau(s),0)$. Thus, the second part of (C5) essentially requires $\prob(0<|\tau(S^*)|\le \epsilon)\le \bar{c}\epsilon^{\alpha}$, which is automatically satisfied with $\alpha=1$ when $\tau(S^*)$ has a bounded probability density function. More generally, it holds when $|\tau(S^*)|^{\alpha}$ has a bounded probability density function. For instance, suppose both the initial reference distribution $\nu$ and the Markov transition function have bounded density functions on $(0,+\infty)$. Then, the distribution of $S^*$, i.e., the mixture distribution of $\{S_t\}_{t\ge 0}$ with weights $\{(1-\gamma)\gamma^t\}_{t\ge 0}$ has a bounded probability density function as well. Suppose $\tau(S^*)=(S^*)^{1/\alpha}$. Then it is immediate to see that $|\tau(S^*)|^{\alpha}$ has a bounded probability density function. Finally, when $|\tau(S^*)|$ is uniformly bounded away from zero, then (C5) holds with $\alpha=+\infty$.} We will see in Theorem \ref{thm4} below that the convergence rate of $\pi_{\new}$ depends crucially on the margin parameter $\alpha$. 

Assumption (C6) characterizes the dependence of the data observations over time. It essentially requires the $\beta$-mixing coefficient \citep[see e.g.,][for a detailed definition]{bradley2005basic} of at lag $q$, which measures the time dependence between the set of variables $\{(S_j,A_j,R_j)\}_{j\le t}$ and $\{(S_j,A_j,R_j)\}_{j\ge t+q}$, to decay to zero at an exponential rate with respect to $q$.  This assumption automatically holds when $\{(S_t,A_t,R_t)\}_{t\ge 0}$ forms a geometrically ergodic Markov chain. Geometric ergodicity is less restrictive than those  imposed in the existing reinforcement learning literature that requires observations to be  independent \citep[see e.g.,][]{degris2012off,farahmand2016regularized} or to follow a uniform-ergodic Markov chain \citep[see e.g.,][]{bhandari2018}. 


\begin{theorem}[Value Enhancement Property]\label{thm4}
	Suppose (C1)-(C6) hold. 
	If the constants $\kappa_1,\kappa_2,\kappa_3$ satisfy $\kappa_{i_1}+\kappa_{i_2}>1/(2+2\alpha)$ for any disjoint $i_1,i_2\in \{1,2,3\}$, and that $\calV(\pi^{\tiny{opt}})-\calV(\pi_{\old}^{(\ell)})=O_p\{(NT)^{-\kappa_0}\}$ for any $\ell$, we have $\calV(\pi^{\tiny{opt}})-\calV(\pi_{\new})=E_1+E_2$ where $E_1=O_p\{(NT)^{-\frac{\kappa_0(2\alpha+1)}{\alpha+1}}\}$, $E_2=o_p\{(NT)^{-1/2}\}$.
\end{theorem}

Theorem \ref{thm4} states that the value difference $\calV(\pi^{\tiny{opt}})-\calV(\pi_{\new})$ can be decomposed into two terms. Here, the first term $E_1$ describes how the input policy $\pi_{\old}$ takes effect. It is due to the presence of the higher-order remainder term $\eta_2(\pi,\pi_{\old})$ resulting from the first order approximation of the value difference $\calV(\pi)-\calV(\pi_{\old})$. 
The second term $E_2$ is due to the estimation error of the $\eta_1(\pi,\pi_{\old})$. \textcolor{black}{In the typical multiply robust setting, it often requires that $\kappa_{i_1}+\kappa_{i_2}>1/2$ so that the bias of estimating $\eta_1(\pi,\pi_{\old})$ is $o_p\{(NT)^{-1/2}\}$. In Theorem \ref{thm4}, since we require slower rates for nuisance parameters, the proposed value difference estimator for $\eta_1(\pi,\pi_{\textrm{old}})$ may converge slower than the $1/2$-root. However, the value enhancement property can still be established under such a slower rate requirement. This is due to that Theorem 2 is concerned with the convergence rate of the estimated optimal policy $\pi_{\textrm{new}}$ in terms of the value instead of the rate of the proposed value difference estimator (denoted by $\widehat{\eta}_1(\pi_{\textrm{new}})$). In particular, the convergence rate of $\pi_{\textrm{new}}$ in terms of the value is primarily determined by the difference between $\widehat{\eta}_1(\pi_{\textrm{new}})$ and $\widehat{\eta}_1(\pi^{\tiny{opt}})$ (see Page 12 of the supplementary material), which converges at a faster rate than $\widehat{\eta}_1(\pi_{\textrm{new}})$ itself. This is because $\pi_{\textrm{new}}$ is consistent to the optimal policy implied by the condition that $\calV(\pi^{\tiny{opt}})-\calV(\pi_{\old}^{(\ell)})=O_p\{(NT)^{-\kappa_0}\}$.}

When $\kappa_0\le 1/2$, it can be seen that the value under the output policy converges at a faster rate than the input policy, leading to the desired ``value enhancement property". One can repeat the one-step update multiple times to guarantee that the value of the estimated optimal policy converges at a rate of $o_p\{(NT)^{-1/2}\}$. When the initial policy already converges faster than the parametric rate (e.g., $\kappa_0> 1/2$), then our proposal is not guaranteed to yield a better policy in theory. However, as shown in our empirical studies (see Section \ref{sec:emp}), the values of the proposed policies are often larger than those computed via state-of-the-art RL algorithms. This suggests that although these initial policies are consistent, they might converge at a suboptimal rate and have room for improvement. 

\subsection{Efficiency of the value difference estimator}\label{sec:theory1}
In this subsection, we show that conditional on $\pi_{\old}^{(\ell)}$, the proposed estimator for $\eta_1(\pi,\pi_{\old}^{(\ell)})$, i.e.,
\begin{eqnarray*}
    \widehat{\eta}_1(\pi,\pi_{\old}^{(\ell)})=\frac{1}{T|\mathcal{I}_{\ell}|} \left\{\sum_{i\in \mathcal{I}_{\ell}} \sum_{t=0}^{T-1} \psi(O_{i,t};\pi, \pi^{(\ell)}_{\old},\widehat{V}^{(\ell)},\widehat{A}^{(\ell)},\widehat{\omega}^{(\ell)},\widehat{d}^{(\ell)})\right\}.
\end{eqnarray*}
is nearly unbiased to $\eta_1(\pi,\pi_{\old}^{(\ell)})$ and its asymptotic variance  matches this efficiency bound. The notion of efficiency bound can be found in Section A.4 of Supplementary Material. Consequently, $\widehat{\eta}_1$ is efficient.

Let $\bar{p}$ denote the conditional distribution of $(R_t,S_{t+1})$ given $(A_t,S_t)$. 
For any given $\pi$ and $\pi_{\old}$, we note that $\eta_1(\pi,\pi_{\old})$ is completely determined by the transition function $\bar{p}$. Let $\{\bar{p}_{\theta_1}:\theta_1\in \Theta_1\}$ be a regular parametric submodel for $\bar{p}$. This requires $\bar{p}_{\theta_1}$ to be a transition matrix for any $\theta_1$ and $\bar{p}=\bar{p}_{\theta_1^*}$ for some $\theta_1^*\in \Theta_1$. Similarly, let $\{\bar{b}_{\theta_2}:\theta_2\in\Theta_2\}$ and $\{\bar{\nu}_{\theta_3}:\theta_3\in \Theta_3\}$ be regular parametric submodels for the behavior policy and the initial state distribution, respectively. Let $\theta=(\theta_1,\theta_2,\theta_3)$ and $\theta^*=(\theta_1^*,\theta_2^*,\theta_3^*)$ where $\theta_2^*$ and $\theta_3^*$ correspond to the true parameters in $\Theta_2$ and $\Theta_3$. Under a given submodel indexed by $\theta$, the log-likelihood function of a single data trajectory can be written as \vspace{-0.1cm}
\begin{eqnarray*}
	\ell_{T}(\{O_{t}\}_{t};\theta)=\log \left[\bar{\nu}_{\theta_3}(S_{0}) \prod_{t=0}^{T} \{\bar{p}_{\theta_1}(R_{t}, S_{t+1}|A_{t},S_{t}) \bar{b}_{\theta_2}(A_{t}|S_{t})\}\right].
\end{eqnarray*}

\vspace{-0.1cm}
\noindent Note that $\eta_1$ can be defined as function of $\theta$ as well. We define the efficiency bound as \vspace{-0.1cm}
\begin{eqnarray*}
	\textrm{EB}(|\mathcal{I}_{\ell}|,T)=|\mathcal{I}_{\ell}|\sup \nabla_{\theta} \eta_1(\theta^*)  \left\{\EE \nabla_{\theta} \ell_{T}(\{O_{t}\}_{t};\theta^*) \nabla_{\theta}^\top  \ell_{T}(\{O_{t}\}_{t};\theta^*) \right\}^{-1}  \nabla_{\theta}^\top \eta_1(\theta^*),
\end{eqnarray*}

\vspace{-0.1cm}
\noindent where the supremum is taken over all regular parametric submodels, and $\nabla_{\theta} \textrm{g}(\theta')$ denotes the derivative of a function $g$ with respect to $\theta$, evaluated at $\theta=\theta'$. As discussed before, $\eta_1$ depends on $\theta$ only through $\theta_1$. 
\begin{theorem}\label{thm5}
	Suppose the conditions in Theorem \ref{thm4} holds with $\kappa_{i_1}+\kappa_{i_2}>1/2$ for any disjoint $i_1,i_2\in \{1,2,3\}$. Then conditional on $\pi_{\old}^{(\ell)}$, we have for any $\pi$ that \vspace{-0.1cm}
	\begin{eqnarray*}
		\frac{\widehat{\eta}_1(\pi,\pi_{\old}^{(\ell)})-\eta_1(\pi, \pi_{\old}^{(\ell)})}{\sqrt{\tsb{\textrm{EB}}(|\mathcal{I}_{\ell}|,T)}}  \stackrel{d}{\to} N(0,1).
	\end{eqnarray*}
\end{theorem}

\vspace{-0.1cm}
\noindent Theorem \ref{thm5} implies that $\widehat{\eta}_1$ is asymptotically unbiased with asymptotic variance $\textrm{EB}(|\mathcal{I}_{\ell}|,T)$. This demonstrates the efficiency of the proposed estimator. 

\section{Numerical examples}\label{sec:emp}
In this section, we use one toy example and real data related studies 
to demonstrate the superior performance of our method. Specifically, in Section \ref{sec:toy}, we use a toy example to demonstrate the multiple robustness of our estimator and the value enhancement property. We then demonstrate the performance of the proposed method on OhioT1DM related datasets in Section \ref{sec:ohio}. In Appendix D, we conduct another simulation study to illustrate the finite-sample performance of our algorithm compared with several existing methods. 

\subsection{A Toy Example}\label{sec:toy}
We design a toy example to illustrate the multiple robustness of our estimator and the desired value enhancement property. Consider a binary state space $\calS = \{0, 1\}$, where $S_0$ takes value $0$ with probability $0.4$ and otherwise. The action space $\calA$ takes values in $ \{0,1\} $. 
The reward function is defined as $r(a, s) = \mathbb{I}(s = a)$,
and then the reward is generated according to $R_t=r(A_t,S_t)+e_t$ where $\{e_t\}_{0 \leq t < T}$ is a sequence of i.i.d. $N(0,2)$ random errors. The transition matrix of $p(S' | A, S)$ and behavior policy can be found in Section D of the Supplementary Material. 

In this tabular case, the oracle values of $Q^{\pi}$, $\omega^{\pi}$ and $p$ can be simulated using Monte Carlo methods. To demonstrate the triply robustness property, we will add some random errors on $Q^{\pi}$, $\omega^{\pi}$ or $p$ to make them biased. Then we compute $\eta_1$ with these  nuisance functions and obtain the resulting estimated optimal policy via our proposed algorithm. Specifically, we consider the following five combinations of nuisance function estimators: (i) ``origin": all the nuisance functions are set to their oracle values. 
	(ii) ``mod1": $\omega^\pi$ is set to a biased value, while other nuisances are oracle;
		(iii) ``mod2": $Q^\pi$ is set to a biased value, while other nuisances are oracle.
		(iv) ``mod3": The transition $p$ is set to a biased value, while other nuisances are oracle.
		(vi) ``mod4": all nuisance functions are set to bias values. Details of these scenarios can be found in Section D of the Supplementary Material

To summarize, Scenario (i) corresponds to the oracle setting where all the nuisance functions are correctly specified. In Scenarios (ii)-(iv), one of the nuisance functions is misspecified. In the last scenario, all the nuisance functions are misspecified. 
We also vary the initial policy to investigate the value enhancement property. In particular, we represent the initial policy $\pi_{\old}$ using a $2\times 2$ matrix and consider the following parametrization,
\begin{equation*}
	\pi_{\text{old}} = 	\bordermatrix{~&A = 0 &A = 1\cr
		S = 0&\kappa&1-\kappa\cr
		S = 1&1-\kappa&\kappa\cr
	},
\end{equation*}
for some $ \kappa\in[0,1]$. According to our data generating mechanism, $\kappa = 0$ corresponds to the optimal policy. The closer $ \kappa $ is to 1, the worse the initial policy is. 
We consider three choices of $\kappa$, corresponding to $0.2, 0.5$ and $0.8$. This yields 
three different initial policies. 
We further consider two choices of sample size, $N = 30, T=30$ and $ N = 50, T=50$. This yields a total of $5\times 3\times 2=30$ settings. 
$\gamma$ is set to $0.9$. It can be shown that the optimal value $\calV(\pi^{\text{opt}})$ equals $(1-\gamma)^{-1} = 10$. Finally, we consider three choices of $\delta$ (see Equation \eqref{eqn:constraint}), corresponding to $0.05$, $0.1$ and $0.2$, respectively. 


Results are reported in Figures \ref{fig:toy0.1}, \ref{fig:toy0.05} and \ref{fig:toy0.2} (see Appendix \ref{sec:addnum} in the supplementary article). All values of estimated policies are computed via Monte Carlo simulations. It can be observed that 
in Scenarios (i)-(iv), 
our proposed algorithm using models in (i)-(iv) substantially improves the performance of the initial policy, demonstrating the desired value enhancement property. In particular, when either one of nuisance functions models is misspecified, the proposed method remains valid. This empirically verifies the triply-robustness property. 

In addition, when all models are misspecified, the proposed method is not guaranteed to improve the value. Specifically, in the first two columns, the proposed method under ``mod4" improves the initial policy after a few iterations. In the last column, however, values of the estimated policies are smaller than the initial one. We suspect that this is because under model misspecification, our procedure may converge to a suboptimal policy whose value is bounded between the value of the initial policy in the second column and that in the last column. Consequently, when the existing policy given by other methods is already close to the optimal, using inconsistent estimators of nuisance functions could possibly degrade the performance. 
Finally, under settings where the initial policy is very different from the optimal one (i.e., the first columns of Figures \ref{fig:toy0.1},  \ref{fig:toy0.05} and \ref{fig:toy0.2} in Appendix D), it requires more iterations and a larger $\delta$ for our method to achieve a larger value. In contrast, when the initial policy is close to the optimal one, fewer iterations are needed and a smaller $\delta$ would be preferred. For instance, it can be seen from the third column of Figure \ref{fig:toy0.2} in Appendix D that when $\delta=0.2$, the values of the estimated policies using models in (ii) and (iii) decrease at the third iteration. 

Finally, to further demonstrate the advantage of the proposed method, we use lookup tables (e.g., linear models with table lookup features) instead of deep learning models to parametrize all nuisance functions (including the Q-function, the probability ratio and the transition kernel), and apply the proposed method to this toy example. Results are reported in Figure S3 of the Supplementary Material. It can be seen that the proposed method is still able to improve the performance of initial policies.




\begin{figure}[!t]
	\centering
	\includegraphics[width=0.6\linewidth,trim=150 50 150 50]{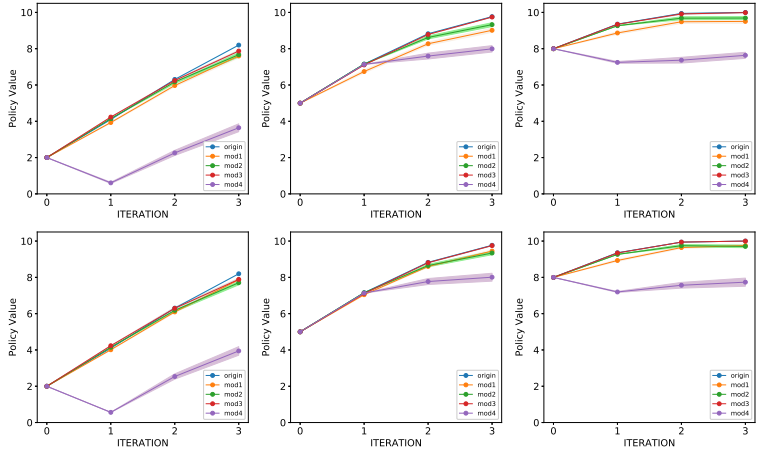}
	\caption{
		Values of estimated policies in a toy example. 
		First row represents results using  $ (T,N) $ pair as $ (30,30) $ while the second row using $ (50,50) $. The three columns represents initial policy factor $ \kappa $ taking values $ 0.8,0.5,0.2 $ respectively. The horizontal axis represents the number of iterations used in our value enhancement procedure. When iteration equals zero, we plot the evaluation value for the initial policy. 
		The optimal value is $10$ and $\delta$ is fixed to 0.1.  The confidence band is computed based on 100 replications. }
	\label{fig:toy0.1}
\end{figure}

\subsection{Application to the OhioT1DM Related Datasets}\label{sec:ohio}
There is an increasing interest in applying RL algorithms to mobile health(mHealth) applications. In this section, we conducted two analyses based on the OhioT1DM dataset. In the first analysis, we generate synthetic data to mimic this real dataset and apply our method to the synthetic dataset. In particular, we use the simulation environment designed in Section 5.2.2 of \cite{shi2020does} for the data generation. In the second analysis, we apply our method to the real dataset.

The OhioT1DM data set contains continuous measurements for six patents with type 1 diabetes over eight weeks. The state $\tilde S_t$ consists of three states, 
corresponding to the average blood glucose levels, the carbohydrate estimate for the meal and the exercise intensity, respectively. The action $A$ is the 
amount of insulin doses. 
We discretize the action space and consider five actions, i.e., $\calA = \{0, 1, 2, 3, 4\}$ from no to high doses of insulin. 
The Markov test developed by \cite{shi2020does} suggests that the data are likely to satisfy a $4$th order Markov property, so we reconstruct the state variable $ S_t = (\tilde{S}_{t-3},A_{t-3},\tilde{S}_{t-2},A_{t-2},\tilde{S}_{t-1},A_{t-1},\tilde{S}_t) $ 
by concatenating past measurements to meet the Markov assumption. This yields to a 15-dimensional state vector. The reward $R_t$ is defined as the Index of Glycemic Control that is a deterministic function of the average blood glucose levels during the time interval $[t, t+1)$. 

\begin{figure}[!t]
	\centering
	\includegraphics[width=0.6\linewidth,trim=100 50 100 50]{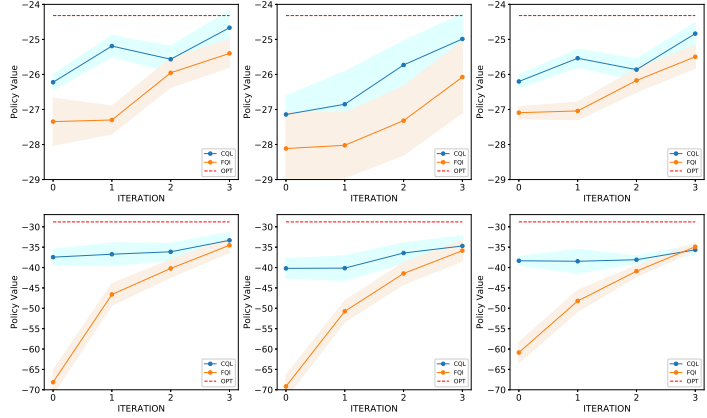}
	\caption{Values of various policies in the real data based simulation study. The initial policies are computed by  CQL and FQI. The first row represents results using $ \gamma=0.9 $ while the second row using $ \gamma=0.95$. The optimal values are equal to $-24.32$ and $-28.79$ respectively (shown by the read dash line). Three columns represents using $ (T,N) $ pair as $ (50,100),(25,200),(100,50) $ respectively. The confidence band is computed based on 
    100 replications.}
	\label{fig:ohio}
\end{figure}

We first apply the proposed method to the synthetic datasets. We use FQI and CQL to compute the initial policy. We did not implement V-learning (VL) here since it requires large computational costs when the dataset is large. 
In Figure \ref{fig:ohio}, it can be seen that we are able to achieve near-optimal policies after iterating the proposed algorithm 3 times. The estimated optimal policy achieves larger values than the initial policies in all cases. The improvement is substantial when $\pi_{\old}$ is not very close to the optimal policy. 

\begin{figure}[h]
	\centering
	\includegraphics[width=0.6\linewidth]{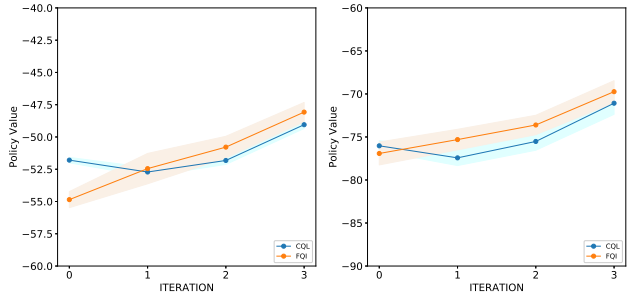}
	\caption{Values of various policy computed based on the real dataset. The initial policies are computed by CQL and FQI. Left figure corresponds to the result of $ \gamma=0.9 $ while the right one considers $ \gamma=0.95 $. The policy values are computed by cross-validation procedure and the confidence band is computed based on 100 replications.}
	\label{fig:realdata}
\end{figure}

We next apply our method to the real dataset. In order to evaluate the estimated optimal policy, we split the data into training and test datasets. After obtaining estimated optimal policies on the training data, we apply FQE on the test data to compute the policy values of all these estimated policies.  Figure \ref{fig:realdata} reports these values. It implies that the proposed algorithm will yield a policy with larger value after 2 to 3 iterations. 
Lastly, we apply the estimated optimal policy based on the proposed algorithm to the whole dataset, with the initial policy computed by CQL. 
The overall proportion of recommending each action ($A= 0, 1, \cdots 4$) by our estimated policy is $15.2\%, 0.5\%, 2\%, 6\%$ and $76\%$ respectively.
The results imply that our estimated policy recommends the largest dose in most scenarios, with a certain proportion of recommending not receiving any insulin doses. 

\section{Discussion}\label{sec:dis}
In this paper, 	we propose a value enhancement policy optimization method for offline RL problems. One of the key ingredients of the proposed methodology lies in developing a triply robust estimator
for the first-order linear term $\eta_1$ which measures the difference between any two policies. There is a rich line of research on multiply robust estimators in causal inference. For instance, \cite{tchetgen2012semiparametric} proposed triply robust estimators of the marginal natural indirect and direct effects in causal mediation analysis. \cite{wang2018bounded} and \cite{shi2020multiply} developed triply robust estimators for the average treatment effect  using instrumental variables and double negative control variables, respectively. \cite{jiang2020multiply} proposed triply robust estimators for the causal effects within principal strata. Our proposed estimator shares similar statistical properties to these estimators in that its consistency only requires 
\change{two out of three nuisance functions} to be correctly specified. In addition, it is efficient when all functions are correctly specified and satisfy certain convergence rates. 

Based on the triply robust estimator, we propose to search an optimal policy that maximizes the value difference subject to a trust region constraint, and iterate this procedure for value enhancement. In practice, the number of iterations can be determined in a data-adaptive manner via cross-validation. Specifically, one can begin with dividing all data trajectories into $K$ disjoint subsets $\cup_{k=1}^K \mathcal{I}_{k}$. Next, for each $k$, one can apply our proposal to one part of data $\mathcal{I}_k^c$ to compute the optimal policy at each iteration, and apply existing state-of-the-art off policy evaluation methods \citep[see e.g.,][]{jiang2016doubly,kallus2019efficiently,liao2019off,pmlr-v139-shi21d} to the remaining part $\mathcal{I}_k$ to evaluate the value of these policies. We then aggregate the value estimators over different $k$ to get full efficiency and select the number of iterations that maximize the estimated value. 

In addition, we show in our numerical studies that the proposed policy achieves larger values compared to other baseline policies. However, we would like to remark that it took more time to implement our method than those baseline methods. Specifically, it took around 14 minutes to implement the proposed method for one iteration under settings in Section D.2 of the Supplementary Material. In contrast, the running time for V-learning was about 17 minutes, and for CQL about 4 minutes. We also remark that in the offline RL domains, the policy is computed based on a pre-collected dataset. Our primary objective lies in learning an optimal policy with the largest possible value. As long as the procedure can be implemented within a reasonable amount of time, the computation time is not a big issue. This is ultimately different from online RL domains where the policy is usually updated immediately upon the arrival of each observation. But it would be interesting to study how to improve the computational efficiency under the framework of our value enhancement algorithm.

Finally, the proposed method can be used as a stand-alone policy iteration algorithm that starts with a completely random initial policy and iteratively updates this policy to improve its performance. However, the resulting algorithm can be computationally intensive in practice, since it might require a large number of iterations to achieve a near-optimal policy.

\appendix
\setcounter{algorithm}{3}
\section{Some additional technical details}

\subsection{More on the trust region policy optimization}\label{sec:moreTRPO}
\begin{lemma}\label{prop1}
	Suppose $\nu$ is uniformly bounded away from zero. Then there exists some positive constant $c^*>0$ such that $|\eta_2(\pi_{\new},\pi_{\old})|$ is bounded from above by
	\begin{eqnarray}\label{eqn:highorder}
		\frac{c^*\gamma}{1-\gamma} \left[\EE^{d^{\pi_{\old},\nu}} \calD_{\TV}\{\pi_{\old}(\bullet|S),\pi_{\new}(\bullet|S)\} \right]^2\le \frac{c^*\gamma}{1-\gamma} \EE^{d^{\pi_{\old},\nu}}\calD_{\KL}\{\pi_{\old}(\bullet|S),\pi_{\new}(\bullet|S)\}.
	\end{eqnarray}
\end{lemma}
We remark that the upper bound on the RHS of \eqref{eqn:highorder} is tighter than that in Theorem 1 of \cite{schulman2015trust}. 

\textit{Proof}: Note that\vspace{-0.1cm}
\begin{eqnarray}\label{eqn:proofeq1}
	|\eta_2(\pi_{\new},\pi_{\old})|\le  \sum_{a\in\calA,s\in \calS} |\pi_{\new}(a|s)-\pi_{\old}(a|s)||A^{\pi_{\old}}(a,s)||d^{\pi_{\old},\nu}(s)-d^{\pi_{\new},\nu}(s)|.
\end{eqnarray}

\vspace{-0.1cm}
\noindent In the following, we provide an upper bound on $|d^{\pi_{\old},\nu}(s)-d^{\pi_{\new},\nu}(s)|$. By definition, we have \vspace{-0.1cm}
\begin{eqnarray}\label{eqn:proofeq2}
	|d^{\pi_{\old},\nu}(s)-d^{\pi_{\new},\nu}(s)|\le (1-\gamma)\gamma \sum_{t=0}^{+\infty} \gamma^t|p_{t+1}^{\pi_{\old}}(s)-p_{t+1}^{\pi_{\new}}(s)|,
\end{eqnarray}

\vspace{-0.1cm}
\noindent For any $t$, we can define a time-varying policy $\pi(t)$ such that the agent follows $\pi_{\old}$ at the initial $t$ time points and $\pi_{\new}$ subsequently. It follows that\vspace{-0.1cm}
\begin{eqnarray*}
	\sum_{t=0}^{+\infty} \gamma^t|p_{t+1}^{\pi_{\old}}(s)-p_{t+1}^{\pi_{\new}}(s)|\le \sum_{t=0}^{+\infty} \sum_{j=0}^t  \gamma^t|p_{t+1}^{\pi(j+1)}(s)-p_{t+1}^{\pi(j)}(s)|\\\le \sum_{t=0}^{+\infty} \sum_{j=0}^t  \gamma^t \left|\sum_{s_-\in \calS}p_j^{\pi_{\old}}(s_-) \sum_a p(s|a,s_-) \{\pi_{\old}(a|s_-)-\pi_{\new}(a|s_-)\} p_{t-j}^{\pi_{\new}}(s) \right|.
\end{eqnarray*}
By the definition of total variation distance, we have
\begin{eqnarray*}	
	\sum_{t=0}^{+\infty} \gamma^t|p_{t+1}^{\pi_{\old}}(s)-p_{t+1}^{\pi_{\new}}(s)|\le 2\sum_{s_-\in \calS}\|\pi_{\new}(\bullet | s_-)- \pi_{\old}(\bullet | s_-)\|_{\textrm{TV}} \sum_{t=0}^{+\infty} \sum_{j=0}^t  \gamma^tp_{t-j}^{\pi_{\new}}(s)p_j^{\pi_{\old}}(s_-)\\=2\sum_{s_-\in \calS}\|\pi_{\new}(\bullet | s_-)- \pi_{\old}(\bullet | s_-)\|_{\textrm{TV}} \sum_{j=0}^{+\infty} \sum_{t=j}^{+\infty}  \gamma^tp_{t-j}^{\pi_{\new}}(s)p_j^{\pi_{\old}}(s_-)
	\\\le \frac{2}{(1-\gamma)^2} \sum_{s_-\in \calS}\|\pi_{\new}(\bullet | s_-)- \pi_{\old}(\bullet | s_-)\|_{\textrm{TV}} d^{\pi_{\old},\nu}(s_-) d^{\pi_{\new},\nu}(s).
\end{eqnarray*}
Under the given conditions, we have $\nu(s)\ge C$ for some constant $C>0$ and any $s\in\calS$. Consequently, we have $d^{\pi_{\old},\nu}(s)\ge C(1-\gamma)$ and hence $d^{\pi_{\new},\nu}(s)/d^{\pi_{\old},\nu}(s)\le 1/d^{\pi_{\old},\nu}(s)\le C^{-1}(1-\gamma)^{-1}$. It follows that
\begin{eqnarray*}
	\sum_{t=0}^{+\infty} \gamma^t|p_{t+1}^{\pi_{\old}}(s)-p_{t+1}^{\pi_{\new}}(s)|\le \frac{2}{C(1-\gamma)^2} \sum_{s_-\in \calS}\|\pi_{\old}(\bullet | s_-)- \pi_{\new}(\bullet | s_-)\|_{\textrm{TV}} d^{\pi_{\old},\nu}(s_-)d^{\pi_{\old},\nu}(s),
\end{eqnarray*}

\vspace{-0.2cm}
\noindent for some constant $C>0$. Combining this together with \eqref{eqn:proofeq1} and the above inequality yields that $|\eta_2(\pi_{\new},\pi_{\old})|$ is upper bounded by \vspace{-0.1cm}
\begin{eqnarray}\label{eqn:proofeq3}
	\begin{split}
		\frac{2C\gamma}{1-\gamma} \sum_{a\in\calA}\sum_{s,s_-\in \calS}|\pi_{\new}(a|s)-\pi_{\old}(a|s)||A^{\pi_{\old}}(a,s)|\|\pi_{\old}(\bullet | s_-)- \pi_{\new}(\bullet | s_-)\|_{\textrm{TV}}\\ \times d^{\pi_{\old},\nu}(s_-)d^{\pi_{\old},\nu}(s).
	\end{split}	
\end{eqnarray}

\vspace{-0.1cm}
\noindent Note that the $Q$-function and value function correspond to some expected discount cumulative rewards. Under the assumption that the immediate reward is bounded, both functions are bounded. Consequently, we have $\sup_{a,s}|A^{\pi_{\old}}(a,s)|\le c$ for some positive constant $O(1)$. It follows that
\begin{eqnarray*}
	|\eta_2(\pi_{\new},\pi_{\old})|\le 	\frac{4 c\gamma}{C(1-\gamma)} \left\{\sum_{s\in \calS} \|\pi_{\old}(\bullet | s)- \pi_{\new}(\bullet | s)\|_{\textrm{TV}}d^{\pi_{\old},\nu}(s)\right\}^2.
\end{eqnarray*}
This yields the upper bound on the left-hand-side (LHS) of \eqref{eqn:highorder}. 

By Cauchy-Schwarz inequality, we obtain
\begin{eqnarray*}
	|\eta_2(\pi_{\new},\pi_{\old})|\le 	\frac{4 c\gamma}{C(1-\gamma)} \sum_{s\in \calS} \|\pi_{\old}(\bullet | s)- \pi_{\new}(\bullet | s)\|_{\textrm{TV}}^2d^{\pi_{\old},\nu}(s).
\end{eqnarray*}
The upper bound on the RHS of \eqref{eqn:highorder} thus follows by Pinsker's inequality. 

\subsection{Some additional details regarding Proposition \ref{prop2}}\label{sec:moreprop2}

We first give some intuition of this proposition:
Specifically, suppose $\widetilde{A}$ satisfies $\sum_{a} \pi_{\old}(a|s)\widetilde{A}(a,s)=0$ for any $s$. We have the following results:
\begin{itemize}
	\item When $\widetilde{A}=A^{\pi_{\old}}$, $\widetilde{V}=V^{\pi_{\old}}$ and $\widetilde{d}=d^{\pi_{\old}}$, the expectations $\EE \psi_2(O_t;\pi,\pi_{\old},\widetilde{V},\widetilde{A},\widetilde{\omega},\widetilde{d})$ and $\EE \psi_3(O_t;\pi,\pi_{\old},\widetilde{A},\widetilde{\omega},\widetilde{d})$ are equal to zero. Consequently, $\EE \psi(O_t;\pi,\pi_{\old},\widetilde{V},\widetilde{A},\widetilde{\omega},\widetilde{d})$ is equal to the expectation of the plug-in estimator, and thus $\psi(O_t;\pi,\pi_{\old},\widetilde{V},\widetilde{A},\widetilde{\omega},\widetilde{d})$ is unbiased to $\eta_1$. 
	
	\item When $\widetilde{\omega}=\omega^{\pi_{\old}}$ and $\widetilde{d}=d^{\pi_{\old}}$, the expectation $\EE \psi_3(O_t;\pi,\pi_{\old},\widetilde{A},\widetilde{\omega},\widetilde{d})$ is equal to zero. The presence of $\widetilde{\omega}$ in $\psi_2$ guarantees the estimating function is robust to the misspecification of $\widetilde{A}$ and $\widetilde{V}$. Under the assumption that $\sum_{a} \pi_{\old}(a|s)\widetilde{A}(a,s)=0$ for any $s$, the expectation of $\psi_1(\pi,\pi_{\old},\widetilde{A},\widetilde{d})+ \psi_2(O_t;\pi,\pi_{\old},\widetilde{V},\widetilde{A},\widetilde{\omega},\widetilde{d})$ is equal to that of the IS estimator \eqref{eqn:IS1}. Since the IS estimator is unbiased when $\widetilde{\omega}=\omega^{\pi_{\old}}$ and $\widetilde{d}=d^{\pi_{\old}}$, so is $ \psi(O_t;\pi,\pi_{\old},\widetilde{V},\widetilde{A},\widetilde{\omega},\widetilde{d})$.
	
	\item When $\widetilde{A}=A^{\pi_{\old}}$ and $\widetilde{\omega}=\omega^{\pi_{\old}}$, the expectation $\EE \psi_2(O_t;\pi,\pi_{\old},\widetilde{V},\widetilde{A},\widetilde{\omega},\widetilde{d})$ is equal to zero. The presence of $\widetilde{\omega}^{\nu}$ (or $\widetilde{\omega}$, as it is completely determined by $\widetilde{\omega}$) in $\psi_3$ guarantees that the estimating equation is robust to the misspecification of $\widetilde{d}$. More specifically, the expectation of $\psi_1(\pi,\pi_{\old},\widetilde{A},\widetilde{d})+ \psi_3(O_t;\pi,\pi_{\old},\widetilde{A},\widetilde{\omega},\widetilde{d})$ is equal to that of the IS estimator \eqref{eqn:IS2} when $\widetilde{\omega}^{\nu}=\omega^{\pi_{\old},\nu}$. Since the IS estimator is unbiased when $\widetilde{A}=A^{\pi_{\old}}$ and $\widetilde{\omega}=\omega^{\pi_{\old}}$, so is $ \psi(O_t;\pi,\pi_{\old},\widetilde{V},\widetilde{A},\widetilde{\omega},\widetilde{d})$.
\end{itemize}
Now we formally show the expectations of the two argumentation terms $\psi_2(O_t;\widetilde{V},\widetilde{A},\widetilde{\omega},\widetilde{d})$ and $\psi_3(O_t;\widetilde{A},\widetilde{\omega},\widetilde{d})$ are equal to zero when some of the nuisance functions are correctly specified. Consider $\psi_2$ first. When $\widetilde{A}=A^{\pi_{\old}}$ and $\widetilde{V}=V^{\pi_{\old}}$, it follows from the Bellman's equation that \vspace{-0.1cm}
\begin{eqnarray}\label{eqn:Bellman}
	\EE \{R_t+\gamma \widetilde{V}(S_{t+1})-\widetilde{V}(S_t)-\widetilde{A}(A_t,S_t)|A_t,S_t\}=0.
\end{eqnarray}

\vspace{-0.1cm}
\noindent Consequently,$\EE \{\psi_2(O_t;\pi,\pi_{\old},\widetilde{V},\widetilde{A},\widetilde{\omega},\widetilde{d})|A_t,S_t\}=0$ and hence $\EE \psi_2(O_t;\pi,\pi_{\old},\widetilde{V},\widetilde{A},\widetilde{\omega},\widetilde{d})=0$. As for $\psi_3$, note that when $\widetilde{d}=d^{\pi_{\old}}$, we have \vspace{-0.1cm}
\begin{eqnarray*}
	&&\EE \left\{\left.\EE_{\substack{a^*\sim \pi_{\old}(\bullet|S_{t+1})\\S^*\sim \widetilde{d}(\bullet|a^*,S_{t+1})}} \pi_{\new}(a'|S^*)\widetilde{A}(a',S^*)\right|A_t,S_t \right\}\\
	&=&\sum_{s,a}\pi_{\new}(a'|s)\widetilde{A}(a',s)\EE \{d^{\pi_{\old}}(s|a,S_{t+1})\pi_{\old}(a|S_{t+1})|A_t,S_t\}\\
	&=&\sum_{s,a}\pi_{\new}(a'|s)\widetilde{A}(a',s)(1-\gamma) \sum_{t\ge 1}\gamma^{t-1} p_t^{\pi_{\old}}(s;A_t,S_t)\\
	&=&\frac{1}{\gamma} \left\{\EE_{S^*\sim \widetilde{d}(\bullet|A_t,S_t)}\pi_{\new}(a'|S^*)\widetilde{A}(a',S^*)-(1-\gamma)\pi_{\new}(a'|S_t)\widetilde{A}(a',S_t) \right\},
\end{eqnarray*}

\vspace{-0.1cm}
\noindent where the last equality follows from the definition of $\widetilde{d}$. This yields $\EE \{\psi_3(O_t;\pi,\pi_{\old},\widetilde{V},\widetilde{A},\widetilde{\omega},\widetilde{d})|A_t,S_t\}\\=0$ and hence $\EE \psi_3(O_t;\pi,\pi_{\old},\widetilde{V},\widetilde{A},\widetilde{\omega},\widetilde{d})=0$. 

Next, suppose $\sum_{a} \pi_{\old}(a|s)\widetilde{A}(a,s)=0$ for any $s$. We show\vspace{-0.1cm}
\begin{eqnarray}\label{eqn:proofpropeq1}
	\widetilde{A}(a,s)-\frac{1}{1-\gamma}\EE \omega^{\pi_{\old}}(A_t,S_t;a,s) \widetilde{A}(A_t,S_t)=0,
\end{eqnarray} 

\vspace{-0.1cm}
\noindent and \vspace{-0.1cm}
\begin{eqnarray}\label{eqn:proofpropeq2}
	\sum_{a} \{\pi_{\new}(a|s)-\pi_{\old}(a|s) \} \EE \omega^{\pi_{\old}}(A_t,S_t;a,s) \{ \gamma \widetilde{V}(S_{t+1})-\widetilde{V}(S_t) \}=0.
\end{eqnarray} 

\vspace{-0.1cm}
\noindent Combining these two equations yields that the expectation $\EE \{\psi_1(\pi,\pi_{\old},\widetilde{A},\widetilde{\omega})+\psi_2(O_t;\widetilde{V},\widetilde{A},\widetilde{\omega},\widetilde{d})\}$ is equal to the expectation of the IS estimator in \eqref{eqn:IS1} when $\widetilde{\omega}$ is correctly specified. Consequently, $\EE \psi(O_t;\widetilde{V},\widetilde{A},\widetilde{\omega},\widetilde{d})$ is unbiased to $\eta_1$ under the assumption in (B2). 

We first show \eqref{eqn:proofpropeq1}. We observe that\vspace{-0.1cm}
\begin{eqnarray*}
	\frac{1}{1-\gamma}\EE \omega^{\pi_{\old}}(A_t,S_t;a,s) \widetilde{A}(A_t,S_t)=\widetilde{A}(a,s)+\sum_{t\ge 1} \gamma^t \sum_{a',s'} p_t(s'|a,s)\pi_{\old}(a'|s')\widetilde{A}(a',s').
\end{eqnarray*}

\vspace{-0.1cm}
\noindent The second term on the RHS is equal to zero under the condition that $\sum_{a} \pi_{\old}(a|s)\widetilde{A}(a,s)=0$ for any $s$. This yields \eqref{eqn:proofpropeq1}. We next show \eqref{eqn:proofpropeq2}. With some calculations, 
\vspace{-0.1cm}
\begin{eqnarray*}
	\EE \omega^{\pi_{\old}}(A_t,S_t;a,s) \{ \gamma \widetilde{V}(S_{t+1})-\widetilde{V}(S_t) \}=(1-\gamma) \sum_{t\ge 1} \gamma^t \sum_{s'} p_t(s'|a,s)\widetilde{V}(s')\\
	-(1-\gamma)\sum_{t\ge 0} \gamma^t \sum_{s'} p_t(s'|a,s)\widetilde{V}(s')=-(1-\gamma)\widetilde{V}(s).
\end{eqnarray*}
Note that the RHS is independent of $a$. Consequently, we have $\sum_{a} \{\pi(a|s)-\pi_{\old}(a|s) \}\widetilde{V}(s)=0$. This yields \eqref{eqn:proofpropeq2}. 

Equation \eqref{eqn:proofpropeq2} implies that $\EE \psi_2(O_t;\widetilde{V},\widetilde{A},\widetilde{\omega},\widetilde{d})=\EE \psi_2(O_t;V^{\pi_{\old}},\widetilde{A},\widetilde{\omega},\widetilde{d})$ when $\widetilde{\omega}=\omega^{\pi_{\old}}$. If further $\widetilde{A}$ is correctly specified, we obtain that $\EE \psi_2(O_t;\widetilde{V},\widetilde{A},\widetilde{\omega},\widetilde{d})=0$.

Finally, we show \vspace{-0.1cm}
\begin{eqnarray*}
	\EE \sum_{a^*\in \calA} \frac{ \omega^{\pi_{\old},\nu}(A_t,S_t)}{1-\gamma}\left[\gamma\EE_{\substack{a'\sim \pi_{\old}(\bullet|S_{t+1})\\S^*\sim \widetilde{d}(\bullet|a',S_{t+1})}}\widetilde{A}(a^*,S^*)\pi(a^*|S^*)-\EE_{S^*\sim \widetilde{d}(\bullet|A_t,S_t)}\widetilde{A}(a^*,S^*)\pi(a^*|S^*)\right]\\=-\psi_1(\pi,\pi_{\old},\widetilde{A},\widetilde{d}).
\end{eqnarray*}

\vspace{-0.1cm}
\noindent This further yields that the expectation of $\psi_1(\pi,\pi_{\old},\widetilde{A},\widetilde{d})+ \psi_3(O_t;\widetilde{A},\widetilde{\omega},\widetilde{d})$ is equal to that of the IS estimator \eqref{eqn:IS2} when $\widetilde{\omega}^{\nu}=\omega^{\pi_{\old},\nu}$. Consequently, $\EE \psi(O_t;\widetilde{V},\widetilde{A},\widetilde{\omega},\widetilde{d})$ is unbiased when $\widetilde{A}$ and $\widetilde{\omega}$ are correctly specified. 

With some calculations, the LHS is equal to \vspace{-0.1cm}
\begin{eqnarray*}
	\EE \sum_{a^*,s^*} \frac{\omega^{\pi_{\old},\nu}(A_t,S_t)}{1-\gamma}\left[\left\{\gamma \sum_{a'} \pi_{\old}(a'|S_{t+1}) \widetilde{d}(s^*|a',S_{t+1})- \widetilde{d}(s^*|A_t,S_t)\right\}\widetilde{A}(a^*,s^*)\pi_{\new}(a^*|s^*)\right]
	\\=\frac{1}{1-\gamma}\sum_{a^*,s^*,a',s'}\pi_{\new}(a^*|s^*) \pi_{\old}(a'|s')\widetilde{d}(s^*|a',s')\widetilde{A}(a^*,s^*) \{d^{\pi_{\old}}(s')-(1-\gamma) \nu(s')\}\\
	-\frac{1}{1-\gamma}\sum_{a^*,s^*,a',s'}\pi_{\new}(a^*|s^*)\pi_{\old}(a'|s')\widetilde{d}(s^*|a',s')\widetilde{A}(a^*,s^*)d^{\pi_{\old}}(s')\\
	=-	\sum_{a^*,s^*,a',s'}\pi_{\new}(a^*|s^*) \pi_{\old}(a'|s')\widetilde{d}(s^*|a',s')\widetilde{A}(a^*,s^*)\nu(s')=-\sum_{a^*,s^*}\pi_{\new}(a^*|s^*) \widetilde{d}^{\nu}(s^*)\widetilde{A}(a^*,s^*),
\end{eqnarray*}

\vspace{-0.1cm}
\noindent where the last equation follows from the definition of $\widetilde{d}^{\nu}$. The proof is hence completed. 

\subsection{VC type class}\label{sec:VCtypeclass}
We introduce the notion of the VC type class in this section. Specifically, let $\mathcal{F}$ denote a class of measurable functions, with a measurable envelope function $F$ such that $\sup_{f\in \mathcal{F}}|f| \le F$. For any probability measure $Q$, let $e_Q$ denote a semi-metric on $\mathcal{F}$ such that $e_Q(f_1,f_2)=\|f_1-f_2\|_{Q,2} = \sqrt{Q |f_1-f_2|^2}$. An $\epsilon$-net of the space $(\mathcal{F}, e_Q)$ is a subset $\mathcal{F}_{\epsilon}$ of $\mathcal{F}$, such that for every $f\in \mathcal{F}$, there exists some $f_{\epsilon}\in \mathcal{F}_{\epsilon}$ satisfying $e_Q(f,f_{\epsilon}) < \epsilon$. We say that $\mathcal{F}$ is a VC type class with envelope $F$, if there exist constants $c_0 > 0, c_1 \ge 1$, such that $\sup_Q \mathbb{N}\left( \mathcal{F},e_Q, \epsilon\|F\|_{Q,2} \right) \le (c_0 / \epsilon)^{c_1}$, for all $0 < \epsilon \le 1$, where the supremum is taken over all finitely discrete probability measures on the support of $\mathcal{F}$, and $\mathbb{N}\left( \mathcal{F},e_Q, \epsilon\|F\|_{Q,2} \right)$ is the infimum of the cardinality of $\epsilon\|F\|_{Q,2}$-nets of $\mathcal{F}$. We refer to $c_1$ as the VC index of $\mathcal{F}$.

\subsection{Semiparametric Efficiency}

In the i.i.d. case, for parametric models, the variance of any unbiased estimator must be greater than or equal to the Cr{\'a}mer-Rao lower bound \citep[][Section 7.3]{casella2002statistical} and the maximum likelihood estimator is known to be efficiency under certain regularity conditions. In semiparametric theory, the efficiency bound is defined as the supremum of Cr{\'a}mer-Rao lower bounds over all regular parametric submodels to move beyond parametric setup \citep[see e.g.,][]{tsiatis2007semiparametric}. In our setup, the observations are time-dependent. We adopt the definition in \cite{komunjer2010semiparametric} and \cite{kallus2019efficiently} that corresponds to a generalization of the classical semiparametric efficiency bound to the non i.i.d. setting. 

\vspace{-0.1cm}
\noindent
\section{More on the algorithm}
\subsection{More on conditional discounted stationary probability ratio}\label{sec:cdspr}
Consider the following optimization problem\vspace{-0.1cm}
\begin{eqnarray}\label{optimize}
	\argmin_{\omega\in \Omega} \sup_{f\in \mathcal{F}} \left|\sum_{\substack{ (i,t)\neq (i',t')}} \Delta(\omega,f,\pi_{\old};i,t,i',t')\right|^2.
\end{eqnarray}

\vspace{-0.1cm}
\noindent We set $\mathcal{F}$ to a unit ball of a reproducing kernel Hilbert space (RFHS), i.e., $\mathcal{F}=\{f\in \mathcal{H}:\|f\|_{\mathcal{H}}=1\}$, 
where 
\begin{eqnarray*}
	\mathcal{H}=\left\{f(\cdot)=\sum_{(i,t)\neq (i',t')} b_{i,t,i',t'} \kappa(X_{i',t'},X_{i,t};\cdot): b_{i,t,i',t'}\in \mathbb{R} \right\},
\end{eqnarray*}
for some positive definite kernel $\kappa(\cdot;\cdot)$, where $X_{i,t}$ is a shorthand for the state-action pair $(S_{i,t},A_{i,t})$. Similar to Theorem 2 of \cite{liu2018breaking}, we can show the optimization problem in \eqref{optimize} is then reduced to\vspace{-0.1cm}
\begin{eqnarray*}
	\argmin_{\omega\in \Omega}  \sum_{ (i_1,t_1)\neq (i_1',t_1')} \sum_{(i_2,t_2)\neq (i_2',t_2')} D(\omega,\pi_{\old};i_1,t_1,i_1',t_1',i_2,t_2,i_2',t_2'),
\end{eqnarray*}

\vspace{-0.1cm}
\noindent where $D(\omega,\pi;i_1,t_1,i_1',t_1',i_2,t_2,i_2',t_2')$ is given by
\begin{eqnarray*}
	\frac{\omega(X_{i_1',t_1'};X_{i_1,t_1})}{(1-\gamma)^{-1}}\Big\{\gamma\EE_{a\sim \pi(\bullet|S_{i_1',t_1'+1})} \kappa(S_{i_1',t_1'+1},a,X_{i_1,t_1};X_{i_2,t_2},X_{i_2,t_2})-\kappa(X_{i_1',t_1'},X_{i_1,t_1};X_{i_2,t_2},X_{i_2,t_2})\Big\}\\+\frac{\omega(X_{i_2',t_2'};X_{i_2,t_2})}{(1-\gamma)^{-1}}\Big\{\gamma\EE_{a\sim \pi(\bullet|S_{i_2',t_2'+1})} \kappa(S_{i_2',t_2'+1},a,X_{i_2,t_2};X_{i_1,t_1},X_{i_1,t_1})-\kappa(X_{i_2',t_2'},X_{i_2,t_2};X_{i_1,t_1},X_{i_1,t_1})\Big\}\\+\omega(X_{i_1',t_1'};X_{i_1,t_1})\omega(X_{i_2',t_2'};X_{i_2,t_2}) \Big\{ \gamma^2\EE_{\substack{a_1\sim \pi(\bullet|S_{i_1',t_1'+1})\\a_2\sim \pi(\bullet|S_{i_2',t_2'+1}) }} \kappa(S_{i_2',t_2'+1},a,X_{i_2,t_2};S_{i_1',t_1'+1},a_1,X_{i_1,t_1}) \\-\gamma \EE_{\substack{a_1\sim \pi(\bullet|S_{i_1',t_1'+1})}} \kappa(S_{i_1',t_1'+1},a,X_{i_1,t_1};X_{i_2',t_2'},X_{i_2,t_2})-\gamma \EE_{\substack{a_2\sim \pi(\bullet|S_{i_2',t_2'+1})}} \kappa(S_{i_2',t_2'+1},a,X_{i_2,t_2};X_{i_1',t_1'},X_{i_1,t_1})\\
	+\kappa(X_{i_2',t_2'},X_{i_2,t_2};X_{i_1',t_1'},X_{i_1,t_1})\Big\}+(1-\gamma)^2 \kappa(X_{i_1,t_1},X_{i_1,t_1};X_{i_2,t_2},X_{i_2,t_2}).
\end{eqnarray*}
In our implementation, we set $\Omega$ to the class of neural networks. The detailed estimating procedure is given in Algorithm \ref{alg1}. 
\begin{algorithm}[t!]
	\caption{Estimation of the density ratio.}
	\label{alg1}
	\begin{algorithmic}
		\item
		\begin{description}
			\item[\textbf{Input}:] The data subset in $\mathcal{I}_{\ell}$.
			
			\item[\textbf{Initial}:] Initial the density ratio $\omega=\omega_{\beta}$ to be a neural network parameterized by $\beta$.
			
			\item[\textbf{for}] iteration $=1,2,\cdots$ \textbf{do}
			\begin{enumerate}
				\item[a] Randomly sample batches $\mathcal{M}$, $\mathcal{M}^*$ from the data transitions.
				
				\item[b] {\textbf{Update}} the parameter $\beta$ by $$\beta\leftarrow \beta-\epsilon {|\mathcal{M}|\choose 2}^{-2}\sum_{\substack{(i_1,t_1),(i_1',t_1')\in \mathcal{M}\\ (i_1,t_1)\neq (i_1',t_1') }}\sum_{\substack{(i_2,t_2),(i_2',t_2')\in \mathcal{M}\\ (i_2,t_2)\neq (i_2',t_2') }} \nabla_{\beta} D(\frac{\omega_{\beta}}{z_{\omega_{\beta}}},\pi_{\old};i_1,t_1,i_1',t_1',i_2,t_2,i_2',t_2'),$$ 
				where $z_{\omega_{\beta}}$ is a normalization constant $$z_{\omega_{\beta}}(\cdot;S_{i,t},A_{i,t})=\frac{1}{|\mathcal{M}^*|} \sum_{(i',t') \in \mathcal{M}^* } \omega_{\beta}(X_{i',t'};X_{i,t}).$$ 
			\end{enumerate}
			\item[\textbf{Output}] $\omega_{\beta}$. 
		\end{description}
	\end{algorithmic}
\end{algorithm}

\subsection{More on conditional discounted visitation probability}\label{sec:morecond}
We first provide an upper bound for $\mathcal{D}_{\TV}(\widehat{d}^{(\ell)}(\bullet|a,s), d^{\pi_{\old}^{(\ell)}}(\bullet|a,s))$ for a given pair $(a,s)$. Notice that the total variation distance corresponds to a special case of $f$-divergence. This allows us to represent $\mathcal{D}_{\TV}(\widehat{d}^{(\ell)}(\bullet|a,s), d^{\pi_{\old}^{(\ell)}}(\bullet|a,s))$ as
\begin{eqnarray*}
	\sup_{\|f\|_{\infty}\le 1/2}|\EE_{S^*\sim d^{\pi_{\old}^{(\ell)}}(\bullet|a,s)} f(S^*)-\EE_{S^*\sim \widehat{d}^{(\ell)}(\bullet|a,s)} f(S^*)|,
\end{eqnarray*}
or equivalently, 
\begin{eqnarray*}
	\sup_{\|f\|_{\infty}\le 1/2}\Big|\EE_{S^*\sim d^{\pi_{\old}^{(\ell)}}(\bullet|a,s)} f(S^*)-\frac{1-\gamma}{M} \sum_{m=1}^M \sum_{t'=0}^{T'} \gamma^t f(\widetilde{S}_{t'}^{(m)})\Big|.
\end{eqnarray*}

By definition, for a given $f$, we can decompose 
the difference into \vspace{-0.1cm}
\begin{eqnarray*}
	\left|\EE_{S^*\sim d^{\pi_{\old}^{(\ell)}}(\bullet|a,s)} f(S^*)-\frac{1-\gamma}{M} \sum_{m=1}^M \sum_{t'=0}^{T'} \gamma^t f(\widetilde{S}_{t'}^{(m)})\right|\\
	\le \sum_{t'=0}^{T'} \gamma^{t'} \left|(1-\gamma)\EE_{S^*\sim p_{t'}^{\pi_{\old}^{(\ell)}}(\bullet|a,s)}f(S^*)-\frac{1-\gamma}{M} \sum_{m=1}^Mf(\widetilde{S}_{t'}^{(m)})\right|\\+(1-\gamma)\sum_{t'=T'+1}^{+\infty} \gamma^{t'} |\EE_{S^*\sim d^{\pi_{\old}^{(\ell)}}(\bullet|a,s)}f(S^*)|.
\end{eqnarray*}
Since $f$ is uniformly bounded by $1/2$, the second term on the RHS is bounded by $\gamma^{T'}$ that converges to zero as $T'\to \infty$.   

As for the first term, it can be further bounded from above by \vspace{-0.1cm}
\begin{eqnarray}\label{eqn:proof}
	\sum_{t'=0}^{T'} (1-\gamma) \gamma^{t'} \left|\EE_{S^*\sim p_{t'}^{\pi_{\old}^{(\ell)}}(\bullet|a,s)}f(S^*)-\EE f(\widetilde{S}_{t'})\right|+\sum_{t'=0}^{T'}(1-\gamma)\gamma^{t'}\left|\frac{1}{M} \sum_{m=1}^Mf(\widetilde{S}_{t'}^{(m)})-\EE f(\widetilde{S}_{t'})\right|.
\end{eqnarray}

\vspace{-0.1cm}
\noindent The expectation of the second term in \eqref{eqn:proof} can be upper bounded by \vspace{-0.1cm}
\begin{eqnarray*}
	\sum_{t'=0}^{T'}(1-\gamma)\gamma^{t'}\sqrt{\Var\left(\frac{1}{M} \sum_{m=1}^Mf(\widetilde{S}_{t'}^{(m)})-\EE f(\widetilde{S}_{t'})\right)}\le M^{-1/2},
\end{eqnarray*}

\vspace{-0.1cm}
\noindent by Cauchy-Schwarz inequality. Consequently, the second term in \eqref{eqn:proof} decays to zero as $M\to \infty$. 

Finally, consider the first term in \eqref{eqn:proof}. We use $(S_t^{(j)},A_t^{(j)})$ to denote the state-action pair measured at time $t$ that follows $\pi_{\old}^{(\ell)}$ and the transition function $p$ at the first $j$th steps conditional on $(A_0,S_0)=(a,s)$, and then follows $\pi_{\old}^{(\ell)}$ and the transition function $\mathcal{N}(\widehat{\mu}^{(\ell)}, \widehat{\Sigma}^{(\ell)})$ in the subsequent steps. For each $t'$, we have\vspace{-0.1cm}
\begin{eqnarray*}
	\left|\EE_{S^*\sim p_{t'}^{\pi_{\old}^{(\ell)}}(\bullet|a,s)}f(S^*)-\EE f(\widetilde{S}_{t'})\right|\le \sum_{j=1}^{t'} \left|\EE f(S_{t'}^{(j-1)})- \EE f( S_{t'}^{(j)})\right|.
\end{eqnarray*}

\vspace{-0.1cm}
\noindent Let $\widehat{p}_{t}^{\pi_{\old}^{(\ell)}}(\bullet|a,s)$ denote the distribution function of the state vector $S_t$ at time $t$ that follows $\pi_{\old}$ and the estimated transition function $\mathcal{N}(\widehat{\mu}^{(\ell)}, \widehat{\Sigma}^{(\ell)})$ conditional on $(A_0=a,S_0=s)$. We omit the subscript $t$ and the superscript $\pi_{\old}^{(\ell)}$ when $t=1$. 
Suppose $f$ is uniformly bounded by some constant $c>0$. Then $|\EE_{S^{*}\sim \widehat{p}_{t'-j}^{\pi_{\old}^{(\ell)}}(\bullet|a^*,s),a^*\sim \pi_{\old}^{(\ell)}(\bullet|s)}f(S^{*})|$ is uniformly bounded by $c$ for any $s$ as well. Consequently,
\vspace{-0.1cm}
\begin{eqnarray*}
	\left|\EE f(S_{t'}^{(j-1)})- \EE f( S_{t'}^{(j)})\right|\le \EE\left| \EE_{\substack{S^*\sim p(\bullet|A_{j-1}^{(j-1)},,S_{j-1}^{(j-1)})}} \left\{\EE_{\substack{S^{**}\sim  \widehat{p}_{t'-j}^{\pi_{\old}^{(\ell)}}(\bullet|a^{**},S^*)\\a^{**}\sim \pi_{\old}^{(\ell)}(\bullet|S^*)}}f(S^{**})\right\}\right.\\-\left.\EE_{\substack{S^*\sim \widehat{p}(\bullet|A_{j-1}^{(j-1)},S_{j-1}^{(j-1)})}} \left\{\EE_{\substack{S^{**}\sim  p_{t'-j}^{\pi_{\old}^{(\ell)}}(\bullet|a^{**},S^*)\\a^{**}\sim \pi_{\old}^{(\ell)}(\bullet|S^*)}}f(S^{**})\right\}  \right|\\
	\le c\EE \calD_{\TV}\{p(\bullet|A_{j-1}^{(j-1)},S_{j-1}^{(j-1)}),\widehat{p}(\bullet|A_{j-1}^{(j-1)},S_{j-1}^{(j-1)})\}.
\end{eqnarray*}
It follows that the first term in \eqref{eqn:proof} can be upper bounded by\vspace{-0.1cm}
\begin{eqnarray}\nonumber
	&&\sum_{t'=0}^{+\infty} (1-\gamma)\gamma^{t'} \left|\EE_{S^*\sim p_{t'}^{\pi_{\old}^{(\ell)}}(\bullet|a,s)}f(S^*)-\EE f(\widetilde{S}_{t'})\right|\\\nonumber
	&\le& c	\sum_{t'=1}^{+\infty} (1-\gamma)\gamma^{t'} \sum_{j=1}^{t'} \EE \calD_{\TV}\{p(\bullet|A_{j-1}^{(j-1)},S_{j-1}^{(j-1)}),\widehat{p}(\bullet|A_{j-1}^{(j-1)},S_{j-1}^{(j-1)})\}\\\nonumber
	&=& c\sum_{j=1}^{+\infty} \gamma^{\ell} \calD_{\TV}\{p(\bullet|A_{j-1}^{(j-1)},S_{j-1}^{(j-1)}),\widehat{p}(\bullet|A_{j-1}^{(j-1)},S_{j-1}^{(j-1)})\}\\\label{eqn:dtv}
	&=& \frac{c\gamma}{1-\gamma}\EE_{(A^*,S^*)\sim q^{\pi_{\old}^{(\ell)}}(\bullet,\bullet;a,s)} \calD_{\TV}\{p(\bullet|A^*,S^*),\widehat{p}(\bullet|A^*,S^*)\},
\end{eqnarray}

\vspace{-0.1cm}
\noindent where $q^{\pi_{\old}^{(\ell)}}(\bullet,\bullet;a,s)$ denotes the conditional discounted visitation probability of the state-action pair, i.e., \vspace{-0.1cm}
\begin{eqnarray*}
	q^{\pi_{\old}^{(\ell)}}(a',s';a,s)=(1-\gamma)\mathbb{I}(a'=a,s'=s)+(1-\gamma)\sum_{t=1}^{+\infty} \gamma^t \pi_{\old}^{(\ell)}(a'|s')p_t^{\pi_{\old}^{(\ell)}}(s'|a,s).
\end{eqnarray*}

\vspace{-0.1cm}
\noindent Let $\widetilde{p}$ denote the normal density function with mean $\widehat{\mu}^{(\ell)}$ and covariance matrix $\Sigma$. It follows from the triangle inequality that
\begin{eqnarray*}
	\calD_{\TV}(p,\widehat{p})\le \calD_{\TV}(p,\widetilde{p})+\calD_{\TV}(\widetilde{p},\widehat{p}). 
\end{eqnarray*}
According to Proposition 2.1 of \cite{devroye2018total}, the total variation distance between $p$ and $\widetilde{p}$ is upper bounded by $0.5\sqrt{(\mu-\widehat{\mu}^{(\ell)})^\top \Sigma^{-1} (\mu-\widehat{\mu}^{(\ell)})}\le c\|\mu-\widehat{\mu}^{(\ell)}\|_2$ for some constant $c>0$ under the condition that the minimum eigenvalue of $\Sigma$ is bounded away from zero.  Meanwhile, it follows from Theorem 1.1 of \cite{devroye2018total} that the total variation distance between $\widehat{p}$ and $\widetilde{p}$ is upper bounded by $1.5\sqrt{\sum_i \lambda_i^2}$ where $\{\lambda_i\}_i$ denote the eigenvalues of $\Sigma^{-1} (\widehat{\Sigma}^{(\ell)}-\Sigma)$, under the conditions that both $\widehat{\Sigma}^{(\ell)}$ and $\Sigma$ are positive definite. The sum of squared eigenvalues equals the squared Frobenious norm of $\Sigma^{-1} (\widehat{\Sigma}^{(\ell)}-\Sigma)$, which can be further upper bounded by $\|\Sigma^{-1}\|_2^2 \|\widehat{\Sigma}^{(\ell)}-\Sigma\|_F^2\le c^2 \|\widehat{\Sigma}^{(\ell)}-\Sigma\|_F^2$. Therefore,
\begin{eqnarray*}
	\calD_{\TV}(p,\widehat{p})\le \frac{c}{2} \|\mu-\widehat{\mu}^{(\ell)}\|_2+\frac{3c}{2}\|\Sigma-\widehat{\Sigma}^{(\ell)}\|_F. 
\end{eqnarray*}

\vspace{-0.1cm}
\noindent To summarize, we have shown that
\begin{eqnarray*}
	&&\mathcal{D}_{\TV}(\widehat{d}^{(\ell)}(\bullet|a,s), d^{\pi_{\old}^{(\ell)}}(a,s))\le \gamma^{T'}+M^{-1/2}\\
	&+&O(1)\EE_{(A^*,S^*)\sim q^{\pi_{\old}^{(\ell)}}(\bullet,\bullet;a,s)} [\|\mu(S^*,A^*)-\widehat{\mu}^{(\ell)}(S^*,A^*)\|_2+\|\Sigma(S^*,A^*)-\widehat{\Sigma}^{(\ell)}(S^*,A^*)\|_F],
\end{eqnarray*}
for some positive constant $O(1)$. 

According to the Cauchy-Schwarz inequality, the aggregated squared total variation distance is upper bounded by
\begin{eqnarray*}
	O(1)\EE_{(a,s)\sim p_{\infty}} \{\EE_{(A^*,S^*)\sim q^{\pi_{\old}^{(\ell)}}(\bullet,\bullet;a,s)} [\|\mu(S^*,A^*)-\widehat{\mu}^{(\ell)}(S^*,A^*)\|_2+\|\Sigma(S^*,A^*)-\widehat{\Sigma}^{(\ell)}(S^*,A^*)\|_F]\}^2\\
	+3\gamma^{2T'}+\frac{3}{M},
\end{eqnarray*}
for some positive constant $O(1)$. Using the Cauchy Schwarz inequality again and notice that $p_{\infty}$ is bounded away from zero, the first term can be further upper bounded by
\begin{eqnarray*}
	O(1)\EE_{(a,s)\sim p_{\infty}} \EE_{(A^*,S^*)\sim q^{\pi_{\old}^{(\ell)}}(\bullet,\bullet;a,s)} [\|\mu(S^*,A^*)-\widehat{\mu}^{(\ell)}(S^*,A^*)\|_2+\|\Sigma(S^*,A^*)-\widehat{\Sigma}^{(\ell)}(S^*,A^*)\|_F]^2\\
	\le O(1)\EE_{(A^*,S^*)\sim p_{\infty}} [\|\mu(S^*,A^*)-\widehat{\mu}^{(\ell)}(S^*,A^*)\|_2+\|\Sigma(S^*,A^*)-\widehat{\Sigma}^{(\ell)}(S^*,A^*)\|_F]^2.
\end{eqnarray*}
This completes the proof. 

\section{Proofs}
Throughout this section, we use $C$ and $c$ to denote some generic constant whose value is allowed to change from place to place. 

\subsection{Proof of Theorem \ref{thm3}}
The proof of Theorem \ref{thm3} is straightforward. We first note that, due to the trust-region condition in \eqref{eqn:constraint}, we have \vspace{-0.1cm}
\begin{eqnarray}\label{eqn:constraint0}
	(1-\gamma) \frac{1}{\mathbb{L}}\sum_{\ell=1}^{\mathbb{L}}\EE_{S^*\sim \nu} \calD_{\KL}(\pi_{\old}^{(\ell)}(\bullet|S^*), \pi_{\new}(\bullet|S^*))\le \delta,
\end{eqnarray}

\vspace{-0.1cm}
\noindent as $\widehat{d}^{(\ell),\nu}(s)\ge (1-\gamma) \nu(s)$ for any $s$. 

Next, using similar arguments in the proof of Lemma \ref{prop1}, we can show \vspace{-0.1cm}
\begin{eqnarray*}
	|\calV(\pi_{\new})-\calV(\pi_{\old})|=\left|\EE_{S^* \sim d^{\pi_{\new}}} \sum_{a} \{\pi_{\new}(a|S^*)-\pi_{\old}(a|S^*)\} A^{\pi_{\old}}(a,S^*)\right|\\
	\le O(1) \EE_{S^*\sim d^{\pi_{\new}}} \|\pi_{\new}(\bullet|S^*)-\pi_{\old}(\bullet|S^*)\|_{\TV}\le O(1)\sqrt{\EE_{S^*\sim d^{\pi_{\new}}}\calD_{\KL}(\pi_{\old}(\bullet|S^*), \pi_{\new}(\bullet|S^*))},
\end{eqnarray*}

\vspace{-0.1cm}
\noindent where $O(1)$ denotes some positive constant. The first equality is due to \eqref{eqn:key}, the first inequality is due to the condition that the immediate reward is uniformly bounded (and so is the advantage function), the second inequality follows from Pinsker's inequality. Under the condition that $\nu(\cdot)$ is bounded uniformly away from zero, we obtain that \vspace{-0.1cm}
\begin{eqnarray*}
	\EE_{S^*\sim d^{\pi_{\new}}} \calD_{\KL}(\pi_{\old}(\bullet|S^*),\pi_{\new}(\bullet|S^*))\le C \EE_{S^*\sim \nu} \calD_{\KL}(\pi_{\old}(\bullet|S^*),\pi_{\new}(\bullet|S^*)),
\end{eqnarray*}

\vspace{-0.1cm}
\noindent for some constant $C>0$. It follows that
\begin{eqnarray*}
	\left|\calV(\pi_{\new})-\frac{1}{\mathbb{L}}\sum_{\ell=1}^{\mathbb{L}}\calV(\pi_{\old}^{(\ell)})\right|\le \frac{1}{\mathbb{L}}\sum_{\ell=1}^{\mathbb{L}} |\calV(\pi_{\new})-\calV(\pi_{\old}^{(\ell)})|\\\le c\sum_{\ell=1}^{\mathbb{L}}\sqrt{\EE_{S^*\sim \nu}\calD_{\KL}(\pi_{\old}^{(\ell)}(\bullet|S^*), \pi_{\new}(\bullet|S^*))}\le c\mathbb{L} \sqrt{\sum_{\ell=1}^{\mathbb{L}}\EE_{S^*\sim \nu}\calD_{\KL}(\pi_{\old}^{(\ell)}(\bullet|S^*), \pi_{\new}(\bullet|S^*))},
\end{eqnarray*}
for some constant $c>0$. Theorem \ref{thm3} thus follows from \eqref{eqn:constraint0}.

\subsection{Proof of Theorem \ref{thm4}}
We begin with some auxiliary lemmas. 
\begin{lemma}\label{lemma:aux}
	Under (C5), there exists some constant $c_0>0$ such that for any $\pi$, \vspace{-0.1cm}
	\begin{eqnarray*}
		\calV(\pi^{\tiny{opt}})-\calV(\pi)\ge c_0 \{\EE_{S^*\sim d^{\pi^{\tiny{opt}},\nu}} \|\pi^{\tiny{opt}}(\bullet|S^*)-\pi(\bullet|S^*)\|_{\TV}\}^{1+1/\alpha},
	\end{eqnarray*}
	where $\alpha$ is the exponent in (C5). 
\end{lemma}

\begin{lemma}\label{lemma:aux1}
	Let $\widehat{\eta}_1^*(\pi,\pi_{\old}^{(\ell)})=(|\mathcal{I}_{\ell}|T)^{-1} \sum_{i\in \mathcal{I}_{\ell}}\sum_{t<T} \psi(O_{i,t};\pi,\pi_{\old}^{(\ell)},Q^{\pi_{\old}^{(\ell)}},\omega^{\pi_{\old}^{(\ell)}},d^{\pi_{\old}^{(\ell)}})$ and $\widehat{\eta}_1^{(\ell)}(\pi)=(|\mathcal{I}_{\ell}|T)^{-1} \sum_{i\in \mathcal{I}_{\ell}}\sum_{t<T} \psi(O_{i,t};\pi,\pi_{\old}^{(\ell)},\widehat{Q}^{(\ell)},\widehat{\omega}^{(\ell)},\widehat{d}^{(\ell)})$.  
	Under the given conditions, 
	\begin{eqnarray*}
		\sup_{\pi_1,\pi_2\in \Pi} \frac{|\widehat{\eta}_1^{(\ell)}(\pi_1)-\widehat{\eta}_1^*(\pi_1,\pi_{\old}^{(\ell)})-\widehat{\eta}_1^{(\ell)}(\pi_2)+\widehat{\eta}_1^*(\pi_2,\pi_{\old}^{(\ell)})|}{\mathbb{E}_{S^*\sim d^{\pi^{\tiny{opt}},\nu}}\|\pi_1(\bullet|S^*)-\pi_2(\bullet|S^*)\|_{\TV}}=o_p(\{NT\}^{-1/(2+2\alpha)})+O_p(\{(NT)\}^{\kappa_4/2-1/2}). 
	\end{eqnarray*}
\end{lemma}

\begin{lemma}\label{lemma:aux2}
	Let $\{Z_t:t\ge 0\}$ be a stationary $\beta$-mixing process with the $\beta$-mixing coefficient $\{\beta(q):q\ge 0\}$. Let $\mathcal{F}$ be a pointwise measurable class of functions that take $Z_t$ as input. For any $f\in \mathcal{F}$, suppose $\EE \{f(Z_0)\} = 0$. Let $\sigma^2>0$ be a positive constant, such that $\sup_{f\in \mathcal{F}} \EE \{f^2(Z_0)\} \le \sigma^2$. Suppose the envelop function is uniformly bounded by some constant $C>0$. In addition, suppose $\mathcal{F}$ belongs to the class of VC-type functions such that $\sup_Q N(\mathcal{F}, e_Q, \varepsilon \|F\|_{Q,2})\le (A/\varepsilon)^{v}$ for some $A\ge e,v\ge 1$. 
	Then \vspace{-0.1cm} 
	\begin{eqnarray*}
		\sup_{f\in \mathcal{F}}\left|\sum_{t=0}^{T-1} f(Z_t)\right| = O_p\left\{\sqrt{v q\sigma^2T \log \left(\frac{AC}{\sigma}\right)} + v C \log \left(\frac{AC}{\sigma}\right)+q\right\},
	\end{eqnarray*}
	for any $1\le q<T/2$ such that $T\beta(q)/q=o(1)$.
\end{lemma}

We next sketch an outline of the proof. We aim to provide an upper bound for the value difference $\calV(\pi^{\tiny{opt}})-\calV(\pi_{\new})$. It can be represented by \vspace{-0.1cm}
\begin{eqnarray*}
	\frac{1}{(1-\gamma)\mathbb{L}}\sum_{\ell=1}^{\mathbb{L}}\left\{\eta_1(\pi^{\tiny{opt}},\pi_{\old}^{(\ell)})+\eta_2(\pi^{\tiny{opt}},\pi_{\old}^{(\ell)})-\eta_1(\pi_{\new},\pi_{\old}^{(\ell)})-\eta_2(\pi_{\new},\pi_{\old}^{(\ell)})\right\}.
\end{eqnarray*} 

\vspace{-0.1cm}
\noindent We break the rest of the proof into several steps. In the first step, we provide upper bounds for the high-order remainder terms $\eta_2(\pi^{\tiny{opt}},\pi_{\old}^{(\ell)})$ and $\eta_2(\pi_{\new},\pi_{\old}^{(\ell)})$. 

We first show that when $\delta$ is set to a sufficiently small constant, the high-order remainder terms $|\eta_2(\pi^{\tiny{opt}},\pi_{\old}^{(\ell)})|+|\eta_2(\pi_{\new},\pi_{\old}^{(\ell)})|$ can be upper bounded from above by $\epsilon \{\calV(\pi^{\tiny{opt}})-\calV(\pi_{\new})\}+c  \{\calV(\pi^{\tiny{opt}})-\calV(\pi_{\old}^{(\ell)})\}^{(\alpha+2)/(\alpha+1)}$ for some constants $c>0,0<\epsilon\le (1-\gamma)/2$, with probability approaching 1 (w.p.a.1). It follows that \vspace{-0.1cm}
\begin{eqnarray*}
	\calV(\pi^{\tiny{opt}})-\calV(\pi_{\new})-\frac{\epsilon}{1-\gamma} \{\calV(\pi^{\tiny{opt}})-\calV(\pi_{\new})\}\le \frac{c}{\mathbb{L}}\sum_{\ell=1}^{\mathbb{L}}  \{\calV(\pi^{\tiny{opt}})-\calV(\pi_{\old}^{(\ell)})\}^{(\alpha+2)/(\alpha+1)}\\
	+\frac{1}{(1-\gamma)\mathbb{L}}\sum_{\ell=1}^{\mathbb{L}} \{\eta_1(\pi^{\tiny{opt}},\pi_{\old}^{(\ell)})-\eta_1(\pi_{\new},\pi_{\old}^{(\ell)})\}.
\end{eqnarray*}

\vspace{-0.1cm}
\noindent Under the given conditions on the initial policy, we have that
\vspace{-0.1cm}
\begin{eqnarray*}
	\calV(\pi^{\tiny{opt}})-\calV(\pi_{\new})\le  O_p\{(NT)^{-\frac{\alpha+2}{\alpha+1}\kappa_0}\}+\frac{2}{(1-\gamma)\mathbb{L}}\sum_{\ell=1}^{\mathbb{L}} \{\eta_1(\pi^{\tiny{opt}},\pi_{\old}^{(\ell)})-\eta_1(\pi_{\new},\pi_{\old}^{(\ell)})\},
\end{eqnarray*}

\vspace{-0.1cm}
\noindent w.p.a.1. 

It suffices to upper bound $\mathbb{L}^{-1}\sum_{\ell=1}^{\mathbb{L}}\{\eta_1(\pi^{\tiny{opt}},\pi_{\old}^{(\ell)})-\eta_1(\pi_{\new},\pi_{\old}^{(\ell)})\}$. Note that $\pi_{\new}$ is obtained by maximizing $\widehat{\eta}_1(\pi)$, we have $\widehat{\eta}_1(\pi_{\new})\ge \widehat{\eta}_1(\pi^{\tiny{opt}})$. Consequently, $\mathbb{L}^{-1}\sum_{\ell=1}^{\mathbb{L}}\{\eta_1(\pi^{\tiny{opt}},\pi_{\old})-\eta_1(\pi_{\new},\pi_{\old})\}\le \mathbb{L}^{-1}\sum_{\ell=1}^{\mathbb{L}}\{\eta_1(\pi^{\tiny{opt}},\pi_{\old})-\eta_1(\pi_{\new},\pi_{\old})-\widehat{\eta}_1(\pi_{\new}) +\widehat{\eta}_1(\pi^{\tiny{opt}})\}$. Thus, it suffices to provide an upper bound for\vspace{-0.1cm}
\begin{eqnarray*}
	\left|\frac{1}{\mathbb{L}}\sum_{\ell=1}^{\mathbb{L}}\{\eta_1(\pi^{\tiny{opt}},\pi_{\old}^{(\ell)})-\eta_1(\pi_{\new},\pi_{\old}^{(\ell)})-\widehat{\eta}_1(\pi_{\new}) +\widehat{\eta}_1(\pi^{\tiny{opt}})\}\right|.
\end{eqnarray*}

\vspace{-0.1cm}
\noindent By Lemma \ref{lemma:aux1}, the above quantity can be upper bounded by \vspace{-0.1cm}
\begin{eqnarray*}
	&&\left|\frac{1}{\mathbb{L}}\sum_{\ell=1}^{\mathbb{L}}\{\eta_1(\pi^{\tiny{opt}},\pi_{\old}^{(\ell)})-\eta_1(\pi_{\new},\pi_{\old}^{(\ell)})-\widehat{\eta}_1^*(\pi_{\new}) +\widehat{\eta}_1^*(\pi^{\tiny{opt}})\}\right|\\&+&[o_p\{(NT)^{-1/(2+2\alpha)}\}+O_p\{(NT)\}^{\kappa_4/2-1/2}]\mathbb{E}_{S^*\sim d^{\pi^{\tiny{opt}},\nu}}\|\pi^{\tiny{opt}}(\bullet|S^*)-\pi_{\new}(\bullet|S^*)\|_{\TV}.
\end{eqnarray*}

\vspace{-0.1cm}
\noindent In the second step, we show the first term can be upper bounded by \vspace{-0.1cm}
\begin{eqnarray*}
	\EE_{S^*\sim d^{\pi^{\tiny{opt}},\nu}}\calD_{\TV}\{\pi_{\new}(\bullet|S^*),\pi^{\tiny{opt}}(\bullet|S^*)\} O_p\{(NT)^{\kappa_4/2-1/2}\log (NT)\}.
\end{eqnarray*}

\vspace{-0.1cm}
\noindent By Lemma \ref{lemma:aux}, we obtain \vspace{-0.1cm}
\begin{eqnarray*}
	\calV(\pi^{\tiny{opt}})-\calV(\pi_{\new})\le O_p \{(NT)^{-\frac{\alpha+2}{\alpha+1}\kappa_0}\}+\{\calV(\pi^{\tiny{opt}})-\calV(\pi_{\new})\}^{\alpha/(1+\alpha)} (I_1+ I_2),
\end{eqnarray*}

\vspace{-0.1cm}
\noindent where $I_1=O_p\{(NT)^{\kappa_4/2-1/2}\log (NT)\}$ and $I_2=o_p\{(NT)^{-1/(2+2\alpha)}\}$. Using H{\"o}lder's inequality, the second term on the right hand side can be upper bounded by $\{\calV(\pi^{\tiny{opt}})-\calV(\pi_{\new})\}/2+I_1^{(1+\alpha)}+I_2^{(1+\alpha)}$. This yields
%
%
$\calV(\pi^{\tiny{opt}})-\calV(\pi_{\new})=O\{(NT)^{-\kappa_0\frac{\alpha+2}{\alpha+1}}\}+o_p\{(NT)^{-1/2}\}$,  under the given condition on $\kappa_4$. The proof is hence completed. 

In the last three steps, we present the proofs of Lemmas \ref{lemma:aux}, \ref{lemma:aux1} and \ref{lemma:aux2}. We next present the details for each of the step. 

\noindent {\bf{Step 1}. }We aim to bound the high-order remainder term $|\eta_2(\pi^{\tiny{opt}},\pi_{\old}^{(\ell)})|$ and $|\eta_2(\pi_{\new},\pi_{\old}^{(\ell)})|$. We first consider $|\eta_2(\pi^{\tiny{opt}},\pi_{\old}^{(\ell)})|$. 
By definition, we have $\eta_2(\pi^{\tiny{opt}},\pi)= \eta_2^{(1)}(\pi^{\tiny{opt}},\pi)+\eta_2^{(2)}(\pi^{\tiny{opt}},\pi)$ for any $\pi$ where \vspace{-0.1cm}
\begin{eqnarray*}
	&\eta_2^{(1)}(\pi^{\tiny{opt}},\pi)&=\sum_{a\in \calA,s\in \calS} \{\pi^{\scriptsize{\tiny{opt}}}(a|s)-\pi(a|s)\}A^{\pi^{\tiny{opt}}}(a,s) \{d^{\pi^{\scriptsize{\tiny{opt}}},\nu}(s)-d^{\pi,\nu}(s)\},\\
	&\eta_2^{(2)}(\pi^{\tiny{opt}},\pi)&=\sum_{a\in \calA,s\in \calS} \{\pi^{\scriptsize{\tiny{opt}}}(a|s)-\pi(a|s)\}\{A^{\pi}(a,s)-A^{\pi^{\tiny{opt}}}(a,s)\} \{d^{\pi^{\scriptsize{\tiny{opt}}},\nu}(s)-d^{\pi,\nu}(s)\}.
\end{eqnarray*}

\vspace{-0.1cm}
\noindent Consequently, it suffices to bound $|\eta_2^{(j)}(\pi^{\tiny{opt}},\pi_{\old}^{(\ell)})|$ for $j=1,2$. 

We first consider $|\eta_2^{(1)}(\pi^{\tiny{opt}},\pi_{\old}^{(\ell)})|$. Using similar arguments in the proof of Lemma \ref{prop1} (see Equation \ref{eqn:proofeq3}), we can show it is upper bounded by \vspace{-0.1cm}
\begin{eqnarray*}
	O(1) \EE_{S^*\sim d^{\pi^{\scriptsize{opt}},\nu}} \|\pi^{\scriptsize{opt}}(\bullet|S^*)-\pi_{\old}^{(\ell)}(\bullet|S^*) \|_{\TV}  \EE_{S^*\sim d^{\pi_{\old}^{(\ell)},\nu}} \sum_{a\in \calA} |\pi_{\old}^{(\ell)}(a|S^*)-\pi^{\tiny{opt}}(a|S^*)||A^{\pi^{\tiny{opt}}}(a,s)|,
\end{eqnarray*}

\vspace{-0.1cm}
\noindent where $O(1)$ denotes some positive constant. In the proof of Lemma \ref{lemma:aux}, we show that $\{\pi(a|s)-\pi^{\tiny{opt}}(a|s)\}A^{\pi^{\tiny{opt}}}(a,s)$ is nonpositive for any $a$, $s$ and $\pi$. Consequently, $\EE_{S^*\sim d^{\pi,\nu}} \sum_{a\in \calA} |\pi(a|S^*)-\pi^{\tiny{opt}}(a|S^*)||A^{\pi^{\tiny{opt}}}(a,S^*)|=\EE_{S^*\sim d^{\pi,\nu_{\old}}} \sum_{a\in \calA} \{\pi^{\tiny{opt}}(a|S^*)-\pi(a|S^*)\}A^{\pi^{\tiny{opt}}}(a,S^*)=\calV(\pi^{\tiny{opt}})-\calV(\pi)$ for any $\pi$. It follows from Lemma \ref{lemma:aux} that for any $\pi$, \vspace{-0.1cm}
\begin{eqnarray}\label{eqn:c1}
	\begin{split}
		|\eta_2^{(1)}(\pi^{\tiny{opt}},\pi_{\old}^{(\ell)})|\le c\{\calV(\pi^{\tiny{opt}})-\calV(\pi_{\old}^{(\ell)})\}\EE_{S^*\sim d^{\pi^{\scriptsize{opt}},\nu}} \|\pi^{\scriptsize{opt}}(\bullet|S^*)-\pi_{\old}^{(\ell)}(\bullet|S^*) \|_{\TV}\\
		\le c_1\{\calV(\pi^{\tiny{opt}})-\calV(\pi_{\old}^{(\ell)})\}^{\frac{2\alpha+1}{\alpha+1}},
	\end{split}	
\end{eqnarray}

\vspace{-0.1cm}
\noindent for some constant $c_1>0$. 

We next consider $|\eta_2^{(2)}(\pi^{\tiny{opt}},\pi_{\old}^{(\ell)})|$. Note that $A^{\pi}(a,s)-A^{\pi^{\tiny{opt}}}(a,s)=\gamma\EE_{S^*\sim p(\bullet|a,s)} \{V^{\pi}(S^*)-V^{\pi^{\tiny{opt}}}(S^*)\} + V^{\pi^{\tiny{opt}}}(s) - V^{\pi}(s)$ for any $\pi$. Under the given conditions, there exists some constant $C>0$ such that $\nu(s)\ge C$ for any $s$. Using the change of measure, it follows that\vspace{-0.1cm}
\begin{eqnarray}\label{eqn:changeofmeasure}
	\begin{split}
		\EE_{S^*\sim p(\bullet|a,s)} \{V^{\pi^{\tiny{opt}}}(S^*)-V^{\pi}(S^*)\}\le \EE_{S^*\sim \nu} \{V^{\pi^{\tiny{opt}}}(S^*)-V^{\pi}(S^*)\}\frac{p(S^*|a,s)}{\nu (S^*)}\\
		\leq  C^{-1}\EE_{S^*\sim \nu} \{V^{\pi^{\tiny{opt}}}(S^*)-V^{\pi}(S^*)\}=C^{-1} \{\calV(\pi^{\tiny{opt}})-\calV(\pi)\},
	\end{split}
\end{eqnarray}
and $V^{\pi^{\tiny{opt}}}(s) - V^{\pi}(s) \leq C^{-1} \{\calV(\pi^{\tiny{opt}})-\calV(\pi)\}$.

\vspace{-0.1cm}
\noindent Using similar arguments in \eqref{eqn:proofeq3}, we obtain \vspace{-0.1cm}
\begin{eqnarray*}
	|\eta_2^{(2)}(\pi^{\tiny{opt}},\pi)|\le O(1) \EE_{S^*\sim d^{\pi^{\tiny{opt},\nu}}} \sum_{a\in \calA} |\pi(a|S^*)-\pi^{\tiny{opt}}(a|S^*)|\max_{a,s} |A^{\pi^{\tiny{opt}}}(a,s)-A^{\pi}(a,s)|\\
	\le O(1) \{\calV(\pi^{\tiny{opt}})-\calV(\pi)\}\EE_{S^*\sim d^{\pi^{\tiny{opt},\nu}}} \|\pi(\bullet|S^*)-\pi^{\tiny{opt}}(\bullet|S^*)\|_{\TV},
\end{eqnarray*}

\vspace{-0.1cm}
\noindent where $O(1)$ denotes some positive constant. Similar to \eqref{eqn:c1}, we obtain $|\eta_2^{(2)}(\pi^{\tiny{opt}},\pi_{\old}^{(\ell)})|\le c_2 \{\calV(\pi^{\tiny{opt}})-\calV(\pi_{\old}^{(\ell)})\}^{(2\alpha+1)/(\alpha+1)}$ for some constant $c_2>0$. This together with \eqref{eqn:c1} yields \vspace{-0.1cm}
\begin{eqnarray*}
	|\eta_2(\pi^{\tiny{opt}},\pi_{\old}^{(\ell)})|\le (c_1+c_2)\{\calV(\pi^{\tiny{opt}})-\calV(\pi_{\old}^{(\ell)})\}^{\frac{2\alpha+1}{\alpha+1}}.
\end{eqnarray*}

We next bound $|\eta_2(\pi_{\new},\pi_{\old}^{(\ell)})|$. We note that $|\eta_2(\pi_{\new},\pi_{\old}^{(\ell)})|$ can be upper bounded by\vspace{-0.1cm}
\begin{eqnarray*}
	|\eta_2(\pi_{\new},\pi_{\old}^{(\ell)})|\le 	|\eta_2^{(3)}(\pi_{\new},\pi_{\old}^{(\ell)})|+|\eta_2^{(4)}(\pi_{\new},\pi_{\old}^{(\ell)})|
	\\\equiv \sum_{a,s} |\pi_{\new}(a|s)-\pi_{\old}^{(\ell)}(a|s)||A^{\pi^{\tiny{opt}}}(a,s)-A^{\pi_{\old}^{(\ell)}}(a,s)| |d^{\pi_{\new},\nu}(s)-d^{\pi_{\old}^{(\ell)},\nu}(s)|\\
	+\sum_{a,s} |\pi_{\new}(a|s)-\pi_{\old}^{(\ell)}(a|s)||A^{\pi^{\tiny{opt}}}(a,s)| |d^{\pi_{\new},\nu}(s)-d^{\pi_{\old}^{(\ell)},\nu}(s)|.
\end{eqnarray*}

\vspace{-0.1cm}
\noindent Using similar arguments in bounding $|\eta_2^{(2)}(\pi^{\tiny{opt}},\pi_{\old}^{(\ell)})|$, $|\eta_2^{(3)}(\pi_{\new},\pi_{\old}^{(\ell)})|$ can be upper bounded by $O(1) \{\calV(\pi^{\tiny{opt}})-\calV(\pi_{\old}^{(\ell)})\} \EE_{S^*\sim d^{\pi^{\tiny{opt}},\nu}}\|\pi_{\new}(a|S^*)-\pi_{\old}^{(\ell)}(a|S^*)\|_{\TV}$ where $O(1)$ denotes some positive constant. By triangle inequality, we have \vspace{-0.1cm}
\begin{eqnarray*}
	\EE_{S^*\sim d^{\pi^{\tiny{opt}},\nu}}\|\pi_{\new}(a|S^*)-\pi_{\old}^{(\ell)}(a|S^*)\|_{\TV}\le \EE_{S^*\sim d^{\pi^{\tiny{opt}},\nu}}\|\pi^{\tiny{opt}}(a|S^*)-\pi_{\old}^{(\ell)}(a|S^*)\|_{\TV}\\+\EE_{S^*\sim d^{\pi^{\tiny{opt}},\nu}}\|\pi^{\tiny{opt}}(a|S^*)-\pi_{\new}(a|S^*)\|_{\TV}.
\end{eqnarray*}

\vspace{-0.1cm}
\noindent By Lemma \ref{lemma:aux}, the two terms on the RHS can be upper bounded by $c_0^{-1} \{\calV(\pi^{\tiny{opt}})-\calV(\pi_{\old}^{(\ell)})\}^{\alpha/(1+\alpha)}$ and $c_0^{-1} \{\calV(\pi^{\tiny{opt}})-\calV(\pi_{\new})\}^{\alpha/(1+\alpha)}$, respectively. Consequently, \vspace{-0.1cm}
\begin{eqnarray*}
	|\eta_2^{(3)}(\pi_{\new},\pi_{\old}^{(\ell)})|\le O(1) \{\calV(\pi^{\tiny{opt}})-\calV(\pi_{\old}^{(\ell)})\}^{\frac{2\alpha+1}{\alpha+1}}\\+O(1) \{\calV(\pi^{\tiny{opt}})-\calV(\pi_{\new})\}^{\frac{\alpha}{\alpha+1}}\{\calV(\pi^{\tiny{opt}})-\calV(\pi_{\old}^{(\ell)})\},
\end{eqnarray*}

\vspace{-0.1cm}
\noindent for some positive constant $O(1)$. 

Similarly, $\eta_2^{(4)}(\pi_{\new},\pi_{\old}^{(\ell)})$ can be upper bounded by \vspace{-0.1cm}
\begin{eqnarray*}
	|\eta_2^{(4)}(\pi_{\new},\pi_{\old}^{(\ell)})|\le \sum_{a,s} |\pi_{\new}(a|s)-\pi^{\tiny{opt}}(a|s)||A^{\pi^{\tiny{opt}}}(a,s)| |d^{\pi_{\new},\nu}(s)-d^{\pi_{\old}^{(\ell)},\nu}(s)|\\
	+\sum_{a,s} |\pi_{\old}^{(\ell)}(a|s)-\pi^{\tiny{opt}}(a|s)||A^{\pi^{\tiny{opt}}}(a,s)| |d^{\tiny{opt},\nu}(s)-d^{\pi_{\old}^{(\ell)},\nu}(s)|\\
	+\sum_{a,s} |\pi_{\old}^{(\ell)}(a|s)-\pi^{\tiny{opt}}(a|s)||A^{\pi^{\tiny{opt}}}(a,s)| |d^{\pi_{\new},\nu}(s)-d^{\tiny{opt},\nu}(s)|.
\end{eqnarray*}

\vspace{-0.1cm}
\noindent Using similar arguments in the proofs of Lemma \ref{prop1} and Theorem \ref{thm3}, the first term on the RHS can be upper bounded by $O(\sqrt{\delta} \{\calV(\pi^{\tiny{opt}})-\calV(\pi_{\new})\})$. The second and third terms can be upper bounded by\vspace{-0.1cm}
\begin{eqnarray*}
	O(1) \{\calV(\pi^{\tiny{opt}})-\calV(\pi_{\old}^{(\ell)})\}^{\frac{2\alpha+1}{\alpha+1}}+O(1) \{\calV(\pi^{\tiny{opt}})-\calV(\pi_{\new})\}^{\frac{\alpha}{\alpha+1}}\{\calV(\pi^{\tiny{opt}})-\calV(\pi_{\old}^{(\ell)})\},
\end{eqnarray*}

\vspace{-0.1cm}
\noindent using similar arguments in bounding $|\eta_2^{(3)}(\pi_{\new},\pi_{\old}^{(\ell)})|$. 

To summarize, we have shown \vspace{-0.1cm}
\begin{eqnarray*}
	|\eta_2(\pi_{\new},\pi_{\old}^{(\ell)})|\le O(1) \sqrt{\delta} \{\calV(\pi^{\tiny{opt}})-\calV(\pi_{\new})\}+O(1) \{\calV(\pi^{\tiny{opt}})-\calV(\pi_{\old}^{(\ell)})\}^{\frac{2\alpha+1}{\alpha+1}}\\+O(1) \{\calV(\pi^{\tiny{opt}})-\calV(\pi_{\new})\}^{\frac{\alpha}{\alpha+1}}\{\calV(\pi^{\tiny{opt}})-\calV(\pi_{\old}^{(\ell)})\},
\end{eqnarray*} 

\vspace{-0.1cm}
\noindent for some positive constant $O(1)$. By H{\"o}lder's inequality, the last term on the second line can be upper bounded by $\sqrt{\delta} \{\calV(\pi^{\tiny{opt}})-\calV(\pi_{\new})\}+O(1) \{\calV(\pi^{\tiny{opt}})-\calV(\pi_{\old}^{(\ell)})\}^{\alpha+1}/\sqrt{\delta}$. When $\calV(\pi_{\old}^{(\ell)})$ is consistent to $\calV(\pi^{\tiny{opt}})$, $\{\calV(\pi^{\tiny{opt}})-\calV(\pi_{\old}^{(\ell)})\}^{\alpha+1}/\sqrt{\delta}=\{\calV(\pi^{\tiny{opt}})-\calV(\pi_{\old}^{(\ell)})\}^{\frac{2\alpha+1}{\alpha+1}}/\sqrt{\delta}$. It follows that
\begin{eqnarray*}
	|\eta_2(\pi_{\new},\pi_{\old}^{(\ell)})|\le O(1) \sqrt{\delta} \{\calV(\pi^{\tiny{opt}})-\calV(\pi_{\new})\}+O(1) \{\calV(\pi^{\tiny{opt}})-\calV(\pi_{\old}^{(\ell)})\}^{\frac{2\alpha+1}{\alpha+1}}.
\end{eqnarray*}	
The proof is hence completed. 

\noindent {\bf{Step 2}. }We begin with some notations. 
For $\ell=1,\cdots,\mathbb{L}$, we use $\psi(o;\pi,\pi_{\old}^{(\ell)},\widehat{Q}^{(\ell)},\widehat{\omega}^{(\ell)},\widehat{d}^{(\ell)})$ to denote $\psi(o;\pi,\pi_{\old}^{(\ell)},\widehat{V}^{(\ell)},\widehat{A}^{(\ell)},\widehat{\omega}^{(\ell)},\widehat{d}^{(\ell)})$ as both $\widehat{V}^{(\ell)}$ and $\widehat{A}^{(\ell)}$ are derived from $\widehat{Q}^{(\ell)}$. 
Similarly, we use the notations $\psi_1(\pi,\pi_{\old}^{(\ell)},\widehat{Q}^{(\ell)},\widehat{d}^{(\ell)})$, $\psi_2(o;\pi,\pi_{\old}^{(\ell)},\widehat{Q}^{(\ell)},\widehat{\omega}^{(\ell)},\widehat{d}^{(\ell)})$ and \\$\psi_3(o;\pi,\pi_{\old}^{(\ell)},\widehat{Q}^{(\ell)},\widehat{\omega}^{(\ell)},\widehat{d}^{(\ell)})$ to denote $\psi_1(\pi,\pi_{\old}^{(\ell)},\widehat{A}^{(\ell)},\widehat{d}^{(\ell)})$, $\psi_2(o;\pi,\pi_{\old}^{(\ell)},\widehat{V}^{(\ell)},\widehat{A}^{(\ell)},\widehat{\omega}^{(\ell)},\widehat{d}^{(\ell)})$ and $\psi_3(o;\pi,\pi_{\old}^{(\ell)},\widehat{A}^{(\ell)},\widehat{\omega}^{(\ell)},\widehat{d}^{(\ell)})$. Notations $\psi_1(\pi,\pi_{\old}^{(\ell)},Q^{\pi_{\old}^{(\ell)}},d^{\pi_{\old}^{(\ell)}})$, $\psi_2(o;\pi,\pi_{\old}^{(\ell)},Q^{\pi_{\old}^{(\ell)}},\omega^{\pi_{\old}^{(\ell)}},d^{\pi_{\old}^{(\ell)}})$,  $\psi_3(o;\pi,\pi_{\old}^{(\ell)},Q^{\pi_{\old}^{(\ell)}},\omega^{\pi_{\old}^{(\ell)}},d^{\pi_{\old}^{(\ell)}})$ and $\psi(o;\pi,\pi_{\old}^{(\ell)},Q^{\pi_{\old}^{(\ell)}},\omega^{\pi_{\old}^{(\ell)}},d^{\pi_{\old}^{(\ell)}})$ can be similarly defined. 

We aim to apply Lemma \ref{lemma:aux2} to show \vspace{-0.1cm}
\begin{eqnarray*}
	\sup_{\pi\in \Pi} \frac{|\eta_1(\pi^{\tiny{opt}},\pi_{\old}^{(\ell)})-\eta_1(\pi,\pi_{\old}^{(\ell)})+\widehat{\eta}_1^{*}(\pi,\pi_{\old}^{(\ell)}) -\widehat{\eta}_1^{*}(\pi^{\tiny{opt}},\pi_{\old}^{(\ell)})|}{\EE_{S^*\sim d^{\pi^{\tiny{opt}}}}\|\pi(\bullet|S^*)-\pi_{\old}^{(\ell)}(\bullet|S^*)\|_{\TV}}=O_p\{ (NT)^{\frac{\kappa_4-1}{2}} \log (NT)\},
\end{eqnarray*}
where $\widehat{\eta}_1^{*}(\pi,\pi_{\old}^{(\ell)})=(|\mathcal{I}_{\ell}|T)^{-1} \sum_{i\in \mathcal{I}_{\ell}}\sum_{t=0}^{T} \psi(O_{i,t};\pi,\pi_{\old}^{(\ell)},Q^{\pi_{\old}^{(\ell)}},\omega^{\pi_{\old}^{(\ell)}},d^{\pi_{\old}^{(\ell)}})$. 
Toward that end, we first note that it follows from (C6) that $\{(S_t,A_t,R_t,S_{t+1})\}_{t\ge 0}$ is exponentially $\beta$-mixing. Since the observed data set consists of multiple independent trajectories, it is exponentially  $\beta$-mixing as well. Consequently, by setting the integer $q$ in Lemma \ref{lemma:aux2} to $c \log (NT)$ for some sufficiently large constant $c$, it follows that $NT \beta(q)/q=o(1)$. 

Under the given conditions, the immediate reward and the conditional probability ratio are uniformly bounded. By (C4), $\Pi$ belongs to a VC type function class with bounded envelop function and VC index bounded by $O\{(NT)^{\kappa_4}\}$. So is the class of functions $\{\psi(o;\pi,\pi_{\old}^{(\ell)},Q^{\pi_{\old}^{(\ell)}},\omega^{\pi_{\old}^{(\ell)}},d^{\pi_{\old}^{(\ell)}})-\psi(o;\pi,\pi^{\tiny{opt}},Q^{\pi_{\old}^{(\ell)}},\omega^{\pi_{\old}^{(\ell)}},d^{\pi_{\old}^{(\ell)}}) :\pi\in \Pi \}$. In view of Lemma \ref{lemma:aux2}, it suffices to show
\begin{eqnarray*}
	\Var\{\psi(O_0;\pi,\pi_{\old}^{(\ell)},Q^{\pi_{\old}^{(\ell)}},\omega^{\pi_{\old}^{(\ell)}},d^{\pi_{\old}^{(\ell)}})-\psi(O_0;\pi,\pi^{\tiny{opt}},Q^{\pi_{\old}^{(\ell)}},\omega^{\pi_{\old}^{(\ell)}},d^{\pi_{\old}^{(\ell)}})\}\\\le c\{\EE_{S^*\sim d^{\pi^{\tiny{opt}}}}\|\pi(\bullet|S^*)-\pi_{\old}^{(\ell)}(\bullet|S^*)\|_{\TV}\}^2,
\end{eqnarray*} 
and 
\begin{eqnarray*}
	|\psi(O_0;\pi,\pi_{\old}^{(\ell)},Q^{\pi_{\old}^{(\ell)}},\omega^{\pi_{\old}^{(\ell)}},d^{\pi_{\old}^{(\ell)}})-\psi(O_0;\pi,\pi^{\tiny{opt}^{(\ell)}},Q^{\pi_{\old}^{(\ell)}},\omega^{\pi_{\old}^{(\ell)}},d^{\pi_{\old}^{(\ell)}})|\\\le c\EE_{S^*\sim d^{\pi^{\tiny{opt}}}}\|\pi(\bullet|S^*)-\pi_{\old}^{(\ell)}(\bullet|S^*)\|_{\TV}, 
\end{eqnarray*}
almost surely for some constant $c>0$. This is immediate to see by the definition of $\psi$. We omit the details for brevity. 

\noindent {\bf{Step 3}. }We prove Lemma \ref{lemma:aux} in this step.  
By \eqref{eqn:key}, we obtain \vspace{-0.1cm}
\begin{eqnarray}\label{eqn:proofeqA1}
	\begin{split}
		\calV(\pi^{\tiny{opt}})-\calV(\pi)=-\frac{1}{1-\gamma}\EE_{S^* \sim d^{\pi,\nu}}\sum_{a\in \calA} \pi(a|S^*)A^{\pi^{\tiny{opt}}}(a,S^*) \\
		=-\frac{1}{1-\gamma}\EE_{S^* \sim d^{\pi,\nu}}\sum_{a\in \calA} \{\pi(a|S^*)-\pi^{\tiny{opt}}(a|S^*)\}A^{\pi^{\tiny{opt}}}(a,S^*).
	\end{split}
\end{eqnarray}

\vspace{-0.1cm}
\noindent As discussed in Section \ref{sec:theory2}, $A^{\pi^{\tiny{opt}}}(a,s)\le 0$ for any $a$ and $s$. We next claim \vspace{-0.1cm}
\begin{eqnarray}\label{eqn:claim}
	\{\pi(a|s)-\pi^{\tiny{opt}}(a|s)\}A^{\pi^{\tiny{opt}}}(a,s)\le 0,\,\,\,\,\forall a,s.
\end{eqnarray}

\vspace{-0.1cm}
\noindent If $a=\argmax_{a'} Q^{\pi^{\tiny{opt}}}(a',s)$, then $A^{\pi^{\tiny{opt}}}(a,s)=0$ and \eqref{eqn:claim} is automatically satisfied. Otherwise, we have $A^{\pi^{\tiny{opt}}}(a,s)<0$ and $\pi^{\tiny{opt}}(a|s)=0$. It follows that \eqref{eqn:claim} holds. 

Combining \eqref{eqn:proofeqA1} with \eqref{eqn:claim} yields that  \vspace{-0.1cm}
\begin{eqnarray*}
	\begin{split}
		\calV(\pi^{\tiny{opt}})-\calV(\pi)
		=\frac{1}{1-\gamma}\EE_{S^* \sim d^{\pi,\nu}}\sum_{a\in \calA} |\pi(a|S^*)-\pi^{\tiny{opt}}(a|S^*)||A^{\pi^{\tiny{opt}}}(a,S^*)| .
	\end{split}
\end{eqnarray*}

\vspace{-0.1cm}
\noindent Since $d^{\pi,\nu}(s)\ge (1-\gamma)\nu(s)$ and $\nu$ is uniformly bounded away from zero, there exists some constant $C>0$ such that $d^{\pi,\nu}(s)\ge (1-\gamma)C$ for any $s$. Similar to \eqref{eqn:changeofmeasure}, we can show\vspace{-0.1cm}
\begin{eqnarray*}
	\begin{split}
		\calV(\pi^{\tiny{opt}})-\calV(\pi)
		=\frac{1}{1-\gamma}\EE_{S^* \sim d^{\pi,\nu}}\sum_{a\in \calA} |\pi(a|S^*)-\pi^{\tiny{opt}}(a|S^*)||A^{\pi^{\tiny{opt}}}(a,S^*)|\frac{d^{\pi,\nu}(S^*)}{d^{\pi^{\tiny{opt}},\nu}(S^*)} \\ \ge C\EE_{S^* \sim d^{\pi^{\tiny{opt}},\nu}}\sum_{a\in \calA} |\pi(a|S^*)-\pi^{\tiny{opt}}(a|S^*)||A^{\pi^{\tiny{opt}}}(a,S^*)|.
	\end{split}
\end{eqnarray*}

\vspace{-0.1cm}
\noindent Denote $a^* = \argmax_{a'} Q^{\pi^{\tiny{opt}}}(a',S^*)$. Then $A^{\pi^{\tiny{opt}}}(a^*,S^*)=0$. For any $\epsilon>0$, it follows that \vspace{-0.1cm}
\begin{eqnarray*}
	\calV(\pi^{\tiny{opt}})-\calV(\pi)\ge C\EE_{S^* \sim d^{\pi^{\tiny{opt}},\nu}}\sum_{a\in \calA, a \neq a^*} |\pi(a|S^*)-\pi^{\tiny{opt}}(a|S^*)||A^{\pi^{\tiny{opt}}}(a,S^*)|\mathbb{I}(A^{\pi^{\tiny{opt}}}(a,S^*)\le -\epsilon)\\
	\ge \epsilon C\EE_{S^* \sim d^{\pi^{\tiny{opt}},\nu}}\sum_{a\in \calA, a \neq a^*} |\pi(a|S^*)-\pi^{\tiny{opt}}(a|S^*)|\mathbb{I}(A^{\pi^{\tiny{opt}}}(a,S^*)\le -\epsilon)\\\ge 
	\epsilon C \EE_{S^* \sim d^{\pi^{\tiny{opt}},\nu}}\sum_{a\in \calA, a \neq a^*} |\pi(a|S^*)-\pi^{\tiny{opt}}(a|S^*)|\\
	-\epsilon C \EE_{S^* \sim d^{\pi^{\tiny{opt}},\nu}}\sum_{a\in \calA, a \neq a^*} |\pi(a|S^*)-\pi^{\tiny{opt}}(a|S^*)|\mathbb{I}(-\epsilon<A^{\pi^{\tiny{opt}}}(a,S^*)<0).
\end{eqnarray*}

\vspace{-0.1cm}
\noindent Under the margin-type condition in (C5), the last line is $O(\epsilon^{\alpha+1})$. We choose\vspace{-0.1cm}
\begin{eqnarray*} 
	\epsilon=c\{\EE_{S^* \sim d^{\pi^{\tiny{opt}},\nu}}\sum_{a\in \calA, a \neq a^*} |\pi(a|S^*)-\pi^{\tiny{opt}}(a|S^*)|\}^{1/\alpha},
\end{eqnarray*}
for some constant $c>0$. With some property choice of $c$, the last line is lower bounded by $\epsilon C\EE_{S^* \sim d^{\pi^{\tiny{opt}},\nu}}\sum_{a\in \calA, a \neq a^*} |\pi(a|S^*)-\pi^{\tiny{opt}}(a|S^*)|/2$. Note that $$
\sum_{a\in \calA, a \neq a^*} |\pi(a|S^*)-\pi^{\tiny{opt}}(a|S^*)| = |\pi(a^*|S^*)-\pi^{\tiny{opt}}(a^*|S^*)|
$$ 
Consequently, \vspace{-0.1cm}
\begin{eqnarray*}
	\calV(\pi^{\tiny{opt}})-\calV(\pi)\ge \frac{c C}{4} \{\EE_{S^* \sim d^{\pi^{\tiny{opt}},\nu}}\sum_{a\in \calA} |\pi(a|S^*)-\pi^{\tiny{opt}}(a|S^*)|\}^{1+1/\alpha}.
\end{eqnarray*}

\vspace{-0.1cm}
\noindent The proof is completed by noting that $\sum_{a\in \calA} |\pi(a|S^*)-\pi^{\tiny{opt}}(a|S^*)|=2\|\pi(\bullet|S^*)-\pi^{\tiny{opt}}(\bullet|S^*)\|_{\TV}$.

\noindent \textbf{Step 4.} We prove Lemma \ref{lemma:aux1} in this step. 
%
We first show the bias term \vspace{-0.1cm} 
\begin{eqnarray*}
	\Delta \psi(\pi_1,\pi_2,\pi_{\old}^{(\ell)},\widehat{Q}^{(\ell)},\widehat{\omega}^{(\ell)},\widehat{d}^{(\ell)})=\EE [\psi(O_0;\pi_1,\pi_{\old},\widehat{Q}^{(\ell)},\widehat{\omega}^{(\ell)},\widehat{d}^{(\ell)})|\{O_{i,t}\}_{i\in \mathcal{I}_{\ell},0\le t<T}]\\-\EE [\psi(O_0;\pi_2,\pi_{\old},\widehat{Q}^{(\ell)},\widehat{\omega}^{(\ell)},\widehat{d}^{(\ell)})|\{O_{i,t}\}_{i\in \mathcal{I}_{\ell},0\le t<T}]-\eta_1(\pi_1,\pi_{\old}^{(\ell)})+\eta_1(\pi_2,\pi_{\old}^{(\ell)})
\end{eqnarray*}

\vspace{-0.1cm}
\noindent where the little-o term is uniform in $\pi_1,\pi_2$. We next show $\sup_{\pi_1,\pi_2\in \Pi} |\widehat{\eta}^{(\ell)}(\pi_1)-\widehat{\eta}^{*}(\pi_1,\pi_{\old}^{(\ell)})-\widehat{\eta}^{(\ell)}(\pi_2)+\widehat{\eta}^{*}(\pi_2,\pi_{\old}^{(\ell)})-\Delta \psi(\pi_1,\pi_2,\pi_{\old}^{(\ell)})|=o_p\{(NT)^{\kappa_4/2-1/2}\}$. The proof is hence completed. 

To bound the bias term, we first observe that when $\widehat{\omega}^{(\ell)}=\omega^{\pi_{\old}^{(\ell)}}$, \vspace{-0.1cm}
\begin{eqnarray*}
	&&\psi_1(\pi,\pi_{\old}^{(\ell)},\widehat{Q}^{(\ell)},\widehat{d}^{(\ell)})+\EE \{\psi_2(O_0;\pi,\pi_{\old}^{(\ell)},\widehat{Q}^{(\ell)},\omega^{\pi_{\old}^{(\ell)}},\widehat{d}^{(\ell)})|\widehat{Q}^{(\ell)},\widehat{d}^{(\ell)}\}=\psi_1(\pi,\pi_{\old}^{(\ell)},Q^{\pi_{\old}^{(\ell)}},\widehat{d}^{(\ell)}),\\
	&&\psi_1(\pi,\pi_{\old}^{(\ell)},\widehat{Q}^{(\ell)},\widehat{d}^{(\ell)})+\EE \{\psi_3(O_0;\pi,\pi_{\old}^{(\ell)},\widehat{Q}^{(\ell)},\omega^{\pi_{\old}^{(\ell)}},\widehat{d}^{(\ell)})|\widehat{Q}^{(\ell)},\widehat{d}^{(\ell)}\}=\psi_1(\pi,\pi_{\old}^{(\ell)},\widehat{Q}^{(\ell)},d^{\pi_{\old}^{(\ell)}}).
\end{eqnarray*}

\vspace{-0.1cm}
\noindent These two assertions can be proven using similar arguments in Section \ref{sec:moreprop2}. Consequently, we have \vspace{-0.1cm}
\begin{eqnarray*}
	\Delta \psi(\pi_1,\pi_2,\pi_{\old}^{(\ell)},\widehat{Q}^{(\ell)},\omega^{\pi_{\old}^{(\ell)}},\widehat{d}^{(\ell)})=\psi_1(\pi_1,\pi_{\old}^{(\ell)},\widehat{Q}^{(\ell)},d^{\pi_{\old}^{(\ell)}})+\psi_1(\pi_1,\pi_{\old}^{(\ell)},Q^{\pi_{\old}^{(\ell)}},\widehat{d}^{(\ell)})\\-\psi_1(\pi_1,\pi_{\old}^{(\ell)},Q^{\pi_{\old}^{(\ell)}},d^{\pi_{\old}^{(\ell)}})
	-\psi_1(\pi_1,\pi_{\old}^{(\ell)},Q^{\pi_{\old}^{(\ell)}},d^{\pi_{\old}^{(\ell)}})-\psi_1(\pi_2,\pi_{\old}^{(\ell)},\widehat{Q}^{(\ell)},d^{\pi_{\old}^{(\ell)}})\\-\psi_1(\pi_2,\pi_{\old}^{(\ell)},Q^{\pi_{\old}^{(\ell)}},\widehat{d}^{(\ell)})+\psi_1(\pi_2,\pi_{\old}^{(\ell)},Q^{\pi_{\old}^{(\ell)}},d^{\pi_{\old}^{(\ell)}})+\psi_1(\pi_2,\pi_{\old}^{(\ell)},Q^{\pi_{\old}^{(\ell)}},d^{\pi_{\old}^{(\ell)}})\\=\sum_{s,a} \{\pi_1(a|s)-\pi_2(a|s)\} \{\widehat{Q}^{(\ell)}(a,s)-Q^{\pi_{\old}^{(\ell)}}(a,s)\}\{d^{\pi_{\old,\nu}}(s)-\widehat{d}^{(\ell),\nu}(s)\}.
\end{eqnarray*}

\vspace{-0.1cm}
\noindent By Cauchy-Schwarz inequality, the RHS can be upper bounded by \vspace{-0.1cm}
\begin{eqnarray}\label{eqn:firstline}
	&&\sqrt{\sum_{s,a} |\pi_1(a|s)-\pi_2(a|s)||\widehat{Q}^{(\ell)}(a,s)-Q^{\pi_{\old}^{(\ell)}}(a,s)|^2p_{\infty}(a,s) }\\\label{eqn:secondline}
	&\times&  \sqrt{\sum_{s,a} |\pi_1(a|s)-\pi_2(a|s)||\widehat{d}^{(\ell),\nu}(s)-d^{\pi_{\old,\nu}^{(\ell)}}(s)|^2/p_{\infty}(a,s)}.
\end{eqnarray}

\vspace{-0.1cm}
\noindent Note that $\max_{a,s} |\pi_1(a|s)-\pi_2(a|s)|\le \sum_{a,s} |\pi_1(a|s)-\pi_2(a|s)|\le C \sum_a \mathbb{E}_{S^*\sim d^{\pi^{\tiny{opt}},\nu}} |\pi_1(a|S^*)-\pi_2(a|S^*)|$ under the assumption that $\nu$ is uniformly bounded away from zero. Under (C1), \eqref{eqn:firstline} is upper bounded by $O_p\{(NT)^{-\kappa_1}\}\mathbb{E}_{S^*\sim d^{\pi^{\tiny{opt}},\nu}}\|\pi_1(\bullet|S^*)-\pi_2(\bullet|S^*)\|_{\TV}$. Since $p_{\infty}$ is uniformly bounded away from zero, \eqref{eqn:secondline} can be upper bounded by $$c\mathbb{E}_{S^*\sim d^{\pi^{\tiny{opt}},\nu}} |\pi_1(a|S^*)-\pi_2(a|S^*)|\sqrt{\max_{s} |\widehat{d}^{(\ell),\nu}(s)-d^{\pi_{\old}^{(\ell)},\nu}(s)|^2 },$$ for some constant $c>0$. Note that $\widehat{d}^{(\ell),\nu}(s)-d^{\pi_{\old}^{(\ell)},\nu}(s)$ can be represented by $\EE_{S^*\sim \widehat{d}^{(\ell),\nu}}\mathbb{I}(S^*=s)-\EE_{S^*\sim d^{\pi_{\old}^{(\ell)},\nu}}\mathbb{I}(S^*=s)=\sum_{a^*,s^*} \pi_{\old}^{(\ell)}(a^*|s^*)\nu(s^*)(\EE_{S^*\sim \widehat{d}^{(\ell)}(\bullet|a^*,s^*)}\mathbb{I}(S^*=s)-\EE_{S^*\sim d^{\pi_{\old}^{(\ell)}}(\bullet|a^*,s^*)}\mathbb{I}(S^*=s))$. 
As such, it can be upper bounded by $$ O(1)\mathbb{E}_{S^*\sim d^{\pi^{\tiny{opt}},\nu}} |\pi_1(a|S^*)-\pi_2(a|S^*)|\sum_a \EE_{S_0\sim \nu} \pi_{\old}^{(\ell)}(a|S_0) \calD_{\TV}\{d^{\pi_{\old}^{(\ell)}}(\bullet|a,S_0),\widehat{d}^{(\ell)}(\bullet|a,S_0)\},$$ 
where $O(1)$ denotes some positive constant. Since $b$ is uniformly bounded away from zero, it can be further upper bounded by \vspace{-0.1cm}
\begin{eqnarray*}
	O\left[\mathbb{E}_{S^*\sim d^{\pi^{\tiny{opt}},\nu}} |\pi_1(a|S^*)-\pi_2(a|S^*)|\EE_{(A,S)\sim p_{\infty}} \calD_{\TV} \{d^{\pi_{\old}^{(\ell)}}(\bullet|A,S),\widehat{d}^{(\ell)}(\bullet|A,S)\}\right].
\end{eqnarray*}

\vspace{-0.1cm}
\noindent Consequently, \eqref{eqn:secondline} is $O_p\{(NT)^{-\kappa_3}\}\mathbb{E}_{S^*\sim d^{\pi^{\tiny{opt}},\nu}} |\pi_1(a|S^*)-\pi_2(a|S^*)|$ under (C3). Under the conditions that $\kappa_1+\kappa_3>1/(2+2\alpha)$, we obtain $	\Delta \psi(\pi_1,\pi_2,\pi_{\old}^{(\ell)},\widehat{Q}^{(\ell)},\omega^{\pi_{\old}^{(\ell)}},\widehat{d}^{(\ell)})=o_p\{(NT)^{-1/(2+2\alpha)}\}\mathbb{E}_{S^*\sim d^{\pi^{\tiny{opt}},\nu}} |\pi_1(a|S^*)-\pi_2(a|S^*)|$. 

It suffices to bound $\Delta \psi(\pi_1,\pi_2,\pi_{\old}^{(\ell)},\widehat{Q}^{(\ell)},\omega^{\pi_{\old}^{(\ell)}},\widehat{d}^{(\ell)})-\Delta \psi(\pi_1,\pi_2,\pi_{\old}^{(\ell)},\widehat{Q}^{(\ell)},\widehat{\omega}^{(\ell)},\widehat{d}^{(\ell)})$, or equivalently, \vspace{-0.1cm}
\begin{eqnarray}\label{thirdline}
	\begin{split}
		\EE \{\psi_2(O_0;\pi_1,\pi_{\old}^{(\ell)},\widehat{Q}^{(\ell)},\omega^{\pi_{\old}^{(\ell)}},\widehat{d}^{(\ell)})-\psi_2(O_0;\pi_1,\pi_{\old}^{(\ell)},\widehat{Q}^{(\ell)},\widehat{\omega}^{(\ell)},\widehat{d}^{(\ell)})\}\\
		- \EE \{\psi_2(O_0;\pi_2,\pi_{\old}^{(\ell)},\widehat{Q}^{(\ell)},\omega^{\pi_{\old}^{(\ell)}},\widehat{d}^{(\ell)})-\psi_2(O_0;\pi_2,\pi_{\old}^{(\ell)},\widehat{Q}^{(\ell)},\widehat{\omega}^{(\ell)},\widehat{d}^{(\ell)})\}
	\end{split}	\\
	\begin{split}\label{fourthline}
		\EE \{\psi_3(O_0;\pi_1,\pi_{\old}^{(\ell)},\widehat{Q}^{(\ell)},\omega^{\pi_{\old}^{(\ell)}},\widehat{d}^{(\ell)})-\psi_3(O_0;\pi_1,\pi_{\old}^{(\ell)},\widehat{Q}^{(\ell)},\widehat{\omega}^{(\ell)},\widehat{d}^{(\ell)})\}\\
		-\EE \{\psi_3(O_0;\pi_2,\pi_{\old}^{(\ell)},\widehat{Q}^{(\ell)},\omega^{\pi_{\old}^{(\ell)}},\widehat{d}^{(\ell)})-\psi_3(O_0;\pi_2,\pi_{\old}^{(\ell)},\widehat{Q}^{(\ell)},\widehat{\omega}^{(\ell)},\widehat{d}^{(\ell)})\}.
	\end{split}
\end{eqnarray}

\vspace{-0.1cm}
\noindent Similarly, we can show \eqref{thirdline} and \eqref{fourthline} can be upper bounded by \vspace{-0.1cm}
\begin{eqnarray*}
	C\sqrt{ \EE_{(A,S)\sim p_{\infty}} |\widehat{Q}^{(\ell)}(A,S)-Q^{\pi_{\old}^{(\ell)}}(A,S)|^2} \sqrt{\EE_{\substack{(A,S)\sim p_{\infty}\\ (\widetilde{A},\widetilde{S}\sim p_{\infty}) }} |\widehat{\omega}^{(\ell)}(\widetilde{A},\widetilde{S};A,S)-\omega^{\pi_{\old}^{(\ell)}}(\widetilde{A},\widetilde{S};A,S)|^2 }\\
	\times \mathbb{E}_{S^*\sim d^{\pi^{\tiny{opt}},\nu}}\|\pi_1(\bullet|S^*)-\pi_2(\bullet|S^*)\|_{\TV},
\end{eqnarray*}

\vspace{-0.1cm}
\noindent and \vspace{-0.1cm}
\begin{eqnarray*}
	C\sqrt{ \EE_{(A,S)\sim p_{\infty}} \calD_{\TV}^2\{d^{\pi_{\old}^{(\ell)}}(\bullet|A,S),\widehat{d}^{(\ell)}(\bullet|A,S)\}} \sqrt{\EE_{\substack{(A,S)\sim p_{\infty}\\ (\widetilde{A},\widetilde{S})\sim p_{\infty} }} |\widehat{\omega}^{(\ell)}(\widetilde{A},\widetilde{S};A,S)-\omega^{\pi_{\old}^{(\ell)}}(\widetilde{A},\widetilde{S};A,S)|^2 }\\
	\times \mathbb{E}_{S^*\sim d^{\pi^{\tiny{opt}},\nu}}\|\pi_1(\bullet|S^*)-\pi_2(\bullet|S^*)\|_{\TV},
\end{eqnarray*}

\vspace{-0.1cm}
\noindent for some constant $C>0$. Under (C2) and (C3), we obtain $\Delta \psi(\pi_1,\pi_2,\pi_{\old}^{(\ell)},\widehat{Q}^{(\ell)},\omega^{\pi_{\old}^{(\ell)}},\widehat{d}^{(\ell)})-\Delta \psi(\pi_1,\pi_2,\pi_{\old}^{(\ell)},\widehat{Q}^{(\ell)},\widehat{\omega}^{(\ell)},\widehat{d}^{(\ell)})=o_p\{(NT)^{-1/(2+2\alpha)}\}$. 

It remains to show $\sup_{\pi_1,\pi_2\in \Pi} |\widehat{\eta}^{(\ell)}(\pi_1)-\widehat{\eta}^{*}(\pi_1,\pi_{\old}^{(\ell)})-\widehat{\eta}^{(\ell)}(\pi_2)+\widehat{\eta}^{*}(\pi_2,\pi_{\old}^{(\ell)})-\Delta \psi(\pi_1,\pi_2,\pi_{\old}^{(\ell)})|=o_p\{(NT)^{\kappa_4/2-1/2}\}$. This can be proven using similar arguments in Step 2. We omit the details to save space. 

\noindent {\bf{Step 5.}} We prove Lemma \ref{lemma:aux2} in the last step. We first use Berbee's coupling lemma \citep[see Lemma 4.1 in][]{Dedecker2002} to approximate $\sup_{f\in \mathcal{F}}|\sum_{t=0}^{T-1} f(Z_t)|$ by sum of i.i.d. variables. Then we apply the maximal inequality in Corollary 5.1 of \cite{chernozhukov2014gaussian} to bound the expectation of the empirical process. 

Following the discussion below Lemma 4.1 of \cite{Dedecker2002},  we can construct a sequence of random variables $\{Z_{t}^0:t\ge 0\}$ such that
\begin{eqnarray}\label{eqn:step1eq1}
	\sup_{f\in \mathcal{F}}\left|\sum_{t=0}^{T-1} f(Z_t)\right|=\sup_{f\in \mathcal{F}}\left|\sum_{t=0}^{T-1} f(Z_t^0)\right|,
\end{eqnarray}
with probability at least $1-T\beta(q)/q$, and that the sequences $\{U_{2i}^0:i\ge 0\}$ and $\{U_{2i+1}^0:i\ge 0\}$ are i.i.d. where $U_i^0=(Z_{iq}^0,Z_{iq+1}^0,\cdots,Z_{iq+q-1}^0)$. 

Recall that $\mathcal{I}_r=\{q\floor{T/q},q\floor{T/q} +1, \cdots,T-1\}$, we have
\begin{eqnarray*}
	\sup_{f\in \mathcal{F}}\left|\sum_{t=0}^{T-1} f(Z_t^0)\right|\le \sum_{j=0}^{q-1}\sup_{f\in \mathcal{F}} \left|\sum_{t=0}^{\floor{T/q}} f(Z_{tq+j}^0)\right|+\sup_{f\in \mathcal{F}}\left|\sum_{t\in \mathcal{I}_r} f(Z_t^0)\right|.
\end{eqnarray*}
Under the boundedness assumption on $F$, the second term on the right-hand-side (RHS) is bounded from above by $Cq$. Without loss of generality, suppose $\floor{T/q}$ is an even number. The first term on the RHS can be bounded from above by $\sum_{j=0}^{2q-1}\sup_{f\in \mathcal{F}} |\sum_{t=0}^{\floor{T/(2q)}} f(Z_{2tq+j}^0)|$. To summarize, we have shown
\begin{eqnarray}\label{eqn:someeqn}
	\sup_{f\in \mathcal{F}}\left|\sum_{t=0}^{T-1} f(Z_t^0)\right|\le \sum_{j=0}^{2q-1}\sup_{f\in \mathcal{F}} \left|\sum_{t=0}^{\floor{T/(2q)}} f(Z_{2tq+j}^0)\right|+Cq.
\end{eqnarray}

By construction, $\{Z_{2tq}^0:t\ge 0\}$ are i.i.d. It remains to bound $\EE \sup_{f\in \mathcal{F}} |\sum_{t=0}^{\floor{T/(2q)}} f(Z_{2tq}^0)|$. It follows from Corollary 5.1 of \cite{chernozhukov2014gaussian} that
\begin{eqnarray*}
	\EE \sup_{f\in \mathcal{F}} \left|\sum_{t=0}^{\floor{T/(2q)}} f(Z_{2tq}^0)\right|\preceq \sqrt{\frac{v \sigma^2T}{q} \log \left(\frac{AC}{\sigma}\right)}+v C \log \left(\frac{AC}{\sigma}\right).
\end{eqnarray*}
The assertion thus follows from \eqref{eqn:step1eq1}, \eqref{eqn:someeqn} and Markov's inequality.

\subsection{Proof of Theorem \ref{thm5}}
We omit the subscript $\ell$ in $\pi_{\old}^{(\ell)}$ to easy notation. 
Similar to Lemma \ref{lemma:aux1}, we can show that $\widehat{\eta}_1(\pi,\pi_{\old})=\widehat{\eta}_1^*(\pi,\pi_{\old})+o_p\{(NT)^{-1/2}\}$ for any $\pi,\pi_{\old}\in \Pi$, when the nuisance functions converge at rates faster than $o_p((NT)^{-1/4})$. Note that $\widehat{\eta}_1^*(\pi,\pi_{\old})$ is unbiased to $\eta_1(\pi,\pi_{\old})$. It suffices to show \vspace{-0.1cm}
\begin{eqnarray}\label{eqn:CLT}
	\frac{\widehat{\eta}_1^*(\pi,\pi_{\old})-\eta_1(\pi,\pi_{\old})}{\sqrt{\textrm{EB}(\pi,\pi_{\old})}}\stackrel{d}{\to} N(0,1). 
\end{eqnarray}

\vspace{-0.1cm}
\noindent By definition, $\widehat{\eta}_1^*(\pi,\pi_{\old})-\eta_1(\pi,\pi_{\old})=(NT)^{-1}\sum_{i=1}^N \sum_{t=0}^{T-1} \{\psi_2(O_{i,t};\pi,\pi_{\old},Q^{\pi_{\old}},\omega^{\pi_{\old}},d^{\pi_{\old}})+\psi_3(O_{i,t};\pi,\pi_{\old},Q^{\pi_{\old}},\omega^{\pi_{\old}},d^{\pi_{\old}})\}$. The sum on the RHS can be represented as \vspace{-0.1cm}
\begin{eqnarray}\label{eqn:CLT1}
	\frac{1}{NT} \sum_{g=1}^{NT} \{\psi_2(O_{i(g),t(g)};\pi,\pi_{\old},Q^{\pi_{\old}},\omega^{\pi_{\old}},d^{\pi_{\old}})+\psi_3(O_{i(g),t(g)};\pi,\pi_{\old},Q^{\pi_{\old}},\omega^{\pi_{\old}},d^{\pi_{\old}})\},
\end{eqnarray} 

\vspace{-0.1cm}
\noindent where the pair $i(g)$ and $t(g)$ are the unique integers that satisfy $\{i(g)-1\}T+t(g)=g-1$. Under (A1) and (A2), \eqref{eqn:CLT1} corresponds to a sum of martingale difference sequence. Suppose we can show \vspace{-0.1cm}
\begin{eqnarray}\label{eqn:CLT2}
	\begin{split}
		\Var\{\psi_2(O_{i(g),t(g)};\pi,\pi_{\old},Q^{\pi_{\old}},\omega^{\pi_{\old}},d^{\pi_{\old}})+\psi_3(O_{i(g),t(g)};\pi,\pi_{\old},Q^{\pi_{\old}},\omega^{\pi_{\old}},d^{\pi_{\old}})\}\\
		=NT\textrm{EB}(\pi,\pi_{\old}).
	\end{split}	
\end{eqnarray}

Under the given assumptions, we can show the conditions in Theorem 1 of \cite{brown1971martingale} are automatically satisfied. It follows from the martingale central limit theorem developed by \cite{brown1971martingale} that \eqref{eqn:CLT} holds. Consequently, it suffices to show \eqref{eqn:CLT2} holds. 

Note that $\eta_1$ depends only on the transition function $\bar{p}$. For a given tuple $O=(S,A,R,S')$ such that $(A,S)\sim p_{\infty}$, $(S',R|A,S)\sim \bar{p}(\bullet,\bullet|A,S)$, suppose we can show \eqref{eqn:CLT3} holds, \vspace{-0.1cm}
\begin{eqnarray}\label{eqn:CLT3}
	\nabla_{\theta_1} \eta_1(\theta_1^*)=\EE \sum_{j=2}^3\psi_j(O;\pi,\pi_{\old},Q^{\pi_{\old}},\omega^{\pi_{\old}},d^{\pi_{\old}}) \nabla_{\theta_1} \log \bar{p}_{\theta_1^*}(\bullet,\bullet|A,S). 
\end{eqnarray}

\vspace{-0.1cm}
\noindent Then it follows from Cauchy-Schwarz inequality that \vspace{-0.1cm}
\begin{eqnarray*}
	NT\textrm{EB}(N,T)=\sup \nabla_{\theta_1} \eta_1(\theta_1^*)  \left\{ \EE \nabla_{\theta_1}\log \bar{p}_{\theta_1^*}(S',R|A,S)\nabla_{\theta_1}\log \bar{p}_{\theta_1^*}^\top(S',R|A,S) \right\}^{-1} \{ \nabla_{\theta_1} \eta_1(\theta_1^*)\}^\top\\
	\le \EE \left[\sum_{j=2}^3\psi_j(O;\pi,\pi_{\old},Q^{\pi_{\old}},\omega^{\pi_{\old}},d^{\pi_{\old}}) \right]\left[\sum_{j=2}^3\psi_j(O;\pi,\pi_{\old},Q^{\pi_{\old}},\omega^{\pi_{\old}},d^{\pi_{\old}}) \right]^\top.
\end{eqnarray*}

\vspace{-0.1cm}
\noindent This implies that the variance of the proposed estimator is asymptotically greater than or equal to the efficiency bound. Moreover, note that for either $j=2$ or $3$, we have $\EE \{\psi_j(O;\pi,\pi_{\old},Q^{\pi_{\old}},\omega^{\pi_{\old}},d^{\pi_{\old}})|A,S\}=0$, almost surely. Using similar arguments in the proof of Theorem 2 of \cite{kallus2019efficiently}, we can show that there exists a sequence of regular parametric submodels whose Cr{\'a}mer-Rao lower bound approaches to \vspace{-0.1cm}
\begin{eqnarray*}
	\EE \left[\sum_{j=2}^3\psi_j(O;\pi,\pi_{\old},Q^{\pi_{\old}},\omega^{\pi_{\old}},d^{\pi_{\old}}) \right]\left[\sum_{j=2}^3\psi_j(O;\pi,\pi_{\old},Q^{\pi_{\old}},\omega^{\pi_{\old}},d^{\pi_{\old}}) \right]^\top.
\end{eqnarray*}

\vspace{-0.1cm}
\noindent This implies that the variance of the proposed estimator is asymptotically equal to the efficiency bound. The proof is hence completed. 

It remains to show \eqref{eqn:CLT3}. 
We first observe that, the advantage function and the discounted visitation probability are completely determined by the transition function $\bar{p}$. By the chain rule, \vspace{-0.1cm}
\begin{eqnarray*}
	\nabla_{\theta_1} \eta_1(\theta_1^*)=\sum_{a,s} \pi(a|s)\{\nabla_{\theta_1} A^{\pi_{\old}}(a,s;\theta_1^*)\} d^{\pi_{\old},\nu}(s)+\sum_{a,s} \pi(a|s) A^{\pi_{\old}}(a,s) \nabla_{\theta_1}d^{\pi_{\old},\nu}(s;\theta_1^*),
\end{eqnarray*}

\vspace{-0.1cm}
\noindent where $A^{\pi_{\old}}(\bullet,\bullet;\theta_1)$ and $d^{\pi_{\old},\nu}(\bullet;\theta_1)$ denote the advantage and the discounted visitation probability under certain parametric submodel for $\bar{p}$ indexed by $\theta_1$. 

Using similar arguments in the proof of Theorem 2 in \cite{kallus2019efficiently}, we can show that \vspace{-0.1cm}
\begin{eqnarray*}
	\{\nabla_{\theta_1} A^{\pi_{\old}}(a,s;\theta_1^*)\}=\frac{1}{1-\gamma}\EE \left\{\omega^{\pi_{\old}}(A,S;a,s)-\sum_{a'} \pi_{\old}(a'|s) \omega^{\pi_{\old}}(A,S;a',s) \right\}\\
	\times \{R+\gamma V^{\pi_{\old}}(S')-Q^{\pi_{\old}}(A,S)\}\nabla_{\theta_1}\log \bar{p}_{\theta_1^*}^\top(S',R|A,S). 
\end{eqnarray*}

\vspace{-0.1cm}
\noindent This in turns yields that \vspace{-0.1cm}
\begin{eqnarray*}
	\sum_{a,s} \pi(a|s)\{\nabla_{\theta_1} A^{\pi_{\old}}(a,s;\theta_1^*)\} d^{\pi_{\old},\nu}(s)=\EE \psi_2(O;\pi,\pi_{\old},Q^{\pi_{\old}},\omega^{\pi_{\old}},d^{\pi_{\old}})
	\nabla_{\theta_1}\log \bar{p}_{\theta_1^*}^\top(S',R|A,S).
\end{eqnarray*}

\vspace{-0.1cm}
\noindent It remains to show
\begin{eqnarray*}
	\sum_{a,s} \pi(a|s) A^{\pi_{\old}}(a,s) \nabla_{\theta_1}d^{\pi_{\old},\nu}(s;\theta_1^*)=\EE \psi_3(O;\pi,\pi_{\old},Q^{\pi_{\old}},\omega^{\pi_{\old}},d^{\pi_{\old}})
	\nabla_{\theta_1}\log \bar{p}_{\theta_1^*}^\top(S',R|A,S).
\end{eqnarray*}

\vspace{-0.1cm}
\noindent For a given parametric submodel $\{\bar{p}_{\theta_1}:\theta_1\in \Theta_1\}$, let $p(s'|a,s;\theta_1)=\sum_{r} \bar{p}(s',r|a,s;\theta_1)$. With some calculations, we have \vspace{-0.1cm}
\begin{eqnarray*}
	\frac{1}{\gamma(1-\gamma)}\nabla_{\theta_1} d^{\pi_{\old},\nu}(s;\theta_1^*)=\sum_{t\ge 0} \gamma^t\sum_{\{(a_j,s_j)\}_{j=0}^t} \nabla_{\theta_1}\left\{\prod_{j=0}^t \pi_{\old}(a_j|s_j)  p(s|a_j,s_j;\theta_1^*)\right\}\nu(s_0)\\
	=\sum_{t\ge 0} \gamma^t\sum_{\{(a_j,s_j)\}_{j=0}^t,s_{t+1}} \mathbb{I}(s_{t+1}=s) \nabla_{\theta_1}\left\{\prod_{j=0}^t \pi_{\old}(a_j|s_j)  p(s_{j+1}|a_j,s_j;\theta_1^*)\right\}\nu(s_0)\\
	= \sum_{t\ge 0} \gamma^t \sum_{\substack{\{(a_j,s_j)\}_{j=0}^t \\ s_{t+1}}} \mathbb{I}(s_{t+1}=s) \prod_{j=0}^t \pi_{\old}(a_j|s_j)  p(s_{j+1}|a_j,s_j)\sum_{k=0}^t \nabla_{\theta_1} \log p(s_{k+1}|a_k,s_k;\theta_1^*) \nu(s_0) \\
	=\frac{1}{1-\gamma} \sum_{k=0}^{+\infty} \gamma^k \sum_{\substack{\{(a_j,s_j)\}_{j=0}^k \\ s_{k+1},a}} \pi_{\old}(a|s_{k+1}) d^{\pi_{\old}}(s;a,s_{k+1}) \nabla_{\theta_1} \log p(s_{k+1}|a_k,s_k;\theta_1^*)\\ \times \left\{\prod_{j=0}^k \pi_{\old}(a_j|s_j)  p(s_{j+1}|a_{j},s_{j})\right\}\nu(s_0).
\end{eqnarray*}

\vspace{-0.1cm}
\noindent By definition of $\omega^{\pi_{\old}}$, the last equation can be rewritten as \vspace{-0.1cm}
\begin{eqnarray*}
	\frac{1}{(1-\gamma)^2} \sum_{a} \pi_{\old}(a|S')d^{\pi_{\old}}(s;a,S')\omega^{\pi_{\old},\nu}(A,S)\nabla_{\theta_1} \log p(S'|A,S;\theta_1^*).
\end{eqnarray*}

Note that $\EE h(A,S) \nabla_{\theta_1}\log p(S'|A,S;\theta_1^*)=0$ for any function $h$, we obtain\vspace{-0.1cm}
\begin{eqnarray*}
	\nabla_{\theta_1} d^{\pi_{\old},\nu}(s;\theta_1^*)=\frac{1}{1-\gamma} \EE \left[\gamma\sum_{a} \pi_{\old}(a|S')d^{\pi_{\old}}(s;a,S')-\EE \left\{\gamma\sum_{a} \pi_{\old}(a|S')d^{\pi_{\old}}(s;a,S')|A,S\right\} \right]\\
	\times \omega^{\pi_{\old},\nu}(A,S)\nabla_{\theta_1} \log p(S'|A,S;\theta_1^*)\\
	=\frac{1}{1-\gamma} \EE\left[\gamma\sum_{a} \pi_{\old}(a|S')d^{\pi_{\old}}(s;a,S')-d^{\pi_{\old}}(s;A,S)+(1-\gamma)\mathbb{I}(s=S) \right]\\
	\times \omega^{\pi_{\old},\nu}(A,S)\nabla_{\theta_1} \log p(S'|A,S;\theta_1^*).
\end{eqnarray*}

\vspace{-0.1cm}
\noindent Consequently, we obtain
\begin{eqnarray*}
	\sum_{a,s} \pi(a|s) A^{\pi_{\old}}(a,s) \nabla_{\theta_1}d^{\pi_{\old},\nu}(s;\theta_1^*)=\EE \psi_3(O;\pi,\pi_{\old},Q^{\pi_{\old}},\omega^{\pi_{\old}},d^{\pi_{\old}})
	\nabla_{\theta_1}\log \bar{p}_{\theta_1^*}^\top(S',R|A,S).
\end{eqnarray*}
The proof is hence completed. 

\section{Some additional numerical details}\label{sec:addnum}

\subsection{Additional Results on The Toy Example}
In this subsection, we first list some details of our toy example. In particular, we consider the following transition matrix:
\begin{equation*}
	p= 	\bordermatrix{~&S' = 0 &S' = 1\cr
		(S = 0, A=0)&0.75&0.25\cr
		(S = 0, A=1)&0.4&0.6\cr
		(S = 1, A = 0)&0.1&0.9\cr
		(S = 1, A = 1)&0.85&0.15\cr
	} 
\end{equation*}
The behavior policy to generate the simulated data is
\begin{equation*}
	b = 	\bordermatrix{~&A = 0 &A = 1\cr
		S = 0&0.7&0.3\cr
		S = 1&0.2&0.8\cr
	} .
\end{equation*}
Below is the detailed design of each scenario for testing the value enhancement property and the triple robustness.
\begin{itemize}
	\item[(i)] origin: all the nuisance functions are set to their oracle values. 
	\item[(ii)] mod1: we inject $ \text{uniform}(0,2) $ noises to the marginal density ratio, whereas other nuisance functions are set to their oracle values.
	\item[(iii)] mod2: we inject $ \text{uniform}(0,2) $ noises to the Q-function, whereas other nuisance functions are set to their oracle values.
	\item[(iv)] mod3: we multiply the transition matrix $p$ by a random variable following $ \text{exponential}(1)$ and clip all values into $ [0,1] $, whereas other nuisance functions are set to their oracle values.
	\item[(vi)] mod4: we inject random errors to all the three nuisance functions according to the procedures described in (ii)-(iv). 
\end{itemize}

\begin{figure}[H]
	\centering
	\includegraphics[width=0.6\linewidth,trim=150 50 150 50]{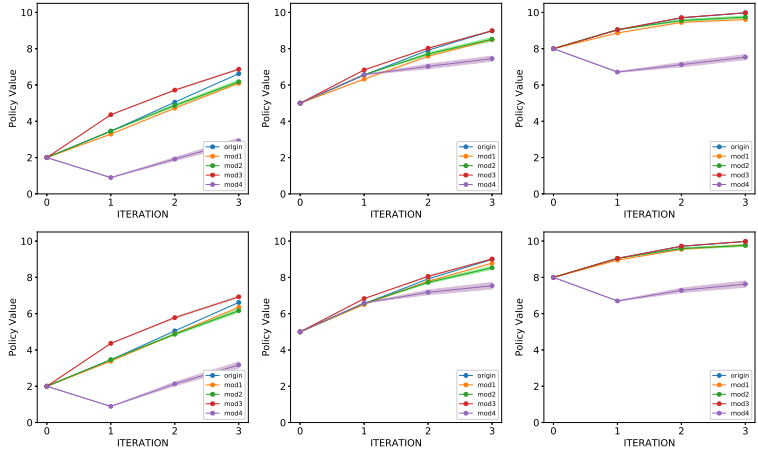}
	\caption{
		Values of estimated policies in a toy example. 
		First row represents results using  $ (T,N) $ pair as $ (30,30) $ while the second row using $ (50,50) $. The three columns represents initial policy factor $ \kappa $ taking values $ 0.8,0.5,0.2 $ respectively. The horizontal axis represents the number of iterations used in our value enhancement procedure. When iteration equals zero, we plot the evaluation value for the initial policy. 
		The optimal value is $10$ and $\delta$ is fixed to 0.05.  The confidence band is computed based on 100 replications. }
	\label{fig:toy0.05}
\end{figure}

\begin{figure}[H]
	\centering
	\includegraphics[width=0.6\linewidth,trim=150 50 150 50]{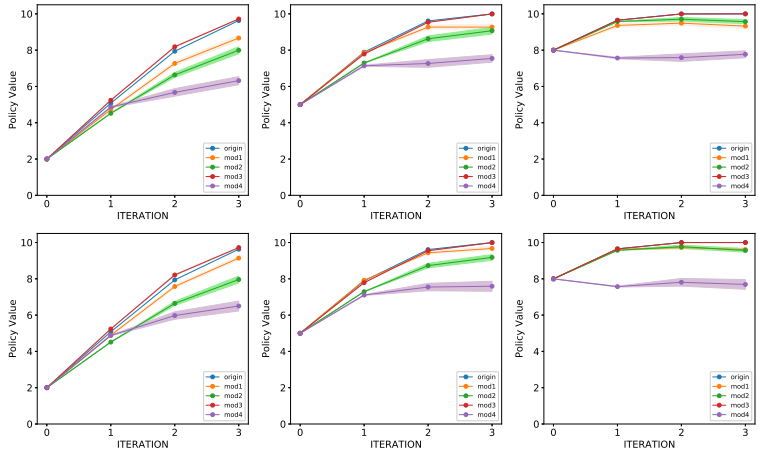}
	\caption{
		Values of estimated policies in a toy example. 
		First row represents results using  $ (T,N) $ pair as $ (30,30) $ while the second row using $ (50,50) $. The three columns represents initial policy factor $ \kappa $ taking values $ 0.8,0.5,0.2 $ respectively. The horizontal axis represents the number of iterations used in our value enhancement procedure. When iteration equals zero, we plot the evaluation value for the initial policy. 
		The optimal value is $10$ and $\delta$ is fixed to 0.2.  The confidence band is computed based on 100 replications. }
	\label{fig:toy0.2}
\end{figure}

We next report values of estimated policies in the toy example (see Section \ref{sec:toy}) where $\delta$ is set to 0.05 and 0.2. These values are depicted in Figure \ref{fig:toy0.05} and \ref{fig:toy0.2}, respectively. 

{\color{black}
	Furthermore, to demonstrate the advantage of the proposed method, we use lookup tables (e.g., linear models with table lookup features) instead of deep learning models to parametrize all nuisance functions (including the Q-function, the probability ratio and the transition kernel), and apply the proposed method to the toy example setting in Section 5.1 of the main text. Results are reported in Figure \ref{fig:toy_est}. It can be seen that the proposed method is still able to improve the performance of initial policies.
	
	\begin{figure}[t]
		\centering
		\includegraphics[width=0.75\linewidth]{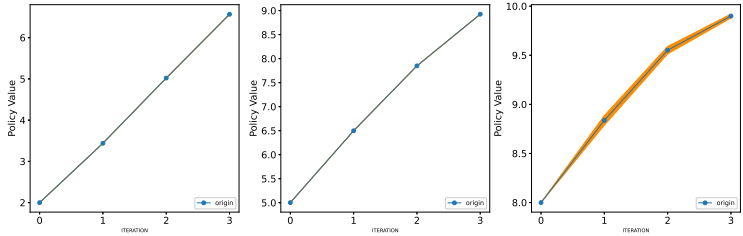}
		\caption{
			Values of estimated policies in the toy example. All nuisance parameters and the policy are modelled by linear functions. 
			Results are computed using  $ (T,N) $ pair as $ (100,100) $. These three figures represents the initial policy factor $ \kappa $ taking values $ 0.8,0.5,0.2 $ respectively. The horizontal axis represents the number of iterations used in our value enhancement procedure. When the iteration equals zero, we plot the evaluation value for the initial policy. 
			The optimal value is $10$ and $\delta$ is fixed to 0.05.  The confidence band is computed based on 100 replications. The variances of the values in the first two panels are very small such that the upper and lower confidence bands are largely overlapped.}
		\label{fig:toy_est}
	\end{figure}

	\subsection{One Additional Simulation Study}\label{sec:candidate}
	In this subsection, we conduct another simulation study to show the finite sample performance of our proposed method. Consider a 15-dimensional state vector $ S_{t} =(S^{(1)}_{t},\cdots,S^{(15)}_{t})$. We set initiate state as a standard normal vector. Let the first two state variables evolve according to: for $t \geq 1$
	\begin{eqnarray*}
		S^{(1)}_{t} = (3/4)(2A_{t-1}-1)S^{(1)}_{t-1}+(1/4)S^{(2)}_{t-1}+\epsilon_t^{(1)},\\
		S^{(2)}_{t} = (3/4)(1- 2A_{t-1})S^{(2)}_{t-1}+(1/4)S^{(1)}_{t-1}+\epsilon_t^{(2)},
	\end{eqnarray*}
	where $ A_{t} $ takes values in $ \{0,1\}$ with equal probabilities and $\{\epsilon_t^{(1)}\}_t$, $\{\epsilon_t^{(2)}\}_t$ are i.i.d. $ N(0,1/4) $ random errors. 
	The data generating mechanism for these two variable are similar to those considered in the simulation settings of  \cite{luckett2019estimating} and \cite{liao2020batch}. 
	Other state variables are sampled independently from the standard normal distribution for all $0 \leq t \leq T-1$. Define the reward function by $ R_{t} = 2S^{(1)}_{t+1} + S^{(2)}_{t+1}-(1/4)(2A_{t}-1) $. 
	The sample size pair $ (T,N) $ is set to be $ (50,100)$, $(25,200)$ or $(100,50) $ in our experiment. 
	We consider two choices of $\gamma$, corresponding to $0.9$ and $0.95$. 
	
	To implement our method, we set the number of folds $ \mathbb{L} $ to 2 and the constant $\delta$ to 0.05. In order to obtain $\pi_{\old}$, as discussed in Section \ref{sec:oldlearn}, we apply VL, FQI and CQL to compute three different initial policies in our experiment.
	Both FQI and CQL require to model the optimal Q-function. In our implementation, we use a rectified linear unit (ReLU) neural network with two hidden layers. 
	To implement V-learning, we use the R-package developed by \cite{luckett2019estimating}. In particular, we use RBF basis functions to model the state value function and linear basis functions to model the policy class. We also use a rectified linear unit (ReLU) neural network with two hidden layers to model $\pi_{\new}$. 
	
	Results are summarized in Figure \ref{fig:candidate}. 
	It can be seen that all these initial policies have the potential to be improved based on our procedure. This
	demonstrates the superior performance of our value enhancement method. In addition, the improvement is substantial when $\pi_{\old}$ is not very close to the optimal policy. This is consistent with our observations in the toy example. It can also be seen that our method may suffer from some slight value loss when $\pi_{\old}$ is very close to the optimal one. This is probably due to that we used a fixed $\delta$ in the simulation. 
	{\color{black}As shown in the toy example}, we should choose a large $\delta$ when $\pi_{\old}$ is far away from the optimal policy and a small $\delta$ otherwise. It will be interesting to study how to adaptively choose $\delta$. However, this is beyond the scope of the current paper. We leave it for future work. {\color{black}Finally, we conduct some additional studies to investigate the performance of the proposed method in settings with a smaller sample size and a higher noise level. In particular, Figure \ref{fig:toy3} reported results where $N\times T=3000$. Figure \ref{fig:toy4} reported results where $\{\epsilon_t^{(1)}\}_t$, $\{\epsilon_t^{(2)}\}_t$ are i.i.d. $N(0,1)$ random errors. Findings are similar to those in Figure \ref{fig:candidate}}.
	%
	
	\begin{figure}[H]
		\centering
		\includegraphics[width=0.6\linewidth,trim=150 50 150 50]{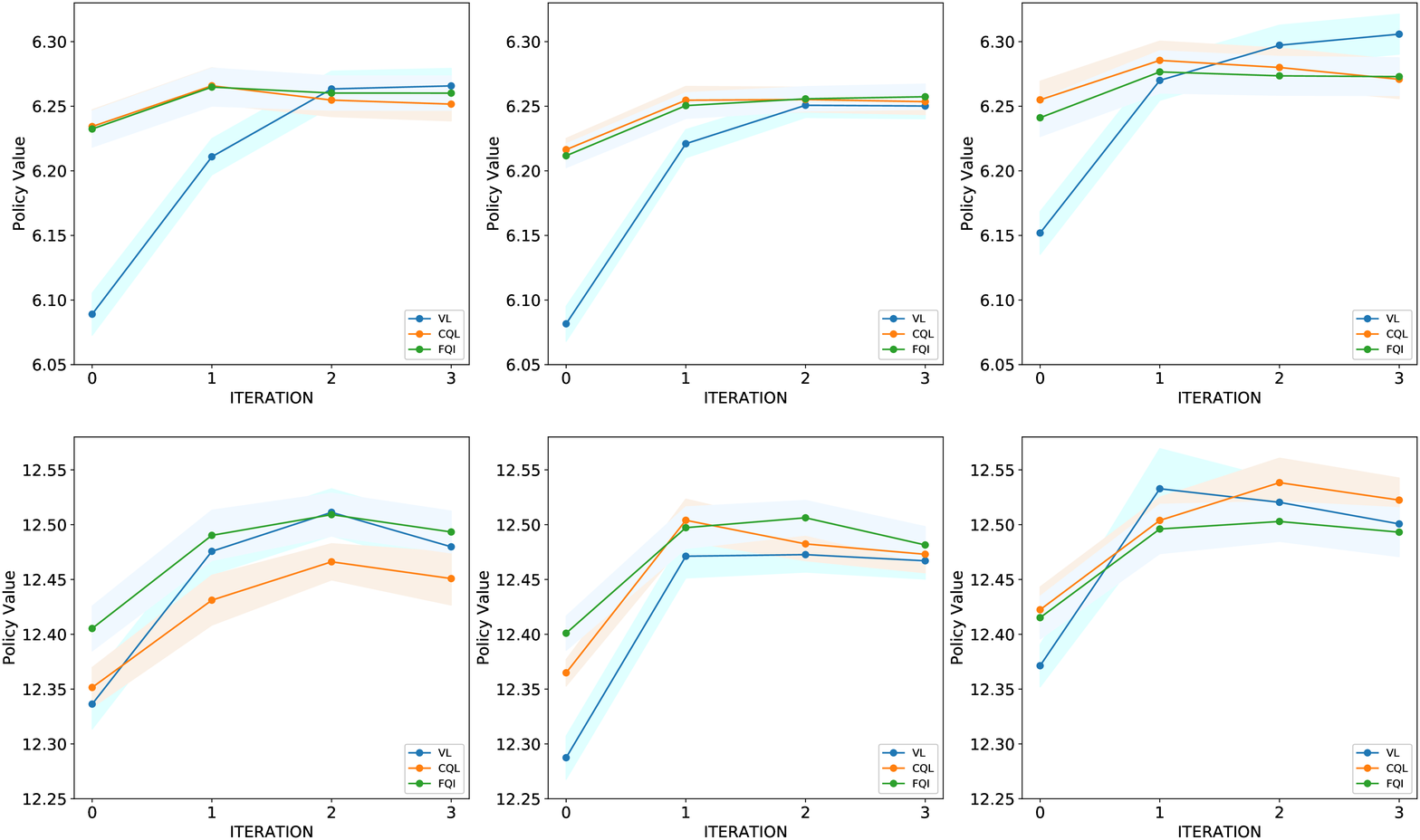}
		\caption{Values of our estimated policies where initial ones are computed by VL, CQL, FQI. The first row represents results using $ \gamma=0.9 $ while the second row using $ \gamma=0.95 $. Three columns represents using $ (T,N) $ pair as $ (50,100),(25,200),(100,50) $ respectively. The optimal value under $\gamma = 0.9$ is approximately $6.89$ and $13.47$ under $\gamma =0.95$. The confidence band is computed based on 100 replications.}
		\label{fig:candidate}
	\end{figure}

	\begin{figure}[H]
		\centering
		\includegraphics[width=0.6\linewidth,trim=150 50 150 50]{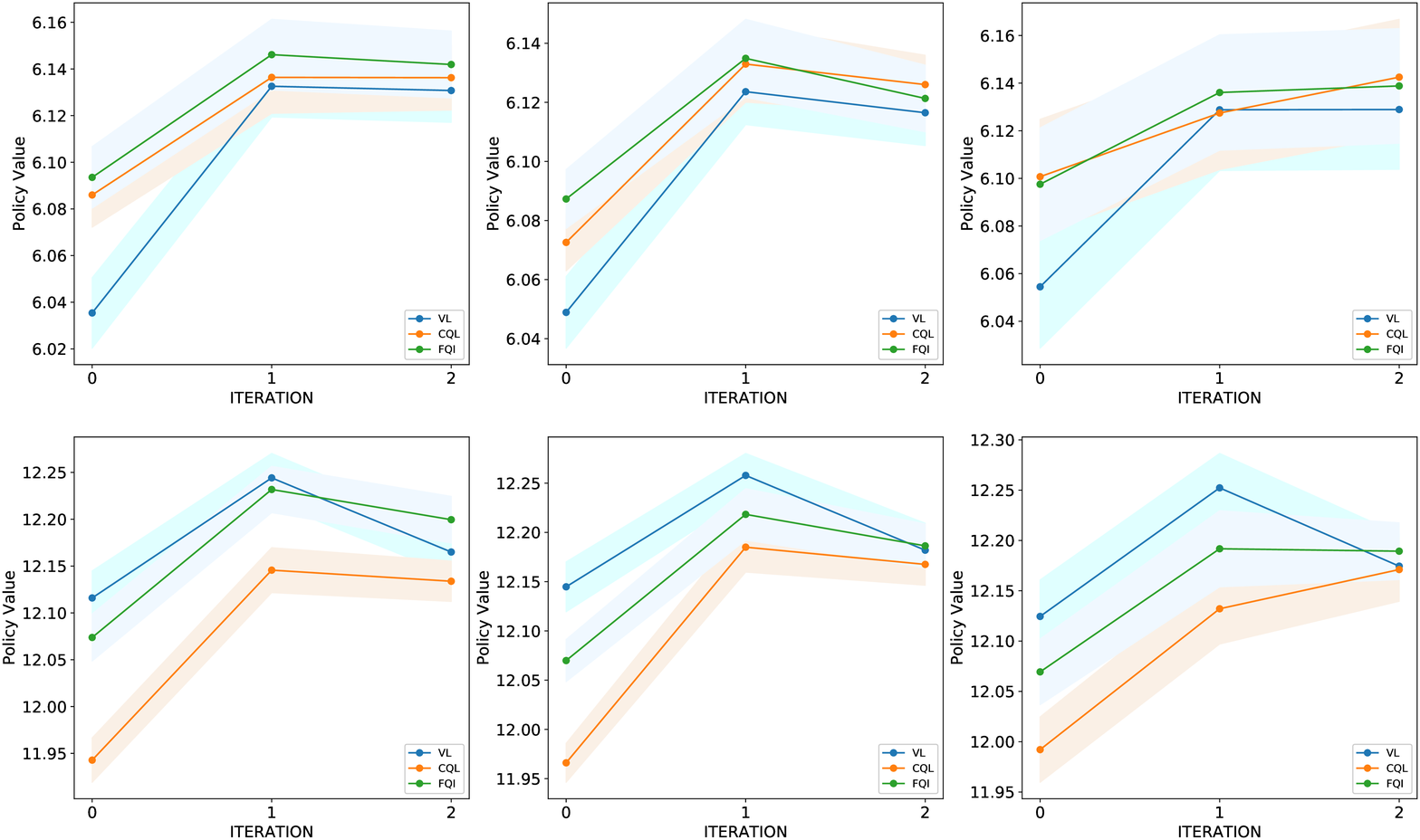}
		\caption{
			Values of our estimated policies where initial ones are computed by VL, CQL, FQI. The first row represents results using $ \gamma=0.9 $ while the second row using $ \gamma=0.95 $. Three columns represents using $ (T,N) $ pair as $ (30,100),(15,200),(100,30) $ respectively. The optimal value under $\gamma = 0.9$ is approximately $6.89$ and $13.47$ under $\gamma =0.95$. The confidence band is computed based on 100 replications.}
		\label{fig:toy3}
	\end{figure}
	
	\begin{figure}[H]
		\centering
		\includegraphics[width=0.6\linewidth,trim=150 50 150 50]{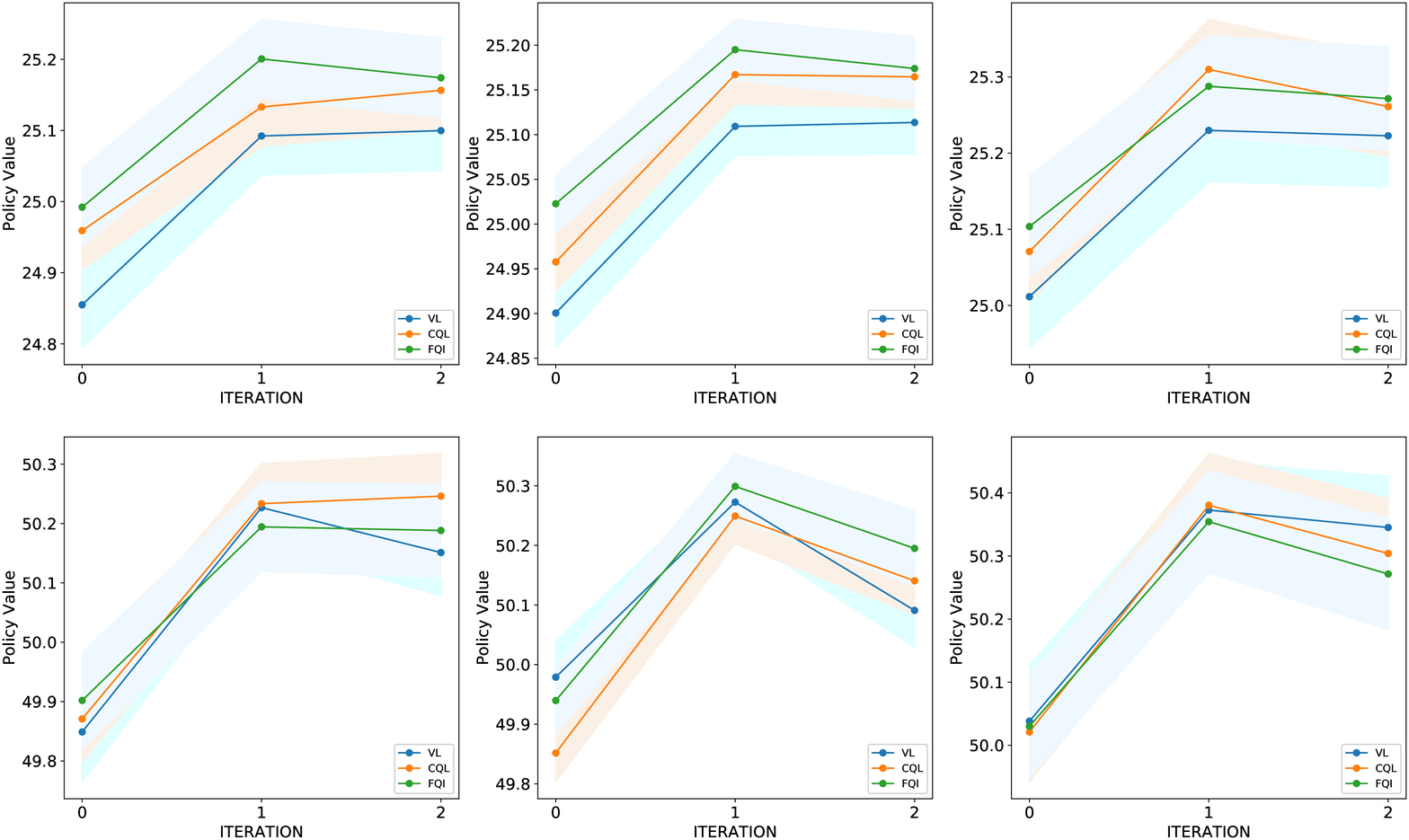}
		\caption{
			Values of our estimated policies where initial ones are computed by VL, CQL, FQI. The first row represents results using $ \gamma=0.9 $ while the second row using $ \gamma=0.95 $. Three columns represents using $ (T,N) $ pair as $ (50,100),(25,200),(100,50) $ respectively. The confidence band is computed based on 100 replications.}
		\label{fig:toy4}
	\end{figure}
}

\bibliographystyle{agsm}
\bibliography{reference}

\end{document}